%% file: main.tex
\documentclass[twoside,11pt]{article}
\usepackage{jair, theapa, rawfonts}
\pdfoutput=1
\usepackage{amsmath}
\usepackage{amsthm}
\usepackage{amssymb}
\usepackage{graphicx}
\usepackage{tabularx}
\usepackage{mathtools}
\usepackage[shortlabels]{enumitem}

\usepackage{algorithm,algorithmic}
\usepackage{multirow}
\usepackage{appendix}

\DeclareSymbolFont{bsfletters}{OT1}{cmss}{bx}{n}
\DeclareSymbolFont{ssfletters}{OT1}{cmss}{m}{n}
\DeclareMathSymbol{\bsfGamma}{0}{bsfletters}{'000}
\DeclareMathSymbol{\ssfGamma}{0}{ssfletters}{'000}
\DeclareMathSymbol{\bsfDelta}{0}{bsfletters}{'001}
\DeclareMathSymbol{\ssfDelta}{0}{ssfletters}{'001}
\DeclareMathSymbol{\bsfTheta}{0}{bsfletters}{'002}
\DeclareMathSymbol{\ssfTheta}{0}{ssfletters}{'002}
\DeclareMathSymbol{\bsfLambda}{0}{bsfletters}{'003}
\DeclareMathSymbol{\ssfLambda}{0}{ssfletters}{'003}
\DeclareMathSymbol{\bsfXi}{0}{bsfletters}{'004}
\DeclareMathSymbol{\ssfXi}{0}{ssfletters}{'004}
\DeclareMathSymbol{\bsfPi}{0}{bsfletters}{'005}
\DeclareMathSymbol{\ssfPi}{0}{ssfletters}{'005}
\DeclareMathSymbol{\bsfSigma}{0}{bsfletters}{'006}
\DeclareMathSymbol{\ssfSigma}{0}{ssfletters}{'006}
\DeclareMathSymbol{\bsfUpsilon}{0}{bsfletters}{'007}
\DeclareMathSymbol{\ssfUpsilon}{0}{ssfletters}{'007}
\DeclareMathSymbol{\bsfPhi}{0}{bsfletters}{'010}
\DeclareMathSymbol{\ssfPhi}{0}{ssfletters}{'010}
\DeclareMathSymbol{\bsfPsi}{0}{bsfletters}{'011}
\DeclareMathSymbol{\ssfPsi}{0}{ssfletters}{'011}
\DeclareMathSymbol{\bsfOmega}{0}{bsfletters}{'012}
\DeclareMathSymbol{\ssfOmega}{0}{ssfletters}{'012}

\DeclareMathOperator*{\argmax}{arg\,max}

\DeclareMathAlphabet{\mathbsf}{OT1}{cmss}{bx}{n}
\DeclareMathAlphabet{\mathssf}{OT1}{cmss}{m}{sl}

\usepackage{booktabs}
\usepackage{dsfont}

\usepackage{subcaption}

\ShortHeadings{Steady-State Planning in Expected Reward Multichain MDPs}
{Atia \textit{et al.}} 
\firstpageno{1}

\input{preamble}

\begin{document}

\title{\color{black} Steady-State Planning in Expected Reward Multichain MDPs}  

\author{\name George K. Atia \email george.atia@ucf.edu \\
\addr Department of Electrical and Computer Engineering\\
Department of Computer Science\\
University of Central Florida, FL 32816, USA
       \AND
       \name Andre Beckus \email andre.beckus@us.af.mil \\
       \addr Air Force Research Laboratory, 
       NY  13441, USA
       \AND
       \name Ismail Alkhouri \email ialkhouri@knights.ucf.edu \\
       \addr Department of Electrical and Computer Engineering\\
       University of Central Florida, FL 32816, USA
       \AND
       \name Alvaro Velasquez \email alvaro.velasquez.1@us.af.mil \\
       \addr Air Force Research Laboratory, 
       NY  13441, USA}

\maketitle

\begin{abstract}  
The planning domain has experienced increased interest in the formal synthesis of decision-making policies. This formal synthesis typically entails finding a policy which satisfies formal specifications in the form of some well-defined logic. While many such logics have been proposed with varying degrees of expressiveness and complexity in their capacity to capture desirable agent behavior, their value is limited when deriving decision-making policies which satisfy certain types of asymptotic behavior in general system models. In particular, we are interested in specifying constraints on the steady-state behavior of an agent, which captures the proportion of time an agent spends in each state as it interacts for an indefinite period of time with its environment. {\color{black} This is sometimes called the average or expected behavior of the agent and the associated planning problem is faced with significant challenges unless strong restrictions are imposed on the underlying model in terms of the connectivity of its graph structure.}
In this paper, we explore this steady-state planning problem that consists of deriving a decision-making policy for an agent such that constraints on its steady-state behavior are satisfied. A linear programming solution for the general case of multichain Markov Decision Processes (MDPs) is proposed and we prove that optimal solutions to the proposed programs yield stationary policies with rigorous guarantees of behavior.  
\end{abstract}



\input{body.tex}


\bibliographystyle{theapa}
\bibliography{main,sample-bibliography,ijcai20}  

\end{document}

%% file: preamble.tex
\newcommand{\Allstates}{S}

\newcommand{\NumErgSets} {m}

\newcommand{\SSSpec} {\Phi^\infty_L}
\newcommand{\LMDPSpec} {\Phi_L}
\newcommand{\absorbprob}[2]{p_{{#1}{#2}}}
\newcommand{\ssdist}{\mathrm{Pr}^\infty}
\newcommand{\ssdistbf}{\mathbf{Pr^\infty}}

\newcommand{\prbetapi}{\mathrm{Pr}^\infty_{\pi}}
\newcommand{\prbetapol}[1]{\mathrm{Pr}^\infty_{#1}}
\newcommand{\setssdistbeta}{\calP^\infty}
\newcommand{\cesarolimit}[1]{#1^\infty}
\newcommand{\classpreservingset}{\Pi_\textit{CP}}
\newcommand{\stationaryset}{\Pi_\textit{S}}
\newcommand{\edgepreservingset}{\Pi_\textit{EP}}

\newcommand{\uptounichainset}{\Pi_\textit{CPU}}

\newcommand{\TransExpect} {\zeta}

\newcommand{\recurrentset}[1]{r(\calM_{#1})}
\newcommand{\tscc}[2]{r_{#1}(\calM_{#2})}
\newcommand{\transientset}[1]{\bar{r}(\calM_{#1})}
\newcommand{\translabelset} {L^{tr}}
\newcommand{\TransSpec}[1] {\Phi^{tr}_{#1}}

\newcommand{\AvgVisitsNoarg}[2]{f_{{#1},{#2}}}
\newcommand{\AvgVisits}[3]{f_{{#1},{#2}}(#3)}
\newcommand{\AvgReward}[2]{g_{{#1},{#2}}}
\newcommand{\VisitsNoarg}[2]{h_{{#1},{#2}}}
\newcommand{\Visits}[3]{h_{{#1},{#2}}(#3)}
\newcommand{\FirstRecurrent}[1]{\phi_{{#1}}}

\newcounter{LPmainconstraintcount}
\renewcommand\theLPmainconstraintcount{(\textit{\roman{LPmainconstraintcount}})}
\makeatletter
\newcommand\LPmainconstraintlabel[1]{\refstepcounter{LPmainconstraintcount}\theLPmainconstraintcount\ltx@label{#1}}
\makeatother

\newcounter{LPunichainconstraintcount}
\renewcommand\theLPunichainconstraintcount{(\textit{\roman{LPunichainconstraintcount}})}
\makeatletter
\newcommand\LPunichainconstraintlabel[1]{\refstepcounter{LPunichainconstraintcount}\theLPunichainconstraintcount\ltx@label{#1}}
\makeatother

\catcode`~=11 \def\UrlSpecials{\do\~{\kern -.15em\lower .7ex\hbox{~}\kern .04em}} \catcode`~=13

\allowdisplaybreaks[3]

\newcommand{\nn}{\nonumber}

\newcommand{\calM}{\mathcal{M}}

\newcommand{\calP}{\mathcal{P}}

\newcommand{\bbR}{\mathbb{R}}

\newtheorem{definition}{Definition}

\newtheorem{lemma}{Lemma}

\newtheorem{theorem}{Theorem}

\newtheorem{corollary}{Corollary}

\newtheorem{proposition}{Proposition}

\newtheorem{data model}{Data Model}
\newtheorem{example}{Example}

\newcommand{\qednew}{\nobreak \ifvmode \relax \else
      \ifdim\lastskip<1.5em \hskip-\lastskip
      \hskip1.5em plus0em minus0.5em \fi \nobreak
      \vrule height0.75em width0.5em depth0.25em\fi}

\newcommand \IndicatorFunction {{\mathds{1}}}

\def\whp/{whp}

\usepackage[dvipsnames]{xcolor}
\usepackage{acronym}
\newcommand{\tred}[1]{\textcolor{red}{#1}}

\definecolor{forestgreen}{rgb}{0.13, 0.55, 0.13}

\newacro{name}[SSPS]{Steady-State Policy Synthesis}
\newacro{LP}[LP]{linear program}
\newacro{LPs}[LPs]{linear programs}

%% file: body.tex
\section{Introduction}
%
	
	{\color{black} The proliferation and mass adoption of automated solutions in recent years has led to an increased concern in the verification, validation, and trust of the prescribed agent behavior \cite{motivation}.} This motivates the need for traditional techniques which can yield guarantees of behavior. The study of such techniques has largely been the focus of areas such as formal synthesis and planning, where it is common to derive decision-making policies for agents acting in an environment such that some given formal specification is satisfied. The majority of prior art in this area entails the coupling of some formal logic with the model of agent-environment dynamics in order to find optimal policies which satisfy specifications expressed in said logic. Examples include planning with Linear Temporal Logic (LTL) \cite{LTLPlanning}, probabilistic LTL (PLTL) \cite{PLTLPlanning}, LTL over finite traces (LTL$_f$) \cite{LTLfPlanning}, Linear Dynamic Logic (LDL$_f$) \cite{LDLfPlanning}, Computation Tree Logic (CTL) \cite{CTLPlanning}, probabilistic CTL (PCTL) \cite{PCTLPlanning}, Signal Temporal Logic (STL) \cite{STLPlanning}, Chance-Constrained Temporal Logic (C2TL) \cite{C2TLPlanning}, Continuous Stochastic Logic (CSL) \cite{CSLControllerSynthesis}, $\mu$-Calculus \cite{MuCalculusPlanning}, Metric Temporal Logic (MTL) \cite{MTLPlanning}, and logic fragments, such as the Rank-1 Generalized Reactivity (GR[1]) formulas of LTL \cite{TuLiP}. Formal multi-agent planning has also been explored using Dynamic Epistemic Logic \cite{DELPlanning} and Alternating-time Temporal Logic (ATL) \cite{ATLPlanning}. 
        
    The use of the foregoing logics has facilitated the growth of solutions to the aforementioned planning problems and are a good conduit for verifying, explaining, and yielding provably correct agent behavior and, consequently, establishing a measure of trust. However, these logics are either insufficient to reason about the asymptotic behavior that is captured by the steady-state distribution of the agent as it follows some decision-making policy, or existing solutions to the corresponding planning problems make strong assumptions on the underlying model of the system. Solutions to these challenges have gained traction in recent years. Indeed, there has been increased interest in what we refer to as the steady-state planning problem of computing decision-making policies that satisfy constraints on the resulting steady-state behavior. In particular, progress has been made in easing the restrictions required on the agent-environment dynamics model, usually expressed in the form of a Markov Decision Process (MDP), in order to derive a solution policy. In this paper, we advance the state-of-the-art in steady-state planning by establishing the first solution to steady-state planning in multichain MDPs such that the resulting stationary policy satisfies constraints imposed on the steady-state distribution of the agent. Our approach also dissolves assumptions of ergodicity or recurrence of the underlying MDP which are often made in the literature when reasoning about steady-state distributions.
        

	
Steady-state planning has applications in several areas, such as deriving maintenance plans for various systems, including aircraft maintenance, where the asymptotic failure rate of components must be kept below some small threshold \cite{minFailureRate_7,minFailureRate_8}. Optimal routing problems for communication networks have also been proposed in which data throughput must be maximized subject to constraints on average delay and packet drop metrics \cite{constrainedRoutingMotivation}. This includes constraints on steady-state network behavior, which include steady-state network frequency and steady-state phase or timing errors \cite{steadyStateNetworkBehavior}. There is also the potential of leveraging solutions in the steady-state planning problem space to the design of intelligent space satellites. Indeed, this is an area where the steady-state distribution of debris following some orbit can be computed to reason about the probability of a satellite colliding with said debris. Such information has been used to determine human-driven control policies for tasks such as debris mitigation or debris removal \cite{debrisMitigation}, sometimes via remote-controlled robots \cite{remoteControl} that are amenable to automated approaches.

The steady-state planning problem has been studied under various names, including steady-state control \cite{SSC}, average- or expected-reward constrained MDPs \cite{altman1999constrained}, and steady-state policy synthesis \cite{IJCAI2019}. As pointed out by \citeA{Altman2019_11}, solutions to this problem often require strong assumptions on the ergodicity of the underlying MDP. These assumptions facilitate the search for efficient algorithms by leveraging the one-to-one correspondence between the optimality of solutions to various mathematical programs and the optimality of policies derived thereof. This has been studied at length in the works of \citeA{Derman:1970}, \citeA{kallenberg1983linear}, \citeA{puterman1994markov}, and \citeA{altman1999constrained}, who have derived mathematical programs for discounted, total, and expected reward formulations of constrained MDPs. In particular, the work of Kallenberg laid the foundation for Markovian control within the context of multichain constrained MDPs. However, it was noted that deriving optimal policies for the expected-reward formulation was intractable by their approach and there was no guarantee of agent behavior in terms of satisfying steady-state constraints.

\paragraph{Summary of contributions.} 
We make four main contributions. {\color{black} First, we introduce the \ac{name} problem of finding a policy from a predefined subset of stationary policies in a multichain MDP that maximizes an expected reward signal while enforcing asymptotic behavior that is correct-by-construction \cite{correctByConstruction} -- in the sense that our policies yield provably correct behavior that satisfies the imposed specifications on the steady-state distribution of the Markov chain induced by said policies.} Our framework generalizes the steady-state planning problems studied by \citeA{SSC} and \citeA{IJCAI2019}, as we do not impose any restrictions on the underlying MDP. 
In particular, we dispense with
the strong assumption made by \citeA{SSC} about the ergodicity of the MDP, according to which every deterministic policy necessarily induces an ergodic Markov chain (i.e., one that is recurrent and aperiodic). In sharp contrast to the work of \citeA{IJCAI2019}, we do not restrict our search to stochastic policies that induce an irreducible Markov chain (i.e., one in which all states form one communicating class).
In general, such a chain may not even exist -- normally, many states in a given MDP are inevitably transient. {\color{black} Our search space consists of subsets of the stationary policies that we term edge- or class-preserving, which, apart from a transient phase, restrict the long-term play in the terminal components of the
given MDP.} 
%
We introduce two distinct notions for class preservation that yield policies with different characteristics. These notions will be made precise in Section \ref{sec:MSSPS}.

As our second contribution, we develop a scalable approach to synthesize policies that provably meet said asymptotic specifications through novel linear programming formulations. While a tractable solution to the \ac{name} problem has heretofore remained elusive and existing solutions require an enormous amount of calculations with no provable guarantees \cite{kallenberg1983linear},
two key ideas underlie our ability to tackle the associated combinatorial difficulties. \textcolor{black}{The first idea is the aforementioned restriction of the domain to edge- or class-preserving policies, which can be provably obtained from solutions to simple \ac{LPs}.}
The second idea is to encode constraints on the limiting distributions of the corresponding Markov chains in formulated LPs, whose solutions yield optimal policies maximizing the expected average reward while meeting desired asymptotic specifications on the limit points of the expected state-action frequencies. These LPs are crafted to capture designated state classifications, absorption probabilities in closed communicating components, and recurrence constraints within such components, along with the steady-state specifications. 


Our third contribution lies in deriving key theoretical results establishing provable performance and behavior guarantees for the derived policies. 
Contracting or transient MDP models that use the expected total reward as the optimality criterion are commonplace in constrained MDPs since optimal stationary policies with regard to this criterion can always be found via mathematical programming in view of a well-established one-to-one correspondence between stationary policies and feasible solutions to such programs \cite{altman_total_cost_98,fein_or_2000,wu_durfee_jair_2010,petrik2009bilinear}. The notoriously more difficult and equally important expected average reward criterion is much less understood considering that such correspondence ceases to exist for general multichain MDPs.  
%
In this paper, we 
tap into this long-standing dilemma and establish such one-to-one correspondence for classes of stationary policies that are edge- or class-preserving.
Theorems \ref{thm:pi_in_ext_edge_set}, \ref{thm:mainEP}, \ref{thm:pol_in_CP} and \ref{thm:mainCP} establish the correctness of linear programs yielding optimal policies from said classes. The proof of these theorems rest on few intermediate results. 
In particular, Lemma \ref{thm:class_under_pi} characterizes the Markov chains induced by the policies of interest, while Lemma \ref{thm:XPi_in_LP} establishes the feasibility of the steady-state distributions induced by these policies. Lemma \ref{thm:mainCPU} gives a sufficient condition for the existence of a one-to-one correspondence between feasible solutions to the linear programs and the stationary policies derived from these solutions.
Theorem \ref{thm:CPU} establishes an existence condition of policies found on a more relaxed notion of class preservation, which inspires a constructive approach in Algorithm \ref{alg:CPU} to compute such policies. Theorem \ref{thm:dual_cone} gives a generic sufficient condition for the existence of an optimal stationary policy meeting the desired specifications beyond class-preserving ones. 

As our fourth contribution, we introduce an alternative type of  specifications applicable in transient states. By augmenting our \ac{LPs} with appropriate constraints, the synthesized policies provably meet specifications on the expected number of visitations to transient states simultaneously with the foregoing steady-state specifications on the asymptotic frequency with which recurrent states are visited (Proposition \ref{thm:unichainpol}). 

We verify the theoretical findings of our work using a comprehensive set of numerical experiments performed in various environments. The results demonstrate the correctness of the proposed \ac{LPs} in yielding policies with provably correct behavior and the scalability of the proposed solutions to large problem sizes. 

This article brings in and substantially extends the scope of our recent work \shortcite{ijcai2020}, which considered policy synthesis over edge-preserving policies. Such policies constitute only a small subset of the policies considered herein. A particularly appealing characteristic of the newly introduced policies is their greater ability to avert MDP transitions of low return without violating the asymptotic constraints.
In turn, they yield larger expected rewards relative to their edge-preserving counterparts -- 
in some cases, we show that this gain can be substantial. Further, we derive general characterizations of optimality over a larger class of policies obtained in terms of the MDP reward signal. In addition, this article advances the aforementioned form of transient specifications that a policy can provably meet together with the steady-state ones. \textcolor{black}{We provide a complete presentation of the steady-state planning problem through 
linear programming formulations over different families of policies, mathematical analyses establishing correctness of such formulations with optimality guarantees, and a comprehensive set of numerical experiments in diverse environments to support the theoretical findings. }


To the best of our knowledge, this work is the first to allow synthesis of stationary policies with provably correct steady-state behavior in general multichain MDPs.
\smallbreak
\noindent\textbf{Organization:} The paper is organized as follows. Notation and preliminaries are covered in Section \ref{sec:bkgnd}. Related work in steady-state planning is summarized in Section \ref{sec:relwork}. The \ac{name} problem is formalized in Section \ref{sec:MSSPS}. We describe our linear programming approach and present the results of our theoretical analysis in Section \ref{sec:lps}. 
Transient specifications and extensions to a larger class of policies are presented in Section \ref{sec:extensions}. Numerical experiments are presented in Section \ref{sec:expresults} to validate our approach and demonstrate its scalability to large problems. Concluding remarks are presented in Section \ref{sec:conclusions}. \textcolor{black}{In \ref{sec:appendx_proofaux}, we present statements and proof of technical lemmas. The proof of the main results are deferred to \ref{sec:appendx_proofmain}.}

\section{Preliminaries and Notation}
\label{sec:bkgnd}
We introduce some notation and preliminary definitions used throughout the paper. For a matrix $A$, $a_{ij}$ and $A(i,j)$ are used interchangeably to denote the element in its $i\textsuperscript{th}$ row and $j\textsuperscript{th}$ column.
%
%
The vectors $e$ and $e_s$ denote the vectors (of appropriate dimension) of all ones, and all zeros except for the $s^\textrm{th}$ entry, respectively. 
Given a vector $x$ and index set $V$, the vector $x_V$ is the vector with entries $x_v, v\in V$, where $x_v$ is the entry corresponding to index $v$. 
By $|S|$, we denote the cardinality of a set $S$. For an integer $n> 0$, the set $[n]: = \{1,\ldots, n\}$, and $A\setminus B$ denotes the set difference of sets $A$ and $B$. The symbols $\exists$ and $\exists!$ mean ``there exists" and ``there exists a unique'', respectively, and $^\top$ is the transpose operator.

\begin{definition}[Markov chain]
A Markov chain is a stochastic model given by a tuple $\calM = (S,T,\beta)$, where $S$ is the state space, $T$ the transition function $T: S\times S\rightarrow [0,1]$ with $T(s'|s)$ denoting the probability of transitioning from state $s$ to state $s'$, and $\beta: S \rightarrow [0, 1]$ the initial state distribution. With slight abuse of notation, the transition function can also be thought of as a matrix $T\in [0,1]^{|S|\times |S|}$, where $T(s,s') = T(s'|s)$. The use of $T$ will be clear from the context. 
\end{definition}

\begin{figure}
    \centering
   \includegraphics[width=0.9\columnwidth]{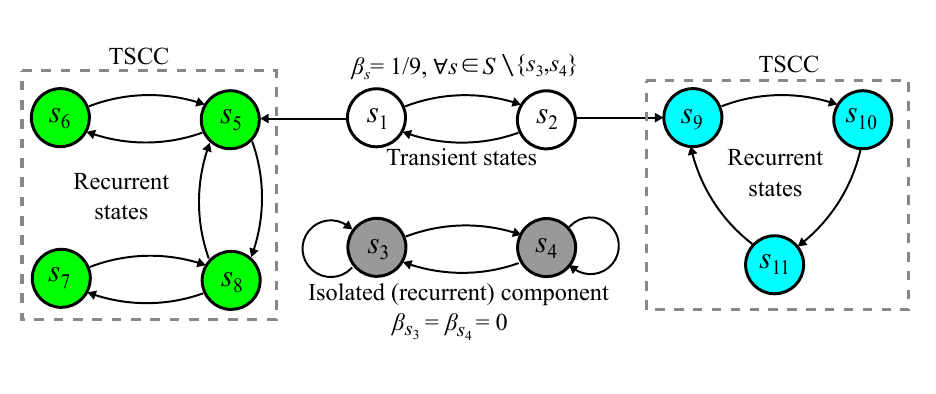}
    \vspace{-0.8cm}
    \caption{State classification in a Markov chain with four communicating classes. The set $\{s_1, s_2\}$ is transient, and the sets $\{s_3, s_4\}$, $\{s_5, s_6, s_7, s_8\}$ and $\{s_9, s_{10}, s_{11}\}$ are recurrent. This Markov chain is not irreducible since the states do not belong to one communicating class. States $s_9, s_{10}$, and $s_{11}$ are periodic. The set $\{s_3, s_4\}$ is isolated since it is not reachable from states in $S\setminus\{s_3, s_4\}$ and it has zero initial distribution. The components colored in green and blue are the two TSCCs of the Markov chain, i.e., $\tscc{1}{} = \{s_5, s_6, s_7, s_8\}$ and $\tscc{2}{} = \{s_9, s_{10}, s_{11}\}$.}  
    \label{fig:example_MC_states}
\end{figure}

\paragraph{Classification of states}\cite{norris_1997,Privault2018}:  {\color{black} Given a finite Markov chain $\calM = (S,T,\beta)$, we say state $s'$ is accessible from state $s$ if $(T^t)(s,s') > 0,$ for some $t > 0$, where $T^t$ is the $t-$step transition matrix, i.e., if there is a positive probability of transitioning to state $s'$ starting from state $s$ in some number of steps. Two states are said to communicate if they are both accessible from each other.}  
Communication is an equivalence relation which partitions the Markov chain $\cal M$ into communicating classes such that only members of the same class communicate with each other. A class is closed if the probability of escaping the class is zero. {\color{black} A state $s \in S$ is said to be \emph{transient} if, starting from $s$, there is a non-zero probability of never returning to $s$}. 
A set of transient states is termed a transient set. {\color{black} Non-transient states are called recurrent, that is, state $s$ is recurrent if, starting from $s$, the probability of returning to state $s$ after some number of steps is one. A Markov chain for which there is only one communicating class consisting of the entire state space is called irreducible, whereas a Markov chain that has a single closed communicating class and (possibly) some transient states is termed unichain. 
A state is periodic with period $k$ if any return to state $s$ must occur in multiples of $k$ time steps, where $k$ is some integer greater than $1$.} 
An example illustrating the classification of states in a Markov chain is shown in Figure \ref{fig:example_MC_states}.

Transience and recurrence describe the likelihood of returning to a state \emph{conditioned} on starting from that state, regardless of the initial state distribution $\beta$. Given $\beta$, we also define an \emph{isolated} component $I$ as a {\color{black} maximal} set of states in $\calM$ that can never be visited, that is, $\beta_I = \sum_{s \in I} \beta_s = 0$, where $\beta_s$ is the initial probability of being in state $s$, and $I$ cannot be reached from any state in $S\setminus I$, i.e., $\sum_{s'\in I} T(s'|s) = 0, \forall s\in S\setminus I$. In Figure \ref{fig:example_MC_states}, the set of states $\{s_3, s_4\}$ is isolated. The term `reachable' refers to states that are not isolated.  

\begin{definition}[Markov decision process (MDP)]
An MDP is a tuple $\calM = (S, A, T, R, \beta)$, in which $S$ denotes the state space, $A$ the set of actions, $T: S\times A\times S\rightarrow [0,1]$ the transition function with $T(s'|s,a)$ denoting the probability of transitioning from state $s$ to state $s'$ under action $a$, $R: S\times A\times S\rightarrow\bbR$ a reward obtained when action $a$ is taken in state $s$ and we end up in state $s'$, and $\beta: S \rightarrow [0, 1]$ the initial distribution. 
By $A(s)\subseteq A$, we denote the set of actions available in state $s$.
\end{definition}

{\color{black}
\begin{definition}[Transition graph]
We define the transition graph of an MDP \sloppy ${\cal M} = (S, A, T, R, \beta)$ as the directed graph whose vertex set is the state space $S$, and in which there is a directed edge from vertex $s$ to vertex $s'$ if there exists an action $a\in A(s)$ such that $T(s'|s,a) > 0$. The transition graph of a Markov chain ${\cal M} = (S, T, \beta)$ is the directed graph with vertex set $S$, and which has a directed edge from vertex $s$ to vertex $s'$ if $T(s'|s) > 0$.
\end{definition}
}

\begin{figure}
    \centering
    \includegraphics[width=\columnwidth]{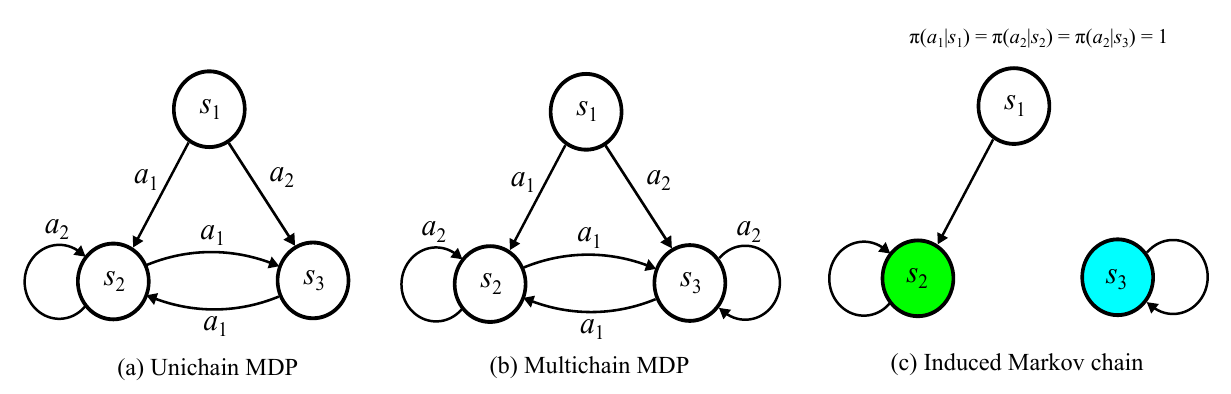}
    \caption{(a) Unichain MDP: every deterministic policy induces a Markov chain that has exactly one recurrent component. (b) Multichain MDP: adding the self-loop to state $s_3$ yields a multichain MDP. For example, the deterministic policy defined by $\pi(a_1|s_1) = \pi(a_2|s_2) = \pi(a_2|s_3) = 1$ induces the Markov chain in (c), which is multichain with two recurrent components $\{s_2\}$ and $\{s_3\}$.}
    \label{fig:uni_multi_chain}
\end{figure}

\begin{definition}[Terminal strongly connected component (TSCC)]
	Consider the {\color{black} transition graph} 
	of a Markov chain or MDP $\mathcal{M}$ with state space $S$ and initial distribution $\beta$. {\color{black} A strongly connected component (SCC) of the digraph is a maximal subset of vertices $C$, where for every pair of vertices $s,s'\in C$, there is a directed path\footnote{\textcolor{black}{There is a directed path from node $v$ to node $w$ if it is possible to reach $w$ from $v$ by traversing the directed edges in the directions in which they point.}} from $s$ to $s'$ and a directed path from $s'$ to $s$ \cite{Tarjan1972DepthFirstSA}.}
	A Terminal Strongly Connected Component (TSCC) $S' \subseteq S$ is an SCC reachable from some initial state $s, \beta_s > 0$ and with no outgoing transitions to any state in $S \setminus S'$. \textcolor{black}{A TSCC is also called a bottom SSC \cite{cour_yann_jacm_1995}}. We denote by $\tscc{k}{}\subseteq S$ the set of states in the $k^\text{th}$ TSCC of $\calM$, and by $\recurrentset{} = \bigcup_{k \in [m]} \tscc{k}{}$ the union of all such sets. 
	The complement set is denoted $\transientset{} := S \setminus \recurrentset{}$, which in the case of Markov chains is the set of transient or isolated states.
	\label{def:TSCC}
\end{definition}
Figure \ref{fig:example_MC_states} illustrates a Markov chain with two TSCCs (highlighted with two separate colors).
{\color{black}
\begin{definition}[Stationary policy]
Given MDP $\calM = (S, A, T, R, \beta)$, a stationary policy $\pi:S\rightarrow \Delta^A$ is a mapping of states to probability distributions over the space of actions $A$, where $\Delta^A$ is the probability simplex over $A$. 
The policy $\pi$ specifies the conditional probability $\pi(a|s)$ that action $a$ is taken in state $s$. 
The set of all stationary policies is denoted $\Pi_S$. 
\end{definition}}
%
\begin{definition}[Markov chain induced by policy] The tuple $\calM_{\pi} = (S, T_{\pi}, \beta)$ is the Markov chain induced by a policy $\pi$ 
in an underlying MDP $\calM = (S,A,T,R, \beta)$, where
\begin{align}
T_\pi(s'|s) = \sum_{s\in A(s)} T(s'|s,a) \pi(a|s)
\label{eq:M_pi}
\end{align}
\label{def:M_pi}
\end{definition}
\begin{definition}[Unichain and multichain MDP]
An MDP is called unichain \cite{puterman1994markov,altman1999constrained} if \textcolor{black}{every} stationary deterministic policy induces a Markov chain that is unichain, that is, consists of exactly one recurrent set and possibly some transient states\footnote{This definition does not require the recurrent class to be ergodic (hence aperiodic). Our analysis dispenses with the aperiodicity precondition as will be clear in the sequel.}. An MDP is said to be multichain if it is not unichain. See Figure \ref{fig:uni_multi_chain} and its caption for an example. 
\end{definition}
\begin{definition}[Stationary distribution] Given a Markov chain $\calM = (S,T, \beta)$, a stationary distribution $\ssdist: S\rightarrow [0,1]$ over the state space is any solution to the set of equations \cite{norris_1997}
\begin{align}
     \ssdist(s) = \sum_{s'\in S}\ssdist(s') T(s|s')&, ~ \ssdist(s)\geq 0, ~\forall s\in S  \label{eq:stationary_dist_1}\\
     \sum_{s\in S} \ssdist(s) &= 1\:.
     \label{eq:stationary_dist_2}
\end{align}
\end{definition}
\textcolor{black}{According to the ergodic theorem of Markov chains, the solution to \eqref{eq:stationary_dist_1} and \eqref{eq:stationary_dist_2} is unique if and only if $T$ is the transition matrix of a unichain \cite[Chapter 4]{Gallager_book}.} If there are multiple recurrent classes,
then in general there will be many stationary distributions. For example, for the Markov chain of Figure \ref{fig:uni_multi_chain}(c), one can verify that the distribution $\ssdist(s_1) = \ssdist(s_2) = 0, \ssdist(s_3) = 1$ and the distribution $\ssdist(s_1) = \ssdist(s_3) = 0, \ssdist(s_2) = 1$ both satisfy \eqref{eq:stationary_dist_1} and \eqref{eq:stationary_dist_2}, thus they are both stationary distributions of the Markov chain. Note that a stationary distribution may not be representative of the true steady-state behavior of the
system (c.f. Definition \ref{def:ss_dist} and the following example).
\begin{definition}[Stationary matrix of a Markov chain] Given a Markov chain $\calM = (S,T, \beta)$, the stationary matrix $T^\infty$ is given by the Ces\`aro limit\footnote{The Ces\`aro limit always exists and accounts for the non-convergence of
powers of transition matrices of periodic chains. Hence, we do not need a precondition about aperiodicity as our analysis does not require that $T^\infty = \lim_{n\rightarrow\infty} T^n$.} \cite{puterman1994markov} 
\begin{align}
\label{eq:def_Tinfty}
\cesarolimit{T} = \lim_{n \rightarrow \infty} \frac{1}{n} \sum_{t=1}^n T^t \:.
\end{align}
\end{definition}
%
%
Given a finite multichain Markov chain $\calM = (S,T, \beta)$ with transient set $F$ and recurrent (i.e., non-transient) components $E_k, k\in [\NumErgSets]$, the transition matrix $T$ can be expressed in the canonical form \textcolor{black}{\cite[Appendix A]{puterman1994markov}}
\begin{align}
T = \begin{bmatrix}
    T_1 & 0 & \ldots & 0  & 0 \\
    0 & T_2 &  & \dots  & 0 \\
    \vdots & \vdots & \vdots & \ddots & \vdots \\
    0 & 0 & \ldots & T_m &  0 \\
    L_1 & L_2 & \ldots & L_m & Z
\end{bmatrix}
\label{eq:T_form}
\end{align}
where the matrices $T_k$ correspond to transitions between states in $E_k$, $L_k$ to transitions from states in $F$ to states in $E_k, k\in[\NumErgSets]$, and $Z$ to transitions between states in $F$. Similarly, we use $T_{\pi,k}, L_{\pi,k},\: k = 1,\ldots,\NumErgSets$, and $Z_\pi$ to denote the corresponding submatrices of the transition matrix $T_\pi$ of the Markov chain $\calM_\pi$ induced by policy $\pi$. Also \cite{puterman1994markov,kallenberg1983linear},  
\begin{align}
    T^\infty(s',s) = \left\{
     \begin{array}{@{}l@{\thinspace}l}\eta_s, & ~~~~~~ s',s\in E_k~\text{for some } k\in[\NumErgSets] \\
     p_{s'k}\eta_s & ~~~~~~ s'\in F, s\in E_k\\
     0 & ~~~~~~ \textit{otherwise}
     \end{array}
     \right.
\label{eq:T_star_form}
\end{align}
where,
$\eta_s = \lim_{n \rightarrow \infty} \frac{1}{n} \sum_{t=1}^n T^t(s',s)$, is the long term proportion of time the chain spends in $s$ from initial states $s'\in E_k$,
$\sum_{s\in E_k}\eta_s = 1$, $p_{s'k}$ is the absorption probability from the transient state $s'\in F$ into the recurrent class $E_k, k\in[m]$, and $\sum_{k=1}^\NumErgSets p_{s'k} = 1, \forall s'\in F$.

In this paper, we are interested in the asymptotic behavior of an agent's policy in an MDP, as captured by the steady-state distribution of the induced Markov chain defined next.  
\begin{definition}[Steady-state distribution]
Given an MDP $\calM$ and policy $\pi$, the steady-state distribution $\prbetapi:S\times A\rightarrow [0,1]$ over the state-action pairs, also known as the occupation measure \textcolor{black}{\cite[Chapter 4]{altman1999constrained}}, is the long-term proportion of time spent in state-action pair $(s,a)$ as the number of transitions approaches $\infty$, i.e.,
\begin{align}
\begin{aligned}
    \prbetapi(s,a)
    &= \lim_{n\rightarrow\infty} \frac{1}{n}\sum_{t=1}^n\mathrm{Pr}(S_t = s, A_t = a | \beta, \pi), ~~~s\in S, a\in A(s)
    \label{eq:occ_measure}
\end{aligned}
\end{align}
if the limit exists, where $S_t$ and $A_t$ are the state and action at time $t$. Also, $\prbetapi(s):= \sum_{a\in A(s)}\prbetapi(s,a)$ is the steady-state probability of being in state $s\in S$. The steady-state distribution is a stationary distribution of the Markov chain induced by the policy $\pi$. 
\label{def:ss_dist}
\end{definition}
As an example, consider the MDP in Figure \ref{fig:uni_multi_chain}(b) with  $\beta_{s_1} = 1, \beta_{s_2} = \beta_{s_3} = 0$. The steady-state distribution of the policy $\pi(a_1|s_1) = \pi(a_2|s_2) = \pi(a_2|s_3) = 1$, which induces the Markov chain in Figure \ref{fig:uni_multi_chain}(c), has $\prbetapi(s_2,a_2) = 1$ and $0$ otherwise.
\textcolor{black}{
\begin{definition}
\label{def:admis_pol_measures}
Given an MDP $\calM = (S,A,T,R,\beta)$ and a set of policies $\Pi\subseteq\Pi_S$, we define 
\[\setssdistbeta(\Pi)
:= \{\prbetapi | \pi\in\Pi\}
\] 
as the set of occupation measures induced by policies in $\Pi$, where $\prbetapi$ is defined in \eqref{eq:occ_measure}. 
\end{definition}
}

\begin{definition}[Steady-state specifications and constraints \cite{IJCAI2019}]
	Given an MDP $\calM = (S, A, T, R, \beta)$ and a set of labels $L = \{L_1, \ldots, L_{n_L}\}$, where $L_i\subseteq S$, a set of steady-state specifications is given by $\SSSpec = \{(L_i, [l_i, u_i])\}_{i = 1}^{n_L}$. Given a policy $\pi$, the specification $(L_i, [l_i, u_i]) \in \SSSpec$ is satisfied if and only if the steady-state constraint 
	{\color{black}
	\begin{align}
	l_i \leq \sum_{s\in L_i} \prbetapi(s) \leq u_i
	\label{eq:ss_spec}
	\end{align}
	is satisfied; that is, if the steady-state probability of being in a state $s \in L_i$ in the Markov chain $\calM_\pi$ falls within the interval $[l_i, u_i]$. }
	\label{def:spec}
\end{definition}


\begin{definition}[Labeled MDP \cite{IJCAI2019}]
	An MDP $\calM = (S,A,T,R,\beta,L,\SSSpec)$ augmented with the label set $L$ and specifications $\SSSpec$ is termed a labeled MDP (LMDP). 
	\label{def:LMDP}
\end{definition}


\begin{lemma}
\textcolor{black}{\cite[Theorem 4.3.2]{kallenberg1983linear}}\cite{10.2307/3690451}
\label{lem:ssdist}
Given an MDP $\calM = (S,A,T,R,\beta)$ and policy $\pi\in\stationaryset$, 
the steady-state distribution $\prbetapi:=\{\prbetapi(s,a)\}_{s,a}$ of the Markov chain $\calM_\pi$ is
\begin{align}
    \prbetapi(s,a) = (\beta^\top \cesarolimit{T_\pi})_s\pi(a|s), ~s\in \Allstates, a\in A(s)
    \label{eq:betaPstar}
\end{align}
where $\cesarolimit{T_\pi}$ is the Ces\`aro limit in \eqref{eq:def_Tinfty}, i.e., $\cesarolimit{T_\pi} = \lim_{n \rightarrow \infty} \frac{1}{n} \sum_{t=1}^n T_\pi^t $. 
\end{lemma}

\begin{definition}[Expected average reward]
Given an MDP $\calM = (S,A,T,R,\beta)$, the expected average reward $R_\pi^\infty(\beta)$ of a policy $\pi$ 
is defined as
\begin{align}
R_\pi^\infty(\beta) = \lim_{\hspace*{14pt}n\rightarrow\infty} \!\!\!\!\! \inf \frac{1}{n} \sum_{t=1}^n \mathbb{E}_{\substack{A_t \sim \pi \\ S_0 \sim \beta}}[R(S_t,A_t)]
\label{eq:avg_reward}
\end{align}
{\color{black} where $R(s,a) := \sum_{s'\in S} T(s'|s,a)R(s,a,s')$}, and the expectation is w.r.t. the probability measure induced by the initial distribution $\beta$ and the policy $\pi$ over the state-action trajectories. 
\end{definition}
It follows from the definition of the expected average reward in \eqref{eq:avg_reward} and the steady-state distribution \eqref{eq:occ_measure} that for a stationary policy $\pi$ \cite{10.2307/3690451,altman1999constrained}
\begin{align}
R_\pi^\infty(\beta) = \sum_{s \in S} \sum_{a \in A(s)} \prbetapi(s,a) R(s,a)  
\label{eq:avgreward_equiv}
\end{align}
where $\prbetapi(s,a)$ is given in \eqref{eq:betaPstar}.

The primary focus of this paper in the context of steady-state planning is to find stationary policies that maximize the expected average reward \eqref{eq:avg_reward} while satisfying specifications $\SSSpec$ on the steady-state distribution (see Definition \ref{def:spec}). We restrict the search to certain classes of stationary policies which will be introduced and defined precisely in Section \ref{sec:MSSPS}. {\color{black} Our solution approach to this constrained MDP problem is based on linear programming formulations, which optimize a linear objective function capturing the expected reward, subject to linear equality and inequality constraints. Such constraints encode restrictions on the steady-state distributions induced by policies of interest, as well as desired steady-state specifications on the long-term frequencies for state-actions pairs. The decision variables of the \ac{LPs} correspond to the occupation
measures, and policies are obtained from their optimal solutions. We establish a one-to-one correspondence between optimal solutions of said \ac{LPs} and optimal policies of the constrained MDP problem. 
}  

\section{Related Work}
\label{sec:relwork}
Research related to steady-state planning often comes from the field of average- or expected-reward constrained MDPs and has its roots in mathematical programming \cite{bertsekas2005dynamic}.
Many solutions proposed in this area utilize linear programming formulations to derive policies \cite{altman1999constrained}.
We illustrate these formulations in order of increasing complexity and elucidate the key differences between the formulations in the literature and our own. First, let us consider the simple problem of deriving a policy for an agent which seeks to maximize expected reward without any constraints on its steady-state distribution.

In the unichain MDP case, one may synthesize a policy by solving a \ac{LP} of the form \cite{10.2307/2627340,DeGhellinck1960}
\begin{equation}
\begin{aligned}
    \max  \quad  &\sum_{s \in S} \sum_{a \in A(s)} x_{sa} \sum_{s' \in S} T(s' | s, a) R(s,a,s')
    \text{   subject to } \\
          \quad  &\sum_{s \in S} \sum_{a \in A(s)} x_{sa} T(s' \mid s,a) 
                    = \sum_{a \in A(s')} x_{s'a} & \forall s' \in S \\
           \quad &x_{sa} \in [0,1]
                  & \forall s \in S, a \in A(s) \\
           \quad &\sum_{s \in S} \sum_{a \in A(s)} x_{sa} = 1\:.
\end{aligned}
 \label{eqn:LPunichain}
\end{equation}
The policy can be derived from the occupation measures given by $x_{sa}$ through a simple calculation. It is worth noting that this combination of occupation measures and linear programming has enabled significant progress in the area of planning within stochastic shortest paths MDPs, where several occupation measure heuristics have been defined to find decision-making policies that maximize the probability of reaching a set of goal states while satisfying multiple cost constraints \cite{iDual,iDual2,occupationMeasurePlanning,occupationPLTL}. While the \ac{LP} in (\ref{eqn:LPunichain}) always produces valid solutions for unichain MDPs, this is not necessarily the case for multichain MDPs due to the fact that there may be more than one ergodic set \cite{puterman1994markov}.
This issue is rectified by modifying \ac{LP} \eqref{eqn:LPunichain} to obtain \cite{10.2307/2099444,kallenberg1983linear}
\begin{equation}
\begin{aligned}
    \max  \quad  &\sum_{s \in S} \sum_{a \in A(s)} x_{sa} \sum_{s' \in S} T(s' | s, a) R(s,a,s')
    \text{ subject to } 
    \\
    \quad &\sum_{s \in S} \sum_{a \in A(s)} x_{sa} T(s' \mid s,a) 
                    = \sum_{a \in A(s')} x_{s'a} & \forall s' \in S 
    \\
          \quad &\sum_{s \in S} \sum_{a \in A(s)} y_{sa}  T(s' \mid s,a) 
                 = \sum_{a \in A(s')} (x_{s'a} + y_{s'a}) - \beta_{s'}
                & \forall s' \in S 
    \\
           \quad &x_{sa} \in [0,1],~ y_{sa} \ge 0 &  \forall s \in S, a \in A(s).
\end{aligned}
\label{eqn:LPmultichain}
\end{equation}
The new $y_{sa}$ variables guide policy formation on the transient states.
Both \ac{LP} \eqref{eqn:LPunichain} and \ac{LP} \eqref{eqn:LPmultichain} yield stationary stochastic policies.
Furthermore, there always exists at least one optimal deterministic policy, which can easily be derived from the stochastic policy solution obtained from the LPs \cite{puterman1994markov}.

For producing control policies \emph{with} steady-state specifications, LPs \eqref{eqn:LPunichain} and \eqref{eqn:LPmultichain} are extended to include linear steady-state constraints on the occupation measures.
When applied to \emph{unichain} MDPs, the constrained version of \ac{LP} \eqref{eqn:LPunichain} encounters minor difficulties, in that there may not be an optimal deterministic policy \cite{altman1999constrained}.
Nonetheless, the \ac{LP} always produces an optimal stochastic stationary policy.
In fact, there exists an optimal policy having at most $n_L$ ``randomizations'', i.e. having at most $|S|+n_L$ state-action pairs with non-zero probability of being selected \cite{10.2307/171066}.
%

On the other hand, serious issues arise when \ac{LP} \eqref{eqn:LPmultichain} is augmented with steady-state constraints and solved for multichain MDPs, as described in the pioneering work of \citeA{kallenberg1983linear}.
In particular, it was shown that there is not a one-to-one correspondence between the feasible solutions of the augmented \ac{LP} and the stationary policies. Instead, the space of feasible solutions is partitioned into equivalence classes of various feasible solutions mapping to the same policy.
The key deficiency is that the steady-state distribution of the Markov chain induced by the synthesized policy does not match the optimal solution to the \ac{LP} in general, and so the derived policy does not always meet the steady-state specifications (see Example \ref{ex:Motivation} in Section \ref{sec:limitations}).
This issue is not easily remedied, since the optimal solution may not be achievable by any stationary policy, or identifying such a policy would generally require combinatorial search. We refer the reader to the paper by \citeA{10.2307/3690451} for an overview. 

In order to mitigate the preceding problem of integrating steady-state constraints, various assumptions have been made in the literature on the structure of the underlying MDP. Multichain MDPs are also frequently excluded from the conversation altogether.
The assumption that the MDP is ergodic, and therefore every policy induces an ergodic Markov chain, has been used by \citeA{SSC} to ensure that steady-state equations and constraints on the same are satisfied. This assumption is relaxed to some extent by \citeA{10.2307/171066,altman1999constrained,4927531}, where unichain MDPs are allowed.
The assumption of either an ergodic or a unichain MDP requires that no stationary deterministic policy induce more than a single recurrent class, thus severely limiting the applicability of these methods.
These assumptions are removed in the recent work of \citeA{IJCAI2019}, where neither ergodic nor recurrence assumptions are made on the underlying MDP.
However, the solution proposed therein finds an irreducible Markov chain 
in the underlying MDP, if one exists, and is therefore suitable for communicating MDPs where, for any two states $s$ and $s'$, there exists a deterministic stationary policy such that $s$ can reach $s'$ in a finite number of steps \cite{puterman1994markov}. This solution, however, is too restrictive, thus not suitable for reasoning over general multichain MDPs.

Another approach taken to address these challenges is to simply allow solutions to take the form of non-stationary policies.
In the work of \citeA{kallenberg1983linear}, this is accomplished by a computationally expensive approach producing a potentially different policy in each time step.
Another approach, proposed by \citeA{10.2307/3690451}, starts by using one policy, and then switches to a second ``tail'' stationary policy.
The time at which the switch occurs is determined by a lottery performed at each time step, and once the switch occurs the tail policy continues to be used indefinitely (thus the policy is ``ultimately'' stationary \emph{once} the switch occurs). However, this approach has three key limitations. First, the constraints must take the form of a target frequency vector, which imposes an equality constraint on the steady-state distribution over all states.
Second, the lottery system does not guarantee that the switch will occur in a finite number of steps, thus meaning that the policy is not guaranteed to be ultimately stationary. Third, the policy depends on a marker to track whether or not the switch has occurred.
This marker is not part of the MDP, and therefore the MDP machinery must be modified to include a so-called marker-augmented history. As an alternative, the authors also propose a way to extend the given MDP with additional states, such that the problem can be solved using a stationary policy applied to the extended MDP. However, this approach still cannot produce a stationary policy to solve the original problem.


While most methods for solving constrained MDPs revolve around the use of mathematical programs, some reinforcement learning approaches have also been proposed for optimizing the average-reward objective and, to a lesser extent, for solving constrained instances of average-reward MDPs. Some noteworthy examples include the constrained actor-critic method proposed by \citeA{ConstrainedActorCritic_12}, wherein a Lagrangian relaxation of the problem is used to incorporate steady-state costs into the objective function being optimized by the constrained actor-critic algorithm.  A similar Lagrangian Q-learning approach is proposed by \citeA{QLearningConstrained_13}.
Both of these reinforcement learning methods assume that every Markov chain induced by a policy is irreducible, which allows only a single recurrent class as with ergodic and unichain assumptions described earlier.
The Lagrangian approach has also been applied to specific stochastic policy linear programming formulations relevant to aircraft maintenance problems where the asymptotic failure is to be kept below some small threshold \cite{minFailureRate_7,minFailureRate_8}.

In contrast to the foregoing efforts, our approach is computationally tractable, works with the most general multichain MDPs, and always produces a stationary policy that satisfies the given steady-state specifications, if one exists.
Additionally, none of the aforementioned methods consider constraints on the expected visits to transient states, as we are considering in our work.

\section{\textcolor{black}{Steady-State Policy Synthesis: Problem Formulation}}
\label{sec:MSSPS}
In this section, we introduce the Steady-State Policy Synthesis (SSPS) problem of finding a stationary policy from predefined classes of policies (edge- and class-preserving) that maximizes the expected average reward subject to steady-state specifications. {\color{black} In contrast to prior work, we do not impose restrictions on the underlying MDP. Before we present our formulation, we briefly discuss the challenges underlying policy synthesis under the average reward optimality criterion and demonstrate the limitations of existing formulations in this context. Subsequently, we specify our search domain of policies and define the \ac{name} problem of synthesizing optimal policies from this domain.} 

\subsection{Challenges and Limitations}
\label{sec:limitations}
\begin{figure}
    \centering
    \includegraphics[width=.8\columnwidth]{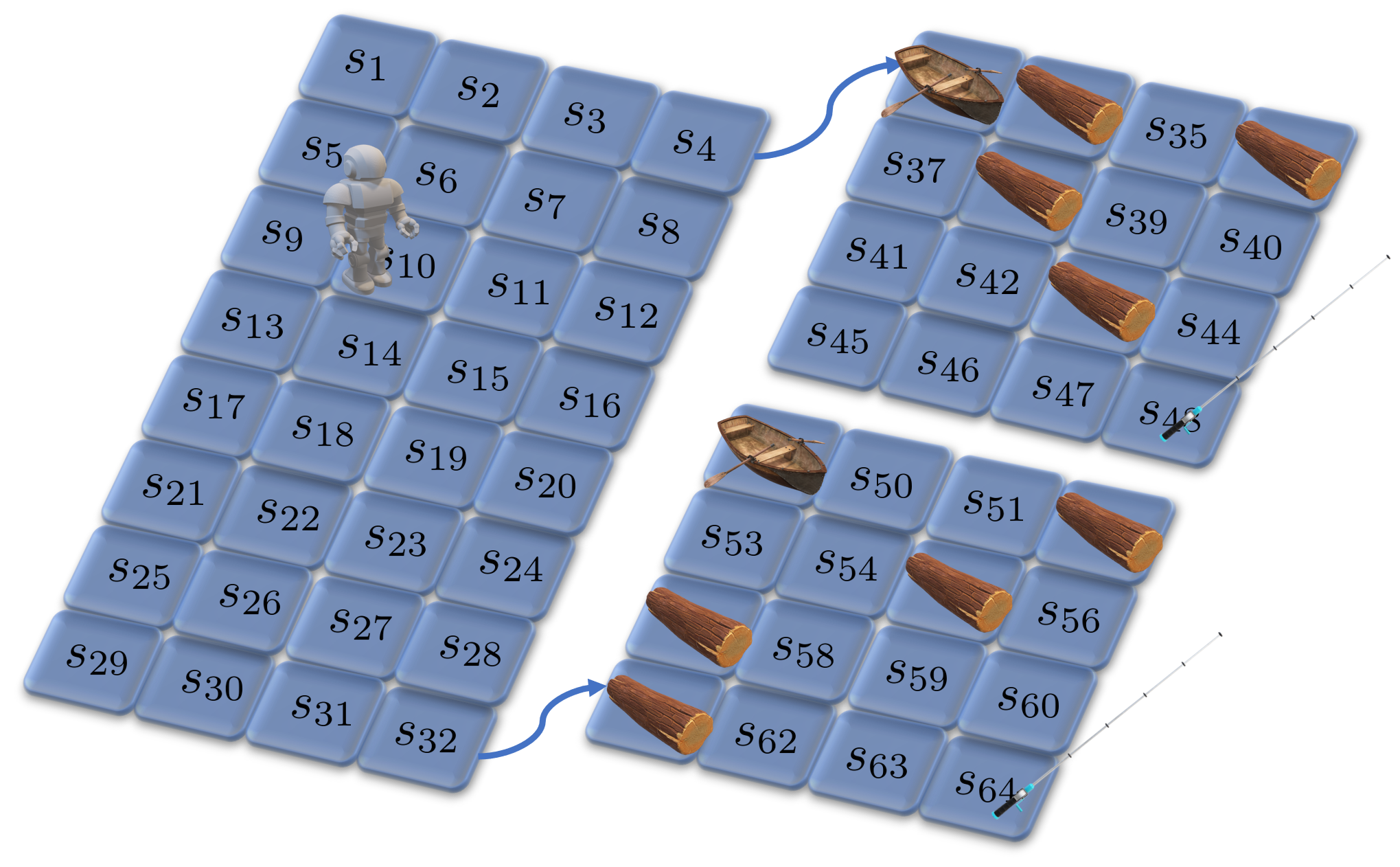}
    \caption{LMDP $\calM = (S,A,T,R,\beta,L,\SSSpec)$ with labels $L_\textrm{log1} = \{s_{34}, s_{36}, s_{38}, s_{43}\}$, $L_\textrm{log2} = \{s_{52}, s_{55}, s_{57}, s_{61}\}$, $L_\textrm{canoe1} = \{s_{33}\}$, $L_\textrm{canoe2} = \{s_{49}\}$, $L_\textrm{fish1} = \{s_{48}\}$, $L_\textrm{fish2} = \{s_{64}\}$, steady-state specifications $(L_\textrm{log1}, [0.25, 1])$, $(L_\textrm{log2}, [0.25, 1])$, $(L_\textrm{canoe1}, [0.05, 1.0])$, $(L_\textrm{canoe2}, [0.05, 1.0])$, $(L_\textrm{fish1}, [0.1, 1.0])$, $(L_\textrm{fish2}, [0.1, 1.0]) \in \SSSpec$, and rewards $R(\cdot, \cdot, L_\textrm{fish1}) = R(\cdot, \cdot, L_\textrm{fish2}) = 1$, $R(\cdot, \cdot, S \setminus (L_\text{fish1} \cup L_\text{fish2})) = 0$.}
    \label{fig:example}
\end{figure}

We motivate this section with a simple example. Suppose an autonomous agent is marooned on a set of three connected frozen islands as shown in Figure \ref{fig:example}.
The agent's goal is to maximize the amount of time it spends fishing for sustenance while at the same time building a canoe to escape the islands. The agent has an equal chance of starting in any state belonging to the larger island of size $n \times n / 2$, i.e., we have $\beta_s = 2 / n^2$ for each state $s$ in the island.
Once the agent moves to one of the two smaller islands, it is unable to return to the larger island. One quarter of the land in the small islands contains logs which can be used to build a canoe, and each of these islands contains one fishing site as well. For the first small island we have steady-state specifications $(L_\textrm{log1}, [0.25, 1])$, $(L_\textrm{log2}, [0.25, 1])$, $(L_\textrm{canoe1}, [0.05, 1.0])$ and reward $R(\cdot, \cdot, L_\textrm{fish1}) = R(\cdot, \cdot, L_\textrm{fish2}) = 1$. Likewise, the second small island has steady-state specifications $(L_\textrm{canoe2}, [0.05, 1.0])$, $(L_\textrm{fish1}, [0.1, 1.0])$, $(L_\textrm{fish2}, [0.1, 1.0])$ and reward $R(\cdot, \cdot, S \setminus (L_\text{fish1} \cup L_\text{fish2})) = 0$. Because the islands are covered in ice, the agent has a chance of slipping in three possible directions whenever it moves.
Specifically, if the agent attempts to go right (left), it has a 90\% chance of transitioning to the right (left), and there is a 5\% chance of transitioning instead to either of the states above or below it. Similarly, if the agent tries to go up (down), it moves to the states above (below) it with 90\% chance, and to the states to the right and left of it with chance 5\% each. This Frozen Island scenario is motivated by that found in OpenAI Gym's FrozenLake environment \shortcite{OpenAIGym}.

For this example, the \ac{LP} by \citeA{IJCAI2019} is infeasible since there exists no policy that induces an irreducible Markov chain, that is, one where all states in $S$ belong to one recurrent class. 
\textcolor{black}{The \ac{LP} by \citeA{kallenberg1983linear} in \eqref{eqn:LPmultichain} will return a solution $(x,y)$, from which the stationary policy $\pi:=\pi(x,y)$ is computed as follows  
\begin{align}
	\pi(a|s) = \left\{
	\begin{array}{@{}l@{\thinspace}l}
	\frac{x_{sa}}{x_s}  & ~~~~~~~ s\in E_x, a\in A(s)\\
	\frac{y_{sa}}{y_s} & ~~~~~~~ s\in E_y \setminus E_x,  a\in A(s) \\
	\text{arbitrary} & ~~~~~~~\text{otherwise}\\
	\end{array}
	\right.
	\label{eq:policy_LP1}
	\end{align}
	where $x_s := \sum_{a\in A(s)}x_{sa}$, $y_s = \sum_{a\in A(s)}y_{sa}$, $E_x := \{s\in \Allstates: x_s > 0\}$ and $E_y := \{s\in \Allstates: y_s > 0\}$.}
However, in general the steady-state distribution induced by the policy \eqref{eq:policy_LP1} will not satisfy the specified constraints. This deficiency is best demonstrated via a simple example. {\color{black} The reader is also referred to Example 1 of \citeA{10.2307/3690451}. 

\begin{example}\label{ex:Motivation}
Consider the MDP in Figure~\ref{fig:uni_multi_chain}(b) with initial probability $\beta_{s_1} = \beta_{s_3} = 0, \beta_{s_2} = 1$. One feasible solution $x$ of the \ac{LP} in \eqref{eqn:LPmultichain}  \cite[Program 4.7.6]{kallenberg1983linear} has $x_{s_2 a_2} = x_{s_3 a_2} = 0.5$. The policy $\pi$ in \eqref{eq:policy_LP1} corresponding to $x$ has $\pi(a_2|s_2) = \pi(a_2|s_3) =1$, hence $\prbetapi(s_2,a_2) = 1$. Therefore, $\prbetapi\ne x$. 
\end{example}
}
\textcolor{black}{The previous example underscores the main challenge underlying steady-state planning in constrained Markov decision models with the average reward criterion: solutions to formulated programs and stationary policies are not in one-to-one correspondence.} 
\textcolor{black}{In other words,
given a feasible LP solution $(x,y)$, 
the steady-state distribution $\prbetapi$ induced by the policy $\pi(x,y)$ derived from that solution is not equal to $x$ in general. 
As a result, unlike unichain MDPs \cite{altman1999constrained}, steady-state specifications encoded as constraints on the state-action variables are generally not met by $\pi$.} 
%
\begin{figure}
    \centering
    \includegraphics[width=0.8\columnwidth]{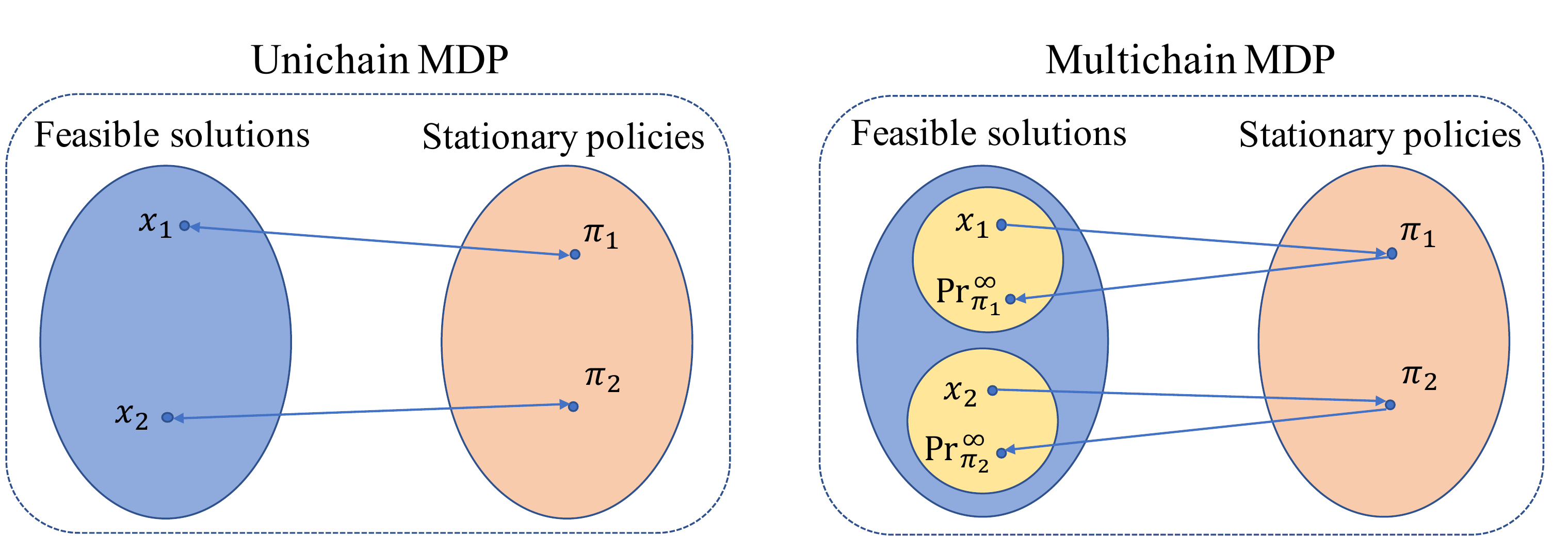}
    \caption{(Left) One-to-one correspondence between the feasible LP solutions and the stationary policies in unichain MDPs. (Right) Equivalence classes of feasible solutions map to stationary policies. 
    The steady-state distribution of the Markov chain induced by a policy need not agree with the LP solution, and could hence fail to meet the LP constraints.}
    \label{fig:example_correspondence}
\end{figure}
Figure 5 (Left) illustrates the one-to-one correspondence between LP solutions and stationary policies found in unichain MDPs.
Figure 5 (Right) illustrates the lack of such correspondence in 
multichain MDPs, where
instead, equivalence classes of feasible LP solutions (yellow circles) map to the same policy 
\cite{kallenberg1983linear,puterman1994markov}.
%

\subsection{Problem Setup}
\label{sec:formulation}
The previous example motivates the work of this paper in which we develop an approach to synthesizing policies with provably correct asymptotic behavior based on the notions of edge preservation and class equivalence. First, we will define sets of policies under which certain class structures are preserved and give an example of such policies, then define the \ac{name} problem of finding an optimal policy from such classes.

\begin{definition}[Edge-preserving policies]
\label{def:edgepreservingset}
Given an MDP $\calM$, we define the set of Edge-Preserving (EP) policies $\edgepreservingset$ as the set of stationary policies that {\color{black} play every action available at states} 
in the TSCCs $\recurrentset{}$ of $\calM$ and for which  $\recurrentset{\pi}=\recurrentset{}$, i.e.,
\begin{align}
    \edgepreservingset = \left\{\pi\in\stationaryset: \begin{matrix} 
    \recurrentset{\pi} = \recurrentset{} ~\wedge~ 
    \pi(a|s) > 0, \forall s\in \recurrentset{}, a\in A(s) \end{matrix} \right\}\:.
    \label{eq:edge_preserving_set}
\end{align}
\end{definition}
Hence, for every state $s\in \recurrentset{}$ (see Definition \ref{def:TSCC}), an EP policy assigns a non-zero probability to every action in $A(s)$, and every state in $\transientset{}$ is either transient or isolated in the Markov chain induced by the policy. 
For example, the uniform policy which has $\pi(a|s) = 1/|A(s)|, \forall s\in S$ is in $\edgepreservingset$. Note that other policies in $\edgepreservingset$ could assign a very small probability (as long as it is non-zero) to non-rewarding transitions in $\recurrentset{}$. {\color{black} 
Using an open set definition in \eqref{eq:edge_preserving_set} simplifies the exposition and the subsequent theoretical analysis, however, it does not guarantee that an optimal policy from the set always exists.
We discuss and analyze variations of the problem formulation to address this issue at length in Section \ref{sec:open_set}.}


Next, we introduce two sets of policies whose definitions rest on two distinct notions of class preservation. 
\begin{definition}[Class-preserving policies]
\label{def:classpreservingset}
Given an MDP $\calM$ with TSCCs ~$\tscc{k}{}, k = 1, \ldots, m$, we define the set of Class-Preserving (CP) policies $\classpreservingset$ as the set of stationary policies that induce Markov chains with the same TSCCs as those of $\calM$, i.e.,  
\begin{align}
    \classpreservingset = \left\{\pi\in\stationaryset: \begin{matrix} 
    \recurrentset{\pi} = \recurrentset{} ~\wedge~ \forall k\in[m], \tscc{k}{\pi} = \tscc{k}{} \end{matrix} \right\}\:.
    \label{eq:class_preserving_set}
\end{align}
\end{definition}

Note that the condition $\recurrentset{\pi}=\recurrentset{}$ in Definitions \ref{def:edgepreservingset} and \ref{def:classpreservingset} implies that $\transientset{}$ consists of transient or isolated states in $\calM_\pi$ for any $\pi$ in $\edgepreservingset$ or $\classpreservingset$. \textcolor{black}{Per \eqref{eq:class_preserving_set}, a CP policy preserves the recurrence of all states in the TSCCs of the MDP but, unlike EP policies, its support need not be the entire set of actions available at said states.} Therefore, CP policies can conceivably achieve larger rewards than EP policies by averting non-rewarding transitions. 

\begin{definition}[Class-preserving up to unichain]
\label{def:uptounichainset}
Given an MDP $\calM$ with TSCCs ~$\tscc{k}{}, k = 1, \ldots, m$, we define the set of Class-Preserving-up-to-Unichain (CPU) policies $\uptounichainset$ as 
\begin{align}
    \uptounichainset = \left\{\pi\in\stationaryset: \begin{matrix} 
    \recurrentset{\pi} \subseteq\recurrentset{} ~\wedge~ \forall k\in[m], \exists! \:\tscc{k}{\pi}\subseteq\tscc{k}{} \end{matrix} \right\}
    \label{eq:uptounichainset}\:,
\end{align}
that is, the set of stationary policies that induce Markov chains $\calM_\pi$ in which the TSCCs of the MDP $\calM$ are reachable and unichain (i.e., each contains exactly one non-isolated, recurrent component) and the recurrent states are a subset of the recurrent states of $\calM$ (Recalling that the notation $\exists!$ in \eqref{eq:uptounichainset} refers to the existence of a unique set). 
\end{definition}
This definition captures a more relaxed notion of class preservation than \eqref{eq:class_preserving_set} for CP policies in that it relaxes the requirement that all states in the TSCCs of $\cal M$ be recurrent and reachable in the Markov chain ${\cal M}_\pi$ induced by the policy $\pi$, to the milder requirement that in ${\cal M}_\pi$ there exists a unique reachable recurrent class in each of the TSCCs of $\cal M$.     

%

\begin{figure}
    \centering
    \includegraphics[scale=1]{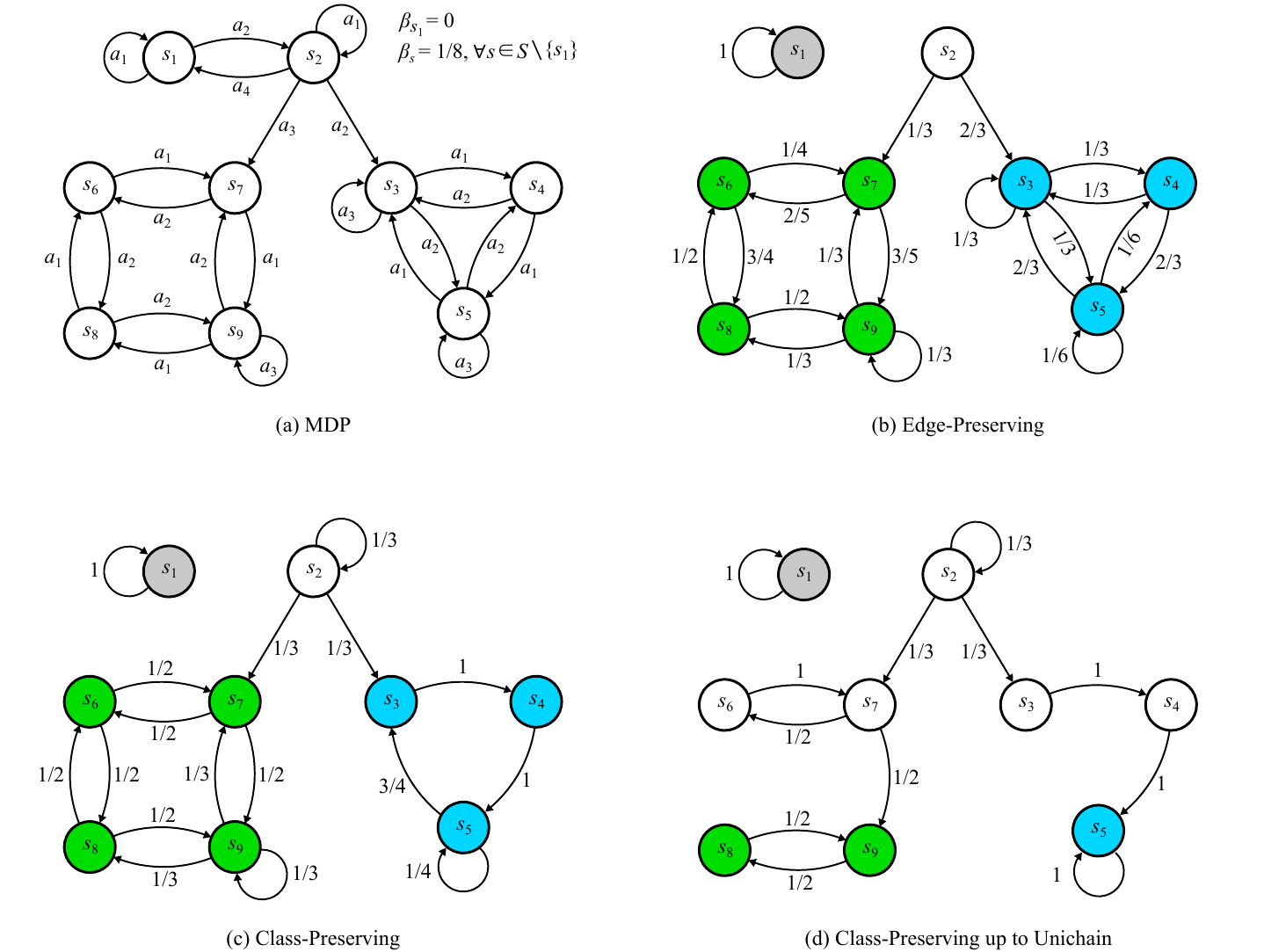}
    \caption{(a) MDP and Markov chains induced by (b) EP, (c) CP, and (d) CPU policies. For the MDP, all transitions are deterministic, i.e., $T(s'|s,a) \in \{0,1\}$ indicating if there is an outgoing edge from $s$ to $s'$ under action $a$, the rewards are defined such that $R(s_5,a_3) = R(s_8,a_1)=1$ and $0$ otherwise, and the initial probabilities are $\beta_{s_1} = 0$ and $\beta_{s} = 1/8, \forall s\in S\setminus\{s_1\}$. The numbers next to the edges of the Markov chains are the conditional probabilities $\pi(a|s)$ of the different actions given the states specifying the policies.}
    \label{fig:PolicyExamples}
\end{figure}
The aforementioned definitions are best illustrated by an example. Figure~\ref{fig:PolicyExamples}(b) illustrates a Markov chain induced by an EP policy, i.e., \textcolor{black}{one that plays every action available 
in the TSCCs of the MDP of Figure~\ref{fig:PolicyExamples}(a) with non-zero probability.} 
As shown, $s_1$ is isolated and $s_2$ is transient -- these would both be transient under the uniform policy. The TSCCs of the induced chain are highlighted with two separate colors. 
Examples of Markov chains 
induced by a CP and a CPU policy are shown in Figure~\ref{fig:PolicyExamples}(c) and (d), respectively. 
The Markov chain of Figure~\ref{fig:PolicyExamples}(c) has the exact same TSCCs of the MDP and of the Markov chain of Figure~\ref{fig:PolicyExamples}(b) induced by the EP policy, \textcolor{black}{with the fundamental difference that the CP policy is not supported on every action available} 
in the TSCCs (e.g., see the recurrent component highlighted in blue). By contrast, states $s_3, s_4, s_6$ and $s_7$ are transient in the Markov chain of Figure~\ref{fig:PolicyExamples}(d). The set consisting of states $s_3, s_4, s_5$ is unichain, having exactly one recurrent component (state $s_5$) and two transient states ($s_3$ and $s_4$). Similarly, the set composed of states $s_6, s_7, s_8, s_9$ is unichain with one recurrent component ($s_8$ and $s_9$) and two transient states ($s_6$ and $s_7$).   

{\color{black}
The classes of policies defined in \eqref{eq:edge_preserving_set}, \eqref{eq:class_preserving_set} and \eqref{eq:uptounichainset} satisfy the following relations.}

\begin{lemma}
$\edgepreservingset\subseteq\classpreservingset\subseteq\uptounichainset\:.$
\label{lem:relation_bet_policies}
\end{lemma}

We remark that the inclusions in Lemma \ref{lem:relation_bet_policies} are generally strict, except for some special MDPs. Specifically, given a general MDP $\cal M$, there may exist a policy $\pi\in\classpreservingset$ for which $\recurrentset{\pi} = \recurrentset{}$ 
and $\pi(a|s) = 0$ for some $a\in A(s), s\in\recurrentset{}$, in which case $\pi\notin\edgepreservingset$. 
Similarly, since a unichain may contain some transient states, $\classpreservingset$ is generally a proper subset of $\uptounichainset$.

\paragraph{Problem Definition.} We can readily define the class of problems \ac{name}$(\Pi)$, parametrized by a predefined set of stationary policies $\Pi$,  of finding a policy in the set $\Pi$  
that maximizes the expected average reward while satisfying a given set of steady-state specifications.

\begin{definition}[Steady-state policy synthesis (\ac{name})] \sloppy Given an LMDP $\calM = (S,A,T,R,\beta,L,\SSSpec)$ and a domain of policies $\Pi\subseteq\stationaryset$, the \ac{name}$(\Pi)$ problem is to find an optimal stochastic policy $\pi\in\Pi$ that maximizes the expected average reward defined in \eqref{eq:avg_reward} 
and satisfies the steady-state specifications $\SSSpec$ (Definition \ref{def:spec}), i.e.,
\begin{align}
\label{eq:MCSSPS}
\begin{aligned}
    &\max_{\pi\in\Pi} \sum_{s\in S}\sum_{a\in A(s)} \prbetapi(s,a) R(s,a)  \text{ subject to } \\
    &\sum_{s \in L_i}\sum_{a\in A(s)} \prbetapi(s,a) \in [l, u], ~~\forall (L_i, [l, u]) \in\SSSpec
\end{aligned}
\end{align}
\textcolor{black}{If the maximum in \eqref{eq:MCSSPS} cannot be attained over the domain $\Pi$, we define \ac{name}$(\Pi)$ as the problem of finding a policy $\pi\in\Pi$ that satisfies the specifications $\SSSpec$ and whose expected average reward $R_\pi^\infty(\beta)\geq \sup_{\pi'\in\Pi} R^\infty_{\pi'}(\beta) - \epsilon$, for some arbitrarily small $\epsilon > 0$ (See Section \ref{sec:open_set}).}   
\label{def:MSSPS}
\end{definition}

In this paper, we present solutions to \ac{name}$(\edgepreservingset)$,  \ac{name}$(\classpreservingset)$ and \ac{name}$(\uptounichainset)$, where $\Pi$ in \eqref{eq:MCSSPS} is set to $\edgepreservingset$, $\classpreservingset$, and $\uptounichainset$, respectively\footnote{Our work \cite{ijcai2020} has presented preliminary results for the \ac{name}$(\edgepreservingset)$ problem.}. To this end, we first determine the TSCCs~ $\recurrentset{}$ of $\calM$ and the complement set $\transientset{}$ using standard techniques from graph theory \cite{Tarjan1972DepthFirstSA}. These are then used to define an \ac{LP} from which the solution policy is derived. 

{\color{black}
\section{Linear Programming Based Solutions}
\label{sec:lps}
In this section, we present our linear-programming-based solution to the \ac{name} problem \eqref{def:MSSPS} over edge- and class-preserving policies. We formulate linear programs that encode constraints on the limiting distributions of said policies to solve \ac{name}$(\edgepreservingset)$ and \ac{name}$(\classpreservingset)$. 
The optimal solutions to the formulated programs provably yield optimal edge- and class-preserving policies that meet the desired specifications. The encoded constraints 
are also at the center of an iterative algorithm described in Section \ref{sec:mmscpu} to generate CPU policies for \ac{name}$(\uptounichainset)$. Our main results on \ac{name} in edge- and class-preserving policies are presented in Sections \ref{sec:mmsep}, \ref{sec:mmscp} and \ref{sec:mmscpu}. 
To simplify the exposition, all proofs are deferred to the appendix.}

\textcolor{black}{We present three main programs. 
The first, 
used for the synthesis of optimal EP policies, is the most constrained as it encodes the requirement that every action in the terminal components must be played with non-zero probability. While this condition is 
not necessary in order to ensure one-to-one
correspondence between the feasible solutions and the induced steady-state distributions, it results in a simple program whose solution provably yields an optimal EP policy that meets the
specifications. The second program 
relaxes this condition to the milder requirement that every state in the terminal components is visited infinitely often (which may not require playing every action
available), but at the expense of additional complexity. Specifically, its solution yields an optimal policy that meets the specifications from the class of CP policies (a superset of
EP policies) but uses more complex flow constraints to encode said requirement. The third program 
is the least constrained and is used to synthesize a policy
from the larger class of CPU policies. While its solution is not guaranteed to yield a CPU policy, we derive a characterization of its optimal solution, which inspires a greedy algorithm to construct such policy. 
We augment the program
by iteratively adding constraints until convergence. The algorithm is guaranteed to
converge in a finite number of steps to a (possibly) suboptimal CPU policy that meets the
specifications.}

\textcolor{black}{
By encoding constraints on 
the limiting distribution of the Markov chain induced by a stationary policy derived from an \ac{LP} solution, the policy is ultimately absorbed in the TSCCs of the MDP.  
This restricts the long-term play to the TSCCs, which once reached, cannot be escaped. By imposing strict positivity on state-action pairs or flow  constraints in the TSCCs, we further ensure that these components are unichain, and in turn, the long-term frequencies induced by the policy 
match the solution from which the policy is generated.}

\subsection{ \ac{name}$(\edgepreservingset)$ -- Synthesis over Edge-Preserving Policies}
\label{sec:mmsep}
In this section, we formulate a linear program to solve \ac{name}$(\edgepreservingset)$ defined in \eqref{eq:MCSSPS}, which seeks to maximize the expected average reward subject to specification constraints over the class of policies $\edgepreservingset$ in \eqref{eq:edge_preserving_set}. 

{\color{black}
Given MDP $\calM$, define $Q_0$ to be the set of vectors $x, y$ satisfying 
\begin{equation}
\left\{\begin{array}{lr}
 \LPmainconstraintlabel{LPmain_x}   \quad  
 \sum_{s \in S} \sum_{a \in A(s)} x_{sa} T(s' \mid s,a) = \sum_{a \in A(s')} x_{s'a}, ~~\forall s' \in S\\ [\medskipamount]
  \LPmainconstraintlabel{LPmain_y}  \quad  
\sum_{s \in S} \sum_{a \in A(s)} y_{sa}  T(s' \mid s,a)  = \sum_{a \in A(s')} (x_{s'a} + y_{s'a}) - \beta_{s'}, ~~ \forall s' \in S\\ [\medskipamount]
 \LPmainconstraintlabel{LPmain_transx}  \quad 
\sum_{f \in \transientset{}} \sum_{a \in A(f)} x_{fa} = 0,\\
[\medskipamount]
x_{sa} \in [0,1],  y_{sa} \ge 0, \forall s \in S, a \in A(s), f \in \transientset{}, k \in [\NumErgSets]  
\end{array}\right.
\label{eq:Q0}
\end{equation}
}
%


We can readily formulate $\mathrm{LP}_1$ \eqref{eqn:LP_EP} to synthesize {\color{black} optimal} EP policies, which incorporates two additional constraints beside the constraints in \eqref{eq:Q0}. 
{\color{black}
\begin{equation}
(\mathrm{LP}_1)\quad
\begin{aligned}
    \max \quad & \sum_{s \in S} \sum_{a \in A(s)} x_{sa} R(s,a) 
    \text{ subject to }   (x,y) \in Q_0       \\
    \LPmainconstraintlabel{LPmain_SSspecs}
   	\quad &l_i \le \sum_{s \in L_i}\sum_{a \in A(s)}  x_{sa} \le u_i,
   	~ \forall (L_i, [l_i,u_i]) \in \SSSpec \\
    \LPmainconstraintlabel{LPmain_strictposx}
         \quad & x_{sa} > 0,
                 ~ \forall s \in \tscc{k}{}, k\in [\NumErgSets], a \in A(s) \\  
\end{aligned}
\label{eqn:LP_EP}
\end{equation}
Constraints \ref{LPmain_x} --  \ref{LPmain_transx} constrain the limiting distributions of Markov chains induced by the policies of interest, and are thus part of the constraint set of all programs we formulate in this work.} In particular, they capture the structure of the stationary matrix $\cesarolimit{T}$ corresponding to the classifications $\recurrentset{}$ and $\transientset{}$
(See Definition \ref{def:TSCC}).
Constraint \ref{LPmain_x} ensures that $x$ is a stationary distribution \cite{altman1999constrained,puterman1994markov};
constraint \ref{LPmain_y}, which is described in \textcolor{black}{\cite[Chapter 4]{kallenberg1983linear} and \cite[Sec 9.3]{puterman1994markov}}, enforces consistency in the expected average number of visits $y_{fa}$ for any transient 
state-action pair $f \in \transientset{}, a \in A(f)$; constraint \ref{LPmain_transx} preserves the non-recurrence of the states $f \in \transientset{}$ by forcing zero steady-state probability.
%
%
Constraint \ref{LPmain_SSspecs} encodes the steady-state specifications. The strict positivity constraint \ref{LPmain_strictposx} preserves the transitions in the TSCCs to yield EP policies. \textcolor{black}{In practice, we transform the strict inequalities to bounded ones by introducing an arbitrarily
small constant on the right-hand side, thereby ensuring an optimal solution always exists (See Section \ref{sec:open_set}).} Enforcing constraints on the occupation measures  
ensures that, from any state $f \in \transientset{}$, the process will be ultimately absorbed into the TSCCs $\tscc{k}{}, k\in[\NumErgSets]$.

\textcolor{black}{The next theorem guarantees that every feasible solution to $\mathrm{LP}_1$ yields an EP policy.} 
\begin{theorem}
\label{thm:pi_in_ext_edge_set}
	Given an LMDP $\calM$, let $(x,y)\in Q_1$, where $Q_1$ is the feasible set of solutions to $\mathrm{LP}_1$ \eqref{eqn:LP_EP}, and let $\pi:=\pi(x,y)$ be defined as in \eqref{eq:policy_LP1}. 
Then, $\pi\in\edgepreservingset$.
\end{theorem}


{\color{black}\noindent We can readily state the following theorem establishing the correctness of $\mathrm{LP}_1$. It guarantees that the policy synthesized from an optimal solution to \eqref{eqn:LP_EP} \textcolor{black}{(if one exists)} is not only in $\edgepreservingset$, but also is optimal among all such policies and meets the steady-state specifications, i.e., solves \ac{name}$(\edgepreservingset)$.}  
\begin{theorem}
\label{thm:mainEP}
Given an LMDP $\calM = (S,A,T,R,\beta,L,\SSSpec)$, $\mathrm{LP}_1$ in \eqref{eqn:LP_EP} is feasible iff there exists a policy $\pi\in\edgepreservingset$ such that the Markov chain $\calM_\pi = (S,T_\pi,\beta)$ satisfies the specifications $\SSSpec$. 
Further, given an optimal solution $x^*, y^*$ of \eqref{eqn:LP_EP}, the policy $\pi^*:= \pi(x^*,y^*)$ as defined in \eqref{eq:policy_LP1} is optimal in the class of policies $\edgepreservingset$ and meets the specifications $\SSSpec$.
\end{theorem}

\subsection{\ac{name}$(\classpreservingset)$ -- Synthesis over Class-Preserving Policies}
\label{sec:mmscp}
The strict positivity constraint \ref{LPmain_strictposx} of $\mathrm{LP}_1$ forces the policy \textcolor{black}{to play every action in the TSCCs of the MDP $\calM$ (by assigning non-zero probability to every action available)}, 
which may be restrictive and often unnecessary. Indeed, as we show,
in order to ensure one-to-one correspondence between the optimal solutions of a formulated \ac{LP} and the optimal policies of the constrained MDP derived from these solutions, it suffices to preserve the recurrence or the unichain property of these components. 

To address this restriction, we introduce flow constraints in $\mathrm{LP}_2$ given in \eqref{eqn:LP_Flow} (replacing constraint \ref{LPmain_strictposx}) \textcolor{black}{to ensure the recurrence of the components $\tscc{k}{}, k\in[\NumErgSets]$, in the induced chain.} 
%
%
It helps to introduce some notation in order to express such constraints. We define the transition relation of an MDP by $T^{\text{rel}} = \{(s, s') \in S \times S | s \neq s' \wedge \exists a \in A(s), T(s' | s, a) > 0\}$ 
\cite{IJCAI2019}. This corresponds to the graph structure of the MDP. For each TSCC $\tscc{k}{}$, we further define its graph structure as $T^\text{rel}_k = T^\text{rel} \cap \tscc{k}{} \times \tscc{k}{}$. We can now add flow constraints 
in order to ensure that, for the Markov chain induced by the solution policy, \textcolor{black}{each set $\tscc{k}{}$ will remain a recurrent class without necessarily having to take every action available in that set.} 

\begin{equation}
(\mathrm{LP}_2)\quad
	\begin{aligned}
	\max \quad & \sum_{s \in S} \sum_{a \in A(s)} x_{sa} \sum_{s' \in S} T(s' | s, a) R(s, a, s')
	\text{ subject to }  (x,y)\in Q_0,  \\ 
	\ref{LPmain_SSspecs}  \quad &  l_i \le \sum_{s \in L_i}\sum_{a \in A(s)}  x_{sa} \le u_i, \text{ } & \hspace*{-64pt} \forall (L_i, [l_i,u_i]) \in \SSSpec \\
	\LPmainconstraintlabel{LP_flowInit} \quad & f_{s_i s'} = \sum_{a \in A(s_i)} T(s' | s_i, a) x_{s_i a} \text{ } & \hspace*{-128pt} \forall (s_i, s') \in T^\text{rel}_k, k \in [m] \\ 
    \LPmainconstraintlabel{LP_flowRevInit} \quad & f^{\text{rev}}_{s_i s'} = \sum_{a \in A(s')} T(s_i | s', a) x_{s' a} & \hspace*{-128pt} \forall (s', s_i) \in T^\text{rel}_k, k \in [m] \\  
    \LPmainconstraintlabel{LP_flowCap} \quad & f_{s s'} \leq \sum_{a \in A(s)} T(s' | s, a) x_{s a} & \hspace*{-96pt} \forall (s, s') \in T^{\text{rel}}_k, k \in [m] \\ 
    \LPmainconstraintlabel{LP_flowRevCap} \quad & f^{\text{rev}}_{s s'} \leq \sum_{a \in A(s')} T(s | s', a) x_{s' a} & \hspace*{-64pt} \forall (s', s) \in T^{\text{rel}}_k, k \in [m] \\
     \LPmainconstraintlabel{LP_flowTransfer} \quad & \sum_{(s', s) \in T^{\text{rel}}} f_{s' s} > \sum_{(s, s') \in T^{\text{rel}}} f_{s s'} & \hspace*{-92pt} \forall s \in \tscc{}{} \setminus \{s_i\} \\ 
    \LPmainconstraintlabel{LP_flowRevTransfer} \quad & \sum_{(s, s') \in T^{\text{rel}}} f^{\text{rev}}_{s' s} > \sum_{(s', s) \in T^{\text{rel}}} f^{\text{rev}}_{s s'} & \hspace*{-92pt} \forall s \in \tscc{}{} \setminus \{s_i\} \\ 
    \LPmainconstraintlabel{LP_flowIn} \quad & \sum_{(s', s) \in T^{\text{rel}}} f_{s' s} > 0 & \hspace*{-92pt} \forall s \in \tscc{}{} \\ 
    \LPmainconstraintlabel{LP_flowRevIn} \quad & \sum_{(s, s') \in T^{\text{rel}}} f^{\text{rev}}_{s' s} > 0 & \hspace*{-92pt} \forall s \in \tscc{}{} \\ 
	\LPmainconstraintlabel{LP_flowVars} \quad
	& f_{s s'}, f^{\text{rev}}_{s s'} \in [0, 1]
	& \hspace*{-96pt} \forall (s, s') \in T^{\text{rel}}
	\end{aligned}
	\label{eqn:LP_Flow}
\end{equation}


The program $\mathrm{LP}_2$ in \eqref{eqn:LP_Flow} is such that every state in $\tscc{k}{}$ can reach and is reachable from every other state in $\tscc{k}{}$. For each $k \in [m]$, constraint \ref{LP_flowInit} induces an initial flow out of a randomly chosen state $s_i \in \tscc{k}{}$ and into its neighbors $s'$, that is proportional to the transition probability $T_\pi (s' | s_i)$ in the Markov chain induced by the solution policy; constraint \ref{LP_flowCap} establishes the flow capacity between states in a similar manner; \ref{LP_flowTransfer} ensures that the incoming flow into every state in $\tscc{k}{}$ is greater than the outgoing flow; finally, constraint \ref{LP_flowIn} ensures that there is incoming flow into every state in $\tscc{k}{}$. These constraints ensure that every state in $\tscc{k}{}$ is reachable from $s_i$, whereas constraints \ref{LP_flowRevInit}, \ref{LP_flowRevCap}, \ref{LP_flowRevTransfer}, \ref{LP_flowRevIn} address the foregoing in the reverse graph structure of the MDP, thereby ensuring that $s_i$ is reachable from all states in $\tscc{k}{}$. {\color{black} We remark that the feasible set in \eqref{eqn:LP_Flow} is a superset of that in \eqref{eqn:LP_EP} since the flow constraints \ref{LP_flowInit}--\ref{LP_flowVars} are implied by \ref{LPmain_strictposx}, hence $\mathrm{LP}_2$ is less-constrained than $\mathrm{LP}_1$.

We can readily state the following two theorems establishing the correctness of $\mathrm{LP}_2$, which are the counterparts of Theorem \ref{thm:pi_in_ext_edge_set} and \ref{thm:mainEP}. In particular, Theorem \ref{thm:pol_in_CP} guarantees that the solution to $\mathrm{LP}_2$ is a CP policy, while Theorem \ref{thm:mainCP} establishes that the policy \eqref{eq:policy_LP1} derived from the optimal solution to $\mathrm{LP}_2$ in \eqref{eqn:LP_Flow} solves \ac{name}$(\classpreservingset)$, i.e., is optimal among the class of CP policies and meets the steady-state specifications.}

\begin{theorem}
\label{thm:pol_in_CP}
Given an LMDP $\calM$, let $(x,y)\in Q_2$ and $\pi$ be defined as in \eqref{eq:policy_LP1}, where $Q_2$ is the feasible set of solutions to $\mathrm{LP}_2$ \eqref{eqn:LP_Flow}. Then, $\pi\in\classpreservingset$.	
\end{theorem}


%
\begin{theorem}
\label{thm:mainCP}
Given an LMDP $\calM = (S,A,T,R,\beta,L,\SSSpec)$, the LP in \eqref{eqn:LP_Flow} is feasible iff there exists a policy $\pi\in\classpreservingset$, where $\classpreservingset$ is defined in \eqref{eq:class_preserving_set}, such that the Markov chain $\calM_\pi = (S,T_\pi,\beta)$ satisfies the specifications $\SSSpec$. 
Further, given an optimal $x^*,y^*$ of \eqref{eqn:LP_Flow}, the policy $\pi(x^*,y^*)$ defined in \eqref{eq:policy_LP1} is optimal in the class of policies $\classpreservingset$ and meets the specifications $\SSSpec$.
\end{theorem}

\subsection{\ac{name}$(\uptounichainset)$ -- Synthesis over Class-Preserving up to Unichain Policies}
\label{sec:mmscpu}
In this section, we discuss policy synthesis over the larger set of policies $\uptounichainset$. We provide a sufficient condition under which we can identify an optimal policy $\pi\in\uptounichainset$ that meets the specifications. Based on this result, we develop an iterative algorithm to construct a policy in $\uptounichainset$ that provably meets the desired specifications.  

%

{\color{black} Next, we give a sufficient condition for \ac{name}$(\uptounichainset)$, characterized in terms of the set of optimal solutions to $\mathrm{LP}_3$ in \eqref{eq:LP0_and_specs}.
\begin{equation}
\mathrm{LP}_3: \max \quad  \sum_{s \in S} \sum_{a \in A(s)} x_{sa} \sum_{s' \in S} T(s' | s, a) R(s, a, s')
	\text{ subject to } (x,y) \in Q_0 \text{ and } \ref{LPmain_SSspecs}
\label{eq:LP0_and_specs}
\end{equation}
Note that the feasible set of $\mathrm{LP}_3$ is the intersection of the set $Q_0$ in \eqref{eq:Q0} and the set of variables satisfying the steady-state specifications, that is, without the positivity or flow constraints in \eqref{eqn:LP_EP} and \eqref{eqn:LP_Flow}, respectively.}

Recall that a strongly connected digraph is one in which it is possible to reach any node starting from any other node by traversing the directed edges in the directions in which they point. Theorem \ref{thm:CPU} states that the policy $\pi$ in \eqref{eq:policy_LP1}, derived from an optimal solution to $\mathrm{LP}_3 $ in \eqref{eq:LP0_and_specs}, solves \ac{name}$(\uptounichainset)$ if the directed subgraphs corresponding to the support of the optimal solution in the TSCCs of $\calM$ are strongly connected. In Theorem \ref{thm:CPU}, we define the digraph associated with the support of a given solution $x$ as the graph whose vertices are all states $s$ with $x_s > 0$ and whose edges correspond to actions $a$ for which $x_{sa} > 0$.
\begin{theorem}
Given LMDP $\calM$, let $Q^*$ be the set of optimal solutions of $\mathrm{LP}_3$ \eqref{eq:LP0_and_specs} and $X^* :=\{x:(x,y)\in Q^* \text{ for some } $y$\}$. Given $x\in X^*$, let 
$V_k^+(x) := \{s\in\tscc{k}{}:x_s>0\}$ 
and $E_k^+(x):=\{(s,a)\in\tscc{k}{}\times A(s): x_{sa} > 0\}$. 
If 
the directed subgraph $(V_k^+(x),E_k^+(x))$ is strongly connected $\forall k\in[m]$, then the policy $\pi$ in \eqref{eq:policy_LP1} is optimal in the class of policies $\uptounichainset$ and meets the specifications in $\SSSpec$.  

\label{thm:CPU}
\end{theorem}


\begin{corollary}
If the condition in the statement of Theorem \ref{thm:CPU} holds for all $x\in X^*$, then $\mathrm{LP}_3$ \eqref{eq:LP0_and_specs} solves \ac{name}$(\uptounichainset)$.
\end{corollary}

\subsubsection{Generation of policies in $\uptounichainset$}
Inspired by Theorem \ref{thm:CPU}, we devise a row-generation-based algorithm to search for a policy $\pi\in\uptounichainset$ as shown in Algorithm \ref{alg:CPU}. 
First, $\mathrm{LP}_3$ \eqref{eq:LP0_and_specs} is solved. If the digraph corresponding to the support of the obtained solution is strongly connected for each of the TSCCs of LMDP $\calM$, the policy in \eqref{eq:policy_LP1} is computed and the search stops. However, if the solution does not correspond to a strongly connected digraph in some TSCC, then there must exist a non-empty set of states with no outgoing edges to the rest of the states in that TSCC. Hence, for some $k\in[m]$ we find a cut, that is, a set of states $C$ that has no outgoing edges to the complement set $\tscc{k}{}\setminus C$. The constraint in \eqref{eq:constraint_cut}  corresponding to this cut is added to include the edges in the support, where $A' = \{a\in A(s): T(s'|s,a) > 0, s'\in\tscc{k}{}\setminus C\}$. The constraint forces the addition of missing edges across this cut in a greedy manner (by forcing the sum of the state-action variables corresponding to these edges to be non-zero) to eventually produce a strongly connected digraph. The process is repeated until a strongly connected solution is found. 
\begin{equation}
    \sum_{s\in C}\sum_{a\in A'} x_{sa} > 0
    \label{eq:constraint_cut}
\end{equation}
\textcolor{black}{Algorithm \ref{alg:CPU} is guaranteed to converge to a (possibly suboptimal) policy in $\uptounichainset$ in a finite number of steps, since in the worst case (when all edges are included) it will yield a policy in $\edgepreservingset\subseteq\uptounichainset$ under which all edges in the TSCCs of $\calM$ are retained.} The finiteness of the number of steps is because the number of cuts in the finite MDP is bounded above by $O(\max_{k\in[m]}2^{|\tscc{k}{}|})$. Our experiments have shown that Algorithm \ref{alg:CPU} converges to a policy in $\uptounichainset$ after a small number of iterations.

\begin{algorithm}
\caption{Generation of a policy $\pi\in\uptounichainset$}
{
\begin{algorithmic}
\footnotesize
\REQUIRE{LMDP $\calM$ with specifications $\SSSpec$.}
\ENSURE{Stationary policy $\pi\in\uptounichainset$ which satisfies $\SSSpec$.}

\STATE{Determine the TSCCs $\tscc{k}{}, k\in\NumErgSets$ of $\calM$}
\STATE{$isSConnected = False$, $C=\{.\}$, $A'=\{.\}$}

\WHILE{$isSConnected = False$}
    \STATE{Solve LP~\eqref{eq:LP0_and_specs} with constraint \eqref{eq:constraint_cut} to get optimal values $x^*_{sa}, y^*_{sa}, \forall (s,a) \in S\times A(s)$.}
    \STATE{Compute the support $E_k^+(x^*)$ of each TSCC corresponding to $x^*$ (See Theorem \ref{thm:CPU})}.

    \IF{digraph $(V_k^+(x^*),E_k^+(x^*))$ forms a SCC for every $k\in [m]$}
        \STATE{compute $\pi$ using \eqref{eq:policy_LP1}}
        \STATE{$isSConnected = True$}
    \ELSE
        \STATE{find a cut and update $C$ and $A'$}
    \ENDIF
\ENDWHILE

\end{algorithmic}
}
\label{alg:CPU}
\end{algorithm}

{\color{black}
\subsection{Additional Insights}
\label{sec:discussion}
This section provides additional remarks and examples to shed more light on the linear programming formulations. The section may be skipped without loss of continuity.
\smallbreak
\subsubsection{Non-surjective mapping} 
All occupation measures induced by the policies of interest are elements of $Q_0$ \eqref{eq:Q0}, that is, $\prbetapol{\pi}\in X_0:=\{x:(x,y)\in Q_0 \text{ for some } y\}$ if $\pi\in\uptounichainset$, which is affirmed by Lemma \ref{thm:XPi_in_LP} stated in \ref{sec:appendx_proofaux}. 
%
However, in general, $\setssdistbeta(\uptounichainset)\subset X_0$, i.e., the set $\setssdistbeta(\uptounichainset)$ is a \emph{proper} subset of $X_0$. 
In mathematical terms, the mapping \eqref{eq:betaPstar} between the set of policies $\uptounichainset$ and the set $X_0$ is injective but non-surjective. In turn, there may exist elements of $X_0$ that are unpaired with policies in $\uptounichainset$. This is illustrated by the following example. 
%
\begin{example}
Consider the MDP in Figure \ref{fig:uni_multi_chain}(b). It is easy to see that $x$ for which $x_{s_2 a_2} = x_{s_3 a_2} = 1/2$, $x_{s_1 a_1} = x_{s_2 a_1} = x_{s_3 a_1} = 0$ is in $X_0$, i.e., $x\in X_{0}$. However, the only policies in $\uptounichainset$ 
that satisfy $\prbetapi(s_2,a_2)+\prbetapi(s_3,a_2) = 1$ are the deterministic policies $\pi_1, \pi_2$, which have $\pi_1(a_2|s_2) = 1, \pi_1(a_2|s_3) = 0$ (for which state $s_2$ is recurrent and $s_3$ is transient) and $\pi_{2}(a_2|s_2) = 0, \pi_2(a_2|s_3) = 1$ (for which state $s_3$ is recurrent and $s_2$ is transient). However, $\prbetapol{\pi_1}(s_2,a_2) = \prbetapol{\pi_2}(s_3,a_2) = 1$. Thus, $x\notin\setssdistbeta(\uptounichainset)$. 
\end{example}

\subsubsection{Insufficient constraint set}
The set $Q_0$ correctly encodes constraints on the limiting distributions of Markov chains induced by policies in $\uptounichainset$ (with the state classification corresponding to $\recurrentset{}$ and $\transientset{}$). However, 
the lack of one-to-one correspondence between feasible solutions and policies (see Section \ref{sec:limitations}) is not fully resolved by the constraint set \eqref{eq:Q0} without the additional constraints in \eqref{eqn:LP_EP} or \eqref{eqn:LP_Flow}.  
In particular, consider the linear program $\mathrm{LP}_0$ with feasible set $Q_0$
\begin{equation}
(\mathrm{LP}_0): \max \quad  \sum_{s \in S} \sum_{a \in A(s)} x_{sa} \sum_{s' \in S} T(s' | s, a) R(s, a, s')
	\text{ subject to }  (x,y) \in Q_0
\label{eq:LP0}
\end{equation}
The steady-state distribution of the policy \eqref{eq:policy_LP1} derived from an optimal solution $(x^*, y^*)$ to $\mathrm{LP}_0$ is generally not equal to
$x^*$. In turn, specifications encoded as constraints on the state-action variables as in \eqref{eq:LP0_and_specs} will not necessarily be met by the policy. This is best illustrated via a simple example.
%

%
%
\begin{example}
\label{ex:lim_LP0}
Revisit the three-state example of Figure \ref{fig:uni_multi_chain}(b) and define the rewards $R(s_1,a_1) = R(s_1,a_2) = R(s_2,a_1) = R(s_3,a_1) = 0$, $R(s_2,a_2) = R(s_3,a_2) = 1$ and initial distribution $\beta_{s_1} = 0, \beta_{s_2} = \beta_{s_3} = 1/2$. The MDP has one TSCC such that, $\recurrentset{} = \tscc{1}{} = \{s_2,s_3\}, \transientset{} = \{s_1\}$. The solution to $\mathrm{LP}_0$ in \eqref{eq:LP0} which has $x^*_{s_1 a_1} = x^*_{s_1 a_2} = x^*_{s_2 a_1} = x^*_{s_3 a_1} = 0, x^*_{s_2 a_2} = 1/3, x^*_{s_3 a_2} = 2/3, y^*_{s_1 a_1} = y^*_{s_1 a_2} = y^*_{s_3 a_1} = 0, y^*_{s_2 a_1} = 1/6$, is optimal (albeit not unique). However, the policy $\pi:=\pi(x^*,y^*)$ has $\prbetapol{\pi}(s_2,a_2) = \prbetapol{\pi}(s_3,a_2) = 1/2$, hence in general $\prbetapol{\pi}\ne x^*$ .
\end{example}

\subsubsection{Remarks on \ac{name}$(\uptounichainset)$}
%
\noindent 1) Note that, in the previous example, the derived policy $\pi\notin\uptounichainset$. 
However, if $\pi:=\pi(x^*,y^*)\in\uptounichainset$, where $(x^*,y^*)$ is an optimal solution to \eqref{eq:LP0_and_specs}, 
then $\pi$ will be optimal over $\uptounichainset\supseteq\classpreservingset$ i.e., solves \ac{name}$(\uptounichainset)$. This follows from the optimality of $(x^*,y^*)$ and Lemma \ref{thm:mainCPU} in the appendix, which gives a sufficient condition for the existence of a one-to-one correspondence between the elements of $Q_0$ and the steady-state distribution of policy \eqref{eq:policy_LP1}. 



\noindent 2) In general, if we dispense with the flow constraints in $\eqref{eqn:LP_Flow}$, we have no guarantee that the TSCCs $\tscc{k}{}$ will be unichain in $\calM_\pi$ under such $\pi$. For example, $\calM_\pi$ induced by the policy $\pi$ given in Example \ref{ex:lim_LP0} has $\tscc{1}{\pi} = \{2\}, \tscc{2}{\pi} = \{3\}$, i.e., $\pi\notin\uptounichainset$.  However, if the rewards  in this example are modified  
such that $R(s_1,a_1)\ne R(s_2,a_1)$ while keeping all other rewards unchanged, then $\pi\in\uptounichainset$. Therefore, under certain sufficient conditions on the reward vector, $\mathrm{LP}_3$ in \eqref{eq:LP0_and_specs} solves \ac{name}$(\uptounichainset)$.
}

{\color{black}
\subsubsection{Existence of Optimal Policies}
\label{sec:open_set}
\noindent\textbf{Modified LP.} The feasible set $Q_1$ for $\mathrm{LP}_1$ is not compact given the strict inequalities of constraint \ref{LPmain_strictposx} in \eqref{eqn:LP_EP}. Therefore, the maximum in \eqref{eqn:LP_EP} may not always be attained on the set. This can be easily remedied by replacing the strict inequalities with bounded ones via introducing an arbitrarily
small constant $\epsilon > 0$ on the right-hand side. Even when such requirement is not made explicit, a constant $\epsilon$ is dictated by the numerical precision of the LP solvers. We define $\mathrm{LP}_1(\epsilon)$ similar to \eqref{eqn:LP_EP}, with constraints \ref{LPmain_strictposx}  replaced with the bounded inequalities in $(v)'$ for some $\epsilon > 0$,
\begin{equation}
\mathrm{LP}_1(\epsilon)\quad
\begin{aligned}
    \max \quad & \sum_{s \in S} \sum_{a \in A(s)} x_{sa} R(s,a) 
    \text{ subject to }   (x,y) \in Q_0,\:\: \ref{LPmain_SSspecs},      \\
    (v)'
         \quad & x_{sa} \geq \epsilon,
                 ~ \forall s \in \tscc{k}{}, k\in [\NumErgSets], a \in A(s)\:. 
\end{aligned}
\label{eqn:LP_EP_modified}
\end{equation}
Theorem \ref{thm:EP_with_eps} stated next is analogous to Theorem \ref{thm:mainEP} with the modified program $\mathrm{LP}_1(\epsilon)$; it establishes that every feasible solution of $\mathrm{LP}_1(\epsilon)$ yields a policy that is in $\edgepreservingset$, and conversely, for every EP policy that meets the steady-state specifications, there exists an $\epsilon>0$ such that its steady-state distribution is $\mathrm{LP}_1(\epsilon)$-feasible. Moreover, the policy obtained from the optimal solution to $\mathrm{LP}_1(\epsilon)$ solves \ac{name}($\edgepreservingset$), that is, its expected average reward 
can be made arbitrarily close to the supremum over the set $\edgepreservingset$ as  $\epsilon\rightarrow 0$. 
\begin{theorem}
Given an LMDP $\calM = (S,A,T,R,\beta,L,\SSSpec)$ and $\mathrm{LP}_1(\epsilon)$ as in \eqref{eqn:LP_EP_modified}, then
\begin{enumerate}[(1)]
    \item The policy $\pi$ in \eqref{eq:policy_LP1} corresponding to a feasible solution of $\mathrm{LP}_1(\epsilon)$ is in $\edgepreservingset$.
    \item If $\exists\pi\in\edgepreservingset$ and $\pi$ meets the specifications $\SSSpec$, then $\exists\epsilon>0$ such that $\prbetapi$ is a feasible solution of $\mathrm{LP}_1(\epsilon)$.
    \item Let $x^*,y^*$ be an optimal solution to $\mathrm{LP}_1(\epsilon)$ and $\pi^*:=\pi(x^*,y^*)$ the corresponding policy in \eqref{eq:policy_LP1}. Then,
    \begin{align}
        \lim_{\epsilon\rightarrow 0} \left(\sup_{\pi\in\edgepreservingset}R^\infty_\pi(\beta) - R^\infty_{\pi^*}(\beta) \right)= 0
    \end{align}
\end{enumerate}
\label{thm:EP_with_eps}
\end{theorem}

A similar result holds for CP policies if the maximum in \eqref{eqn:LP_Flow} cannot be attained over the feasible set by transforming constraints \ref{LP_flowTransfer}--\ref{LP_flowRevIn} to bounded ones. The generalization is straightforward, thus omitted for brevity. 
\bigbreak
\noindent\textbf{Compact policy set -- policies with bounded support.} The foregoing existence issue stems from the open set definition of $\edgepreservingset$ in \eqref{eq:edge_preserving_set}, a result of which is that an optimal policy from the set (i.e., one that maximizes the average reward) may not always exist. Therefore, we introduce a slightly modified definition next, in which we force a lower bound on the values a policy assumes on its support, i.e., require that $\pi(a|s)\geq\delta$, for some arbitrarily small constant $\delta>0$. 
We formally introduce the definition of the compact set of policies, then state a result analogous to Theorem \ref{thm:mainEP} based on this definition for completeness. 
}

{\color{black}
\begin{definition}
\label{def:modifiedEP}
Given an MDP $\calM$ and some small $\delta$, where ~$0< \delta < 1/\max_{s\in\recurrentset{}} |A(s)|$, we define the set $\:\edgepreservingset(\delta)\subset\edgepreservingset$ of EP policies of bounded support as, 
\begin{align}
    \edgepreservingset(\delta) = \left\{\pi\in\stationaryset: \begin{matrix} 
    \recurrentset{\pi} = \recurrentset{} ~\wedge~ 
    \pi(a|s) \geq \delta, \forall s\in \recurrentset{}, a\in A(s) \end{matrix} \right\}\:.
    \label{eq:modifiedEP}
\end{align}
\end{definition}


\begin{theorem}
Given an LMDP $\calM = (S,A,T,R,\beta,L,\SSSpec)$ and the set $\edgepreservingset(\delta)$ in \eqref{eq:modifiedEP},
\begin{enumerate}[(1)]
    \item The policy $\pi$ in \eqref{eq:policy_LP1} corresponding to a feasible solution of $\mathrm{LP}_1(\delta)$ is in $\edgepreservingset(\delta)$.
    \item Let $x^*,y^*$ be an optimal solution to $\mathrm{LP}_1(\delta)$ and $R^*(\delta)$ the average reward of the corresponding policy in \eqref{eq:policy_LP1}. Then,
    \begin{align}
        \lim_{\delta\rightarrow 0} \max_{\pi\in\edgepreservingset(\delta)}R_\pi^\infty(\beta) - R^*(\delta) = 0. 
    \end{align}
\end{enumerate}
\label{thm:modifiedEP}
\end{theorem}

According to Theorem \ref{thm:modifiedEP}, every feasible solution to $\mathrm{LP}_1(\delta)$ yields a policy in $\Pi_{EP}(\delta)$. Also, the gap between the optimal expected average reward over the set $\Pi_{EP}(\delta)$ and the optimal reward of $\mathrm{LP}_1(\delta)$ approaches zero as $\delta\rightarrow 0$.
}

{\color{black}
\section{Extensions}
\label{sec:extensions}
In this section, we explore extensions beyond class-preserving policies, as well as an alternative type of specifications applicable to transient states. 
}
\subsection{Beyond Class-Preserving Policies}
\label{sec:beyond_CP}
In this section, we derive an alternative condition given in Theorem \ref{thm:dual_cone} under which $\mathrm{LP}_3$ in \eqref{eq:LP0_and_specs} is guaranteed to yield a stationary policy whose steady-state distribution meets the desired specifications. The policy generated need not be in $\uptounichainset$. The proof of Theorem \ref{thm:dual_cone} follows from the sufficient and necessary optimality conditions of program \eqref{eq:LP0_and_specs} \cite{linear_opt_book}. The condition is characterized in terms of the rewards vector $R = [R(s,a)], s\in S, a\in A(s)$. First, we introduce the following definition.
%
\begin{definition}[Cone of feasible directions]
\label{def:cone}
    The cone $V(x,y)$, where $(x,y)$ is any feasible solution to LP \eqref{eq:LP0_and_specs}, is defined as
\begin{equation}
V(x,y) :=
\begin{Bmatrix}
\begin{array}{lr}
v = (h,z)\in\mathbb{R}^{2|S||A|}: & \text{ } \\ 
\sum_{a\in A(s)} h_{sa} = \sum_{s'\in S}\sum_{a\in A(s')} h_{s'a} T(s|s',a),  & \text{  } \forall s\in S, \\ 
\sum_{a\in A(s)}(h_{sa}+z_{sa}) = \sum_{s'\in S} z_{s'a}T(s|s',a), & \text{  } \forall s\in S, \\
\sum_{s\in L_i}\sum_{a\in A(s)} h_{sa}\leq 0, & \text{  } i\in u(x),\\
\sum_{s\in L_j}\sum_{a\in A(s)} h_{sa}\geq 0, & \text{  } j\in l(x),\\
h_{fa} = 0, & \text{  } \forall f\in\transientset{}, a\in A(f),\\
h_{sa} \geq 0, & \text{  } \forall (s,a)\in n(x),\\
z_{sa}\geq 0, & \text{  } \forall (s,a)\in m(y)
\end{array}
\end{Bmatrix}
\label{eq:cone}
\end{equation}
where $u(x) := \{i:\sum_{L_i}\sum_{a\in A(s)} x_{sa} = u_i\}, l(x) := \{j:\sum_{L_j}\sum_{a\in A(s)} x_{sa} = l_j\}, n(x):=\{(s,a)\in\recurrentset{}\times A(s): x_{sa} = 0\}, m(y):=\{(s,a)\in S\times A(s): y_{sa} = 0\}$.
\end{definition}
%
%
\begin{theorem}
\label{thm:dual_cone}
Given LMDP $\calM$, let $(x,y)$ be a feasible solution of \eqref{eq:LP0_and_specs}. If $R = [R(s,a)], s\in S, a\in A$ is an interior point of the dual cone \[V^*(x,y)
 :=\{u\in\mathbb{R}^{2|S||A|} : ~\langle u,v\rangle \leq 0, ~\text{for every } v\in V(x,y)\}\:,
\] 
where $\langle , \rangle$ denotes the inner product, then the policy $\pi$ in \eqref{eq:policy_LP1} meets the specifications $\SSSpec$. Further, $\pi$ is the unique optimal policy in the class of policies for which   $\transientset{}\subseteq\transientset{\pi}$.
\end{theorem}
We remark that the policy could be outside of $\uptounichainset$, but preserves the transience (or isolation) of the states in $\transientset{}$.
While the statement of Theorem \ref{thm:dual_cone} imposes a conservative assumption on the rewards vector which may be generally hard to verify,
it opens up possibilities for further research on steady-state planning over larger sets of policies (beyond $\edgepreservingset$, $\classpreservingset$ and $\uptounichainset$ considered in this paper) -- in this case, sets of policies that preserve the transience of $\transientset{}$. Ultimately, one would hope to tackle \ac{name}$(\Pi)$ for arbitrary sets of stationary policies $\Pi$. These are directions for future investigation. 

\subsection{Transient Specifications}
\label{sec:trans_specs}

In Definition~\ref{def:spec}, we introduced specifications on the steady-state distribution.
However, such specifications are only useful in the recurrent sets where states are visited infinitely often.
A transient state $f \in \transientset{\pi}$ on the other hand will only be visited a finite number of times, i.e., $\prbetapi(f) = 0$ for any stationary policy $\pi\in\stationaryset$.
In this section, we present an alternative specification type which can be applied to transient states.

We first describe a suitable property of transient states against which specifications can be applied. We then define transient specifications based on this property.
\begin{definition}[Expected number of visits \cite{kemeny1963markov}]
Given an MDP $\calM$ and policy $\pi \in \Pi_S$, the expected total number of times that state $f \in \transientset{\pi}$ is visited under policy $\pi$ is
\begin{align}
\TransExpect_{\pi}(f) = \beta_{\transientset{\pi}}^{T} (I-Z_\pi)^{-1} e_{f}\:.
\end{align}
\label{def:TransExpect}
\end{definition}
\begin{definition}[Transient specification]
Given an MDP and a set of labels \sloppy $L = \{L_1, \ldots, L_{n_L}\}$, where $L_i\subseteq \transientset{}$, a set of transient specifications is given by $\TransSpec{L} = \{(L_i, [l_i, u_i])\}_{i = 1}^{n_L}$. Given a policy $\pi$, the specification $(L_i, [l_i, u_i]) \in \TransSpec{L}$ is satisfied if and only if $\sum_{f \in L_i} \TransExpect_{\pi}(f) \in [l_i, u_i]$; that is, if the expected number of visits to transient states $f \in L_i$ in the Markov chain $\calM_\pi$ falls within the interval $[l_i, u_i]$. 
	\label{def:spec_ys}
\end{definition}

Suppose that we have a set of labels $\translabelset$ over transient states, and a set of transient specifications $\TransSpec{\translabelset}$.
We can augment the LMDP found in Definition~\ref{def:spec} to incorporate these transient specifications as follows.
Let $L^\infty$ be the set of steady-state labels, and let $\Phi^{\infty}_{L^\infty}$ be corresponding steady-state specifications.
We define a complete set of labels $L = (L^\infty,\translabelset)$ and specifications $\LMDPSpec = (\Phi^{\infty}_{L^\infty}, \Phi^{tr}_{L^{tr}})$, and define an LMDP as $\calM = (S,A,T,R,\beta,L,\LMDPSpec)$.
{\color{black} Our next result regarding $y$ and $\TransExpect_{\pi}$ for the transient states gives a sufficient condition for the policy to meet the transient specification.}  
\begin{proposition}
Given an MDP $\calM$, let $(x,y)\in Q_0$ and $\pi$ as in \eqref{eq:policy_LP1}, where $Q_0$ is defined in \eqref{eq:Q0}.
If $\pi\in\uptounichainset$, then $y_f = \TransExpect_{\pi}(f)$ for any state $f \in \transientset{}$.
\label{thm:unichainpol}
\end{proposition}
We remark that this is analogous to Lemma~\ref{thm:mainCPU} stated in the appendix, which establishes that if we have a point $(x,y)\in Q_0$ for which \eqref{eq:policy_LP1} yields a CPU policy, then $\prbetapi = x$.

Given this characterization, we can augment $\mathrm{LP}_1$ \eqref{eqn:LP_EP} and $\mathrm{LP}_2$ \eqref{eqn:LP_Flow} with constraint \ref{LP_Transspecs} to synthesize policies subject to transient specifications.
\begin{align}
	\LPmainconstraintlabel{LP_Transspecs}
   	\quad &l_i \le \sum_{s \in L_i}\sum_{a \in A(s)}  y_{sa} \le u_i,
   	~ \forall (L_i, [l_i,u_i]) \in \TransSpec{L^{tr}}
\end{align}

\section{Numerical Results}
\label{sec:expresults}
In this section, we present a set of numerical results to corroborate the findings of the theoretical analysis. In Section \ref{subsec:exp_SSPEC}, we verify the steady-state behavior of the policies derived from the proposed \ac{LPs} using the Frozen Islands example of Figure \ref{fig:example}, followed by a study of their behavior in the presence of additional transient specifications in Section \ref{sec:exp_TRSPEC}. In Section \ref{sec:exp_empirical}, we present the results of a study which shows that the empirical steady-state distributions and average number of state visitations induced by the derived policies converge to values that meet the desired specifications. In Section \ref{sec:exp_compare_policies}, we evaluate the average reward achieved by said policies and examine the impact of various restrictions in their respective \ac{LPs} on the optimal values of the objective using the Toll Collector example of Figure \ref{fig:tax collector}. A case study is also presented featuring the progress of the iterative Algorithm \ref{alg:CPU} for generating a CPU policy. A natural generalization of the specifications to the product space of state-action pairs is presented in Section \ref{sec:exp_product_space}. We present two numerical experiments to support the theoretical findings of Section \ref{sec:open_set} in Section \ref{subsec:exist_issues}.
Finally, we examine the scalability of the proposed formulations in Section \ref{sec:exp_runtime}, where we present the runtime results for problems with increasing size conducted in various environments.           

\subsection{Steady-State Specifications}
\label{subsec:exp_SSPEC}
\begin{figure}[h]
    \centering
    \includegraphics{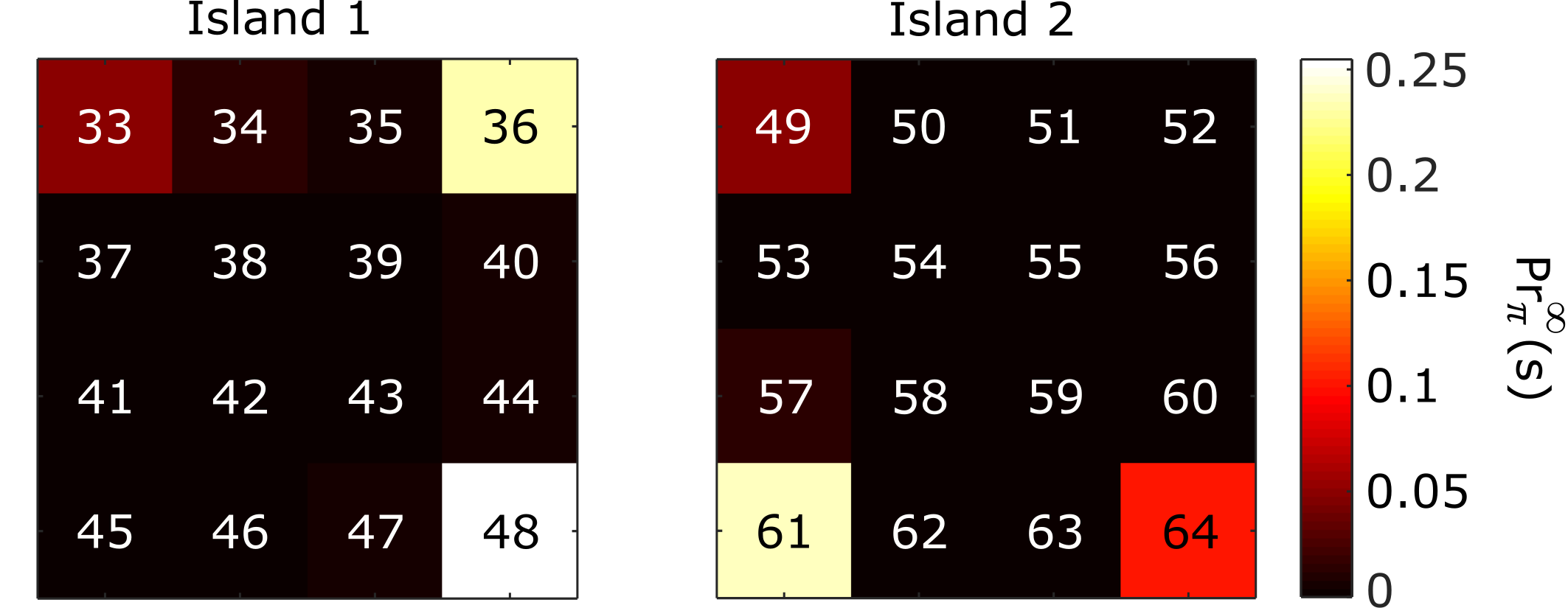}
    \caption{Heat maps showing the steady-state probabilities $\prbetapi(s)$ for states $s \in \recurrentset{}$ belonging to the two TSCCs of the Frozen Lakes example in Figure \ref{fig:example}.}
    \label{fig:Heatmap}
\end{figure}

In this section, we demonstrate the correct-by-construction behavior of the policies proposed. As an illustrative example, we first examine the behavior of a policy in $\edgepreservingset$ using the Frozen Island example shown in Figure~\ref{fig:example}. We run our proposed $\mathrm{LP}_1$ \eqref{eqn:LP_EP} to calculate the steady-state distribution $\prbetapi(s)$, and show the values for the two TSCCs (the two small islands) in Figure~\ref{fig:Heatmap}.

The heat map gives insight into the means by which the agent satisfies the specifications.
After the agent enters an island, it spends a large amount of time in states $s_{33}$, $s_{36}$, $s_{48}$, $s_{49}$, $s_{61}$, and $s_{64}$, in the sense of asymptotic frequency of visits as given by $\prbetapi(s)$. The agent also frequently visits states $s_{36}$ and $s_{61}$ to satisfy the steady-state specifications $(L_\textrm{log1}, [0.25, 1])$ and $(L_\textrm{log2}, [0.25, 1])$, respectively.
Likewise, to meet specifications $(L_\textrm{canoe1}, [0.05, 1.0])$, $(L_\textrm{canoe2}, [0.05, 1.0])$ $(L_\textrm{fish1}, [0.1, 1.0])$, $(L_\textrm{fish2}, [0.1, 1.0])$ the agent often visits states $s_{33}$, $s_{49}$, $s_{48}$, and $s_{64}$, respectively.
In addition to visiting the aforementioned states to satisfy the constraints, the agent also visits state $s_{48}$ over 25\% of the time to maximize its expected reward (recall that $R(\cdot, \cdot, s_{48}) = R(\cdot, \cdot, s_{64}) = 1$).

%
\begin{figure}[h]
    \centering
    \includegraphics[width = 15.5cm]{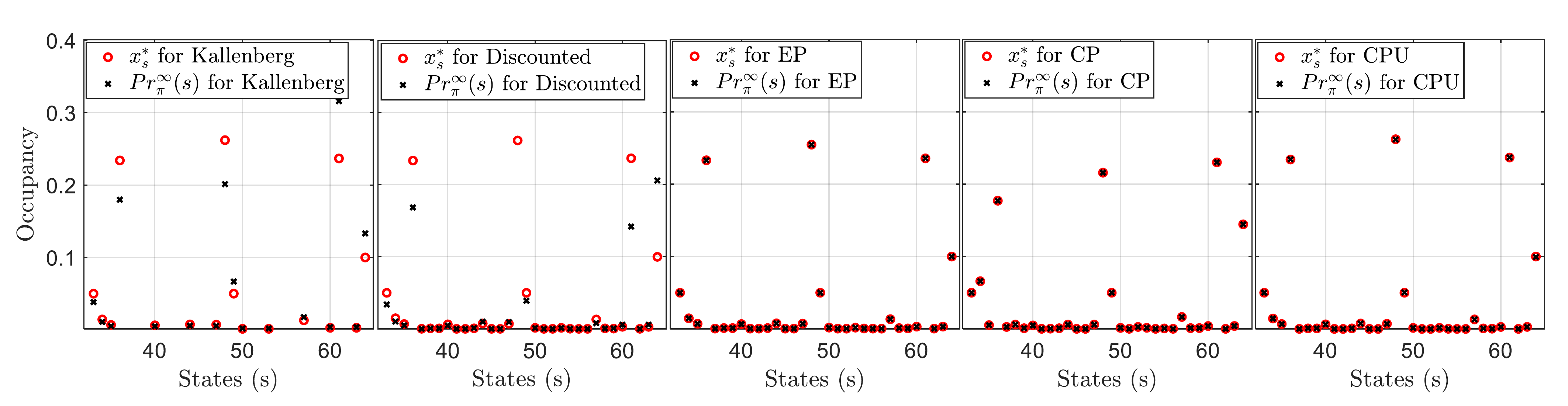}
    \caption{Example showing that $\prbetapi(s) = x^*_s, s \in \recurrentset{}$ for policies in $\edgepreservingset, \classpreservingset$ and $\uptounichainset$ derived from the proposed $\mathrm{LP}_1, \mathrm{LP}_2$ and Algorithm \ref{alg:CPU}, but not for Kallenberg's and the discounted case (discount factor $\gamma=0.9999$).}
    \label{fig:x_Prinf}
\end{figure}

The right three plots of Figure~\ref{fig:x_Prinf} show the values of $\prbetapi(s)$ along with the optimal values $x^*_{s}$ obtained from $\mathrm{LP}_1$ \eqref{eqn:LP_EP} for \ac{name}$(\edgepreservingset)$, $\mathrm{LP}_2$ for \ac{name}$(\classpreservingset)$, and Algorithm \ref{alg:CPU} for \ac{name}$(\uptounichainset)$. In each of these, the steady-state distribution matches the one estimated by the LP for every state.
This in fact holds for all state-action pairs as well, i.e., $\ssdist \!=\! x^*$.
This condition is essential to the proof of Theorems \ref{thm:mainEP}, \ref{thm:mainCP} and \ref{thm:CPU} and ensures that the policy is both optimal and satisfies the steady-state specifications.
We calculate the policy corresponding to the optimal solution of LP (4.7.6) by \citeA{kallenberg1983linear} given in \eqref{eqn:LPmultichain} with the additional specification constraints for comparison. As shown in Figure \ref{fig:x_Prinf} (\textit{left}), the derived policy fails  
to give a steady-state distribution equal to $x^*$. In addition, 
we obtain a policy from the solution to LP (3.5) by  \citeA{altman1999constrained} of a discounted reward MDP  (with the additional specification constraints)
using a discount factor $\gamma = 0.9999$. As observed in the second from the left plot of Figure \ref{fig:x_Prinf}, the steady-state distribution of the derived policy does not match $x^*$. 
\begin{table*}[!t]
	\resizebox{\textwidth}{!}{%
		\begin{tabular}{c rr rr rr rr rr rr rr}
			\toprule
			\multirow{4}{*}{\textbf{Method}} & \multicolumn{12}{c}{\textbf{SS Specifications}} & \multicolumn{2}{c}{\textbf{Rewards}} \\ 
			\cmidrule(lr){2-13} \cmidrule(lr){14-15}
			& \multicolumn{4}{c}{\textbf{Logs ($\mathbf{\ge 0.25}$)}} & \multicolumn{4}{c}{\textbf{Canoes ($\mathbf{\ge 0.05}$)}} & \multicolumn{4}{c}{\textbf{Fish Rods ($\mathbf{\ge 0.1}$)}} & \multirow{4}{*}{$\mathbf{R^*}$} & \multirow{4}{*}{$\mathbf{R^\infty_\pi}$} \\ 
			\cmidrule(lr){2-5} \cmidrule(lr){6-9} \cmidrule(lr){10-13}
			& \multicolumn{2}{c}{\textbf{Island 1}} & \multicolumn{2}{c}{\textbf{Island 2}} & \multicolumn{2}{c}{\textbf{Island 1}} & \multicolumn{2}{c}{\textbf{Island 2}} & \multicolumn{2}{c}{\textbf{Island 1}} & \multicolumn{2}{c}{\textbf{Island 2}} & & \\ 
			\cmidrule(lr){2-3} \cmidrule(lr){4-5} \cmidrule(lr){6-7} \cmidrule(lr){8-9} \cmidrule(lr){10-11} \cmidrule(lr){12-13}
			& $\boldsymbol{x^*}$ & $\ssdistbf$
			& $\boldsymbol{x^*}$ & $\ssdistbf$
			& $\boldsymbol{x^*}$ & $\ssdistbf$
			& $\boldsymbol{x^*}$ & $\ssdistbf$
			& $\boldsymbol{x^*}$ & $\ssdistbf$
			& $\boldsymbol{x^*}$ & $\ssdistbf$
			& &
			\\
			\midrule
			\multicolumn{1}{c}{CPU}
			& 0.25 & 0.25
			& 0.25 & 0.25
			& 0.05 & 0.05
			& 0.05 & 0.05
			& 0.26 & 0.26
			& 0.10 & 0.10
			& 0.3621 & 0.3621 \\
			\multicolumn{1}{c}{CP}
			& 0.26 & 0.26
			& 0.25 & 0.25
			& 0.05 & 0.05
			& 0.05 & 0.05
			& 0.21 & 0.21
			& 0.14 & 0.14
			& 0.3605 & 0.3605 \\
			\multicolumn{1}{c}{EP}
			& 0.25 & 0.25
			& 0.25 & 0.25
			& 0.05 & 0.05
			& 0.05 & 0.05
			& 0.25 & 0.25
			& 0.10 & 0.10
			& 0.3547 & 0.3547 \\
			\multicolumn{1}{c}{Kallenberg}
			& 0.25 & \textbf{\tred{0.17}}
			& 0.25 & 0.36
			& 0.05 & \textbf{\tred{0.04}}
			& 0.05 & 0.07
			& 0.26 & 0.19
			& 0.10 & 0.14  
			& 0.3621 & \textbf{0.3278}  \\
			\multicolumn{1}{c}{Discounted ($\gamma=0.999$)}
			& 0.25 & \textbf{\tred{0.04}}
			& 0.25 & \textbf{\tred{0}}
			& 0.05 & \textbf{\tred{0.0037}}
			& 0.05 & \textbf{\tred{0.0013}}
			& 0.26 & 0.52
			& 0.10 & 0.39  
			& 0.3576 & \textbf{0.9061}  \\
			\multicolumn{1}{c}{Discounted ($\gamma=0.9999$)}
			& 0.25 & \textbf{\tred{0.18}}
			& 0.25 & \textbf{\tred{0.15}}
			& 0.05 & \textbf{\tred{0.03}}
			& 0.05 & \textbf{\tred{0.04}}
			& 0.26 & 0.35
			& 0.10 & 0.21  
			& 0.3617 & \textbf{0.5530}  \\
			\bottomrule
		\end{tabular}%
	}
	\caption{Steady-state specification comparison. Bold red text indicates violated steady-state specifications. Constraints are specified in the header for each label type.}
	\label{tbl_comparison}
\end{table*}
In Table \ref{tbl_comparison}, we show the ramifications when $\prbetapi \!\ne\! x^*$.
For each specification $(L_i, [l_i, u_i]) \in \SSSpec$, Table \ref{tbl_comparison} shows the values of $e^\top x^*_{L_i} := \sum_{s \in L_i} x^*_s$ and $\prbetapi(L_i):=\sum_{s \in L_i} \prbetapi(s)$, demonstrating that all of the specifications are met for the proposed methods.
For Kallenberg's and the discounted formulations, however, although $x^*_{L_\textrm{log1}}$ and $x^*_{L_\textrm{canoe1}}$ satisfy the specification, the policy yields steady-state distributions $\prbetapi(L_\textrm{log1})$ and $\prbetapi(L_\textrm{canoe1})$ which violate the specifications (these violations are highlighted with bold red text). In other words, $e^\top x^*_{L_\textrm{canoe1}} \neq \prbetapi(L_\textrm{canoe1})$ and $e^\top x^*_{L_\textrm{log1}} \neq \prbetapi(L_\textrm{log1})$ for the Kallenberg and discounted formulations.
The table also shows the optimal reward $R^*$ given by our proposed methods, as well as the expected average reward yielded by the policy, i.e., $R^\infty_\pi := \sum_{s \in S} \sum_{a \in A(s)} \prbetapi(s,a) R(s,a)$.
While $R^*$ obtained by Kallenberg's formulation is larger than that of the EP and CP methods, the proposed LPs produce policies which yield larger values of $R^\infty$. Additionally, as the discount factor $\gamma$ approaches $1$, the discounted reward does not converge to the expected reward. The policies obtained from the discounted reward formulation achieve larger rewards $R^\infty$ by violating the steady-state constraints and spending larger proportions of time in the rewarding fishing sites.   

\subsection{Synthesis for Transient Specifications}
\label{sec:exp_TRSPEC}
\begin{figure}[h]
    \centering
    \includegraphics[trim = 0cm 0.7cm 0cm 0cm, clip, width = 15.5cm]{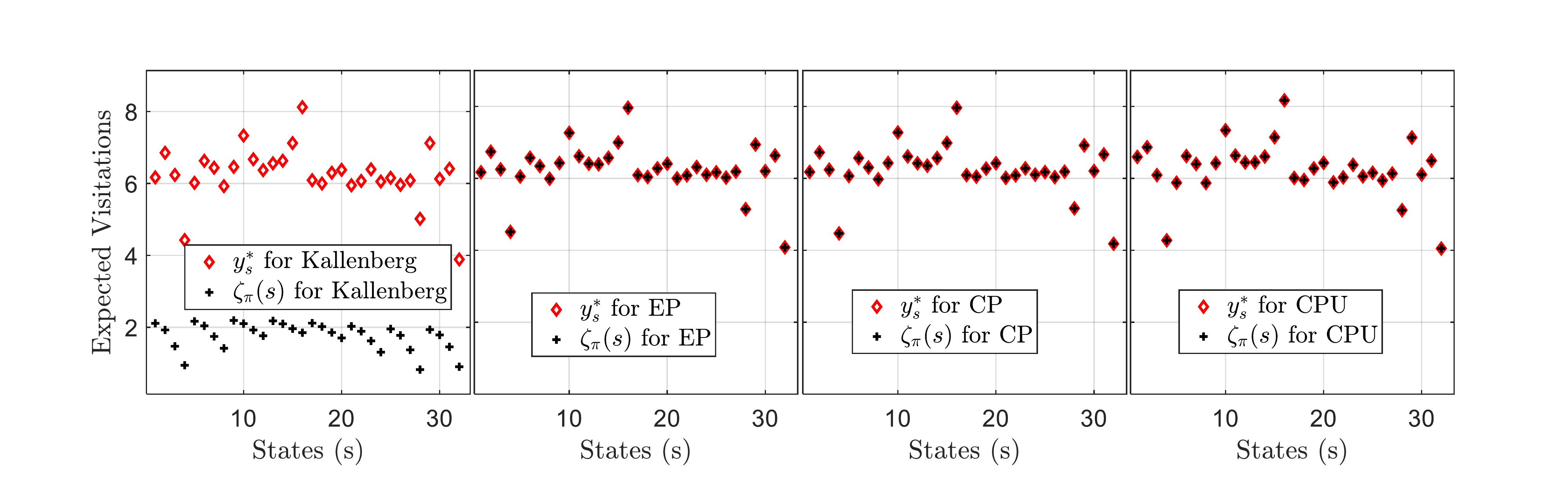}
    \caption{Example showing that $\zeta_\pi(s) = y^*_s, s \in \transientset{}$ for the proposed methods, but not for Kallenberg's formulation. }
    \label{fig:y_Prinf}
    \includegraphics[scale=0.4]{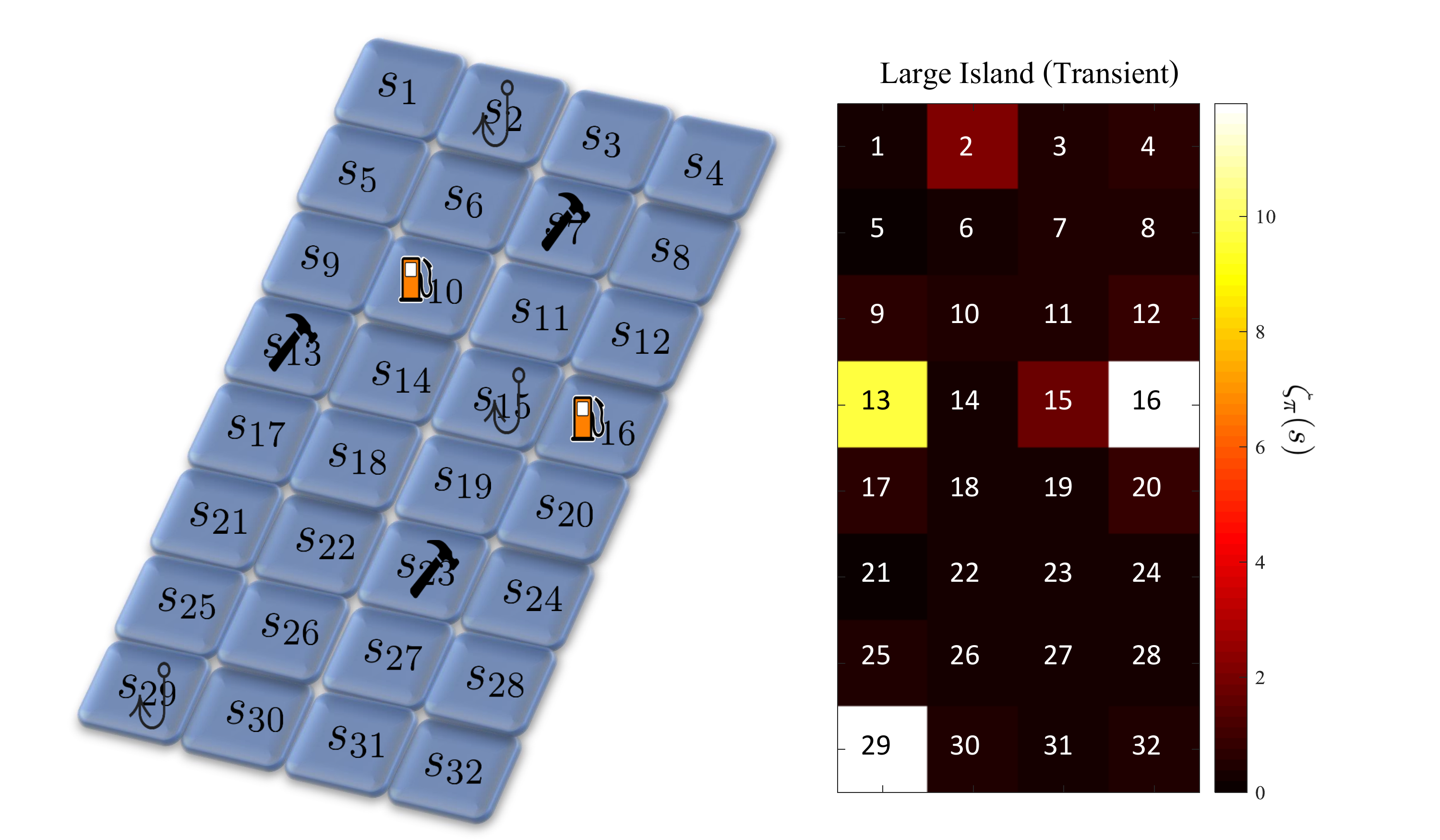}
    \caption{Distribution of labels on the large island, $L_\textrm{tools}=\{s_7,s_{13},s_{23}\}$, $L_\textrm{gas}=\{s_{10},s_{16}\}$, and $L_\textrm{supplies}=\{s_{2},s_{15},s_{29}\}$  (\textit{left}). Heat map showing the expected number of visits $\TransExpect_\pi(s)$ for states $s \in \transientset{}$, i.e., the states belonging to the large island in Figure \ref{fig:example} (\textit{left}). 
    }
    \label{fig:HeatmapWithYConstraints}
\end{figure}

In this section, we demonstrate the behavior of the policies derived subject to transient specifications following the framework described in Section \ref{sec:trans_specs}. We again compare the policies derived from our proposed formulations to that of Kallenberg with regard to meeting such specifications for the Frozen Islands example of Figure~\ref{fig:example}. The labels over states in $\transientset{}$ are set to $L_\textrm{tools}=\{s_7,s_{13},s_{23}\}$, $L_\textrm{gas}=\{s_{10},s_{16}\}$, and $L_\textrm{supplies}=\{s_{2},s_{15},s_{29}\}$ as shown in Figure~\ref{fig:HeatmapWithYConstraints} (\textit{left}). The agent sets out to collect some tools, fill up enough gas, and pick up the required fishing supplies before transitioning to one of the smaller islands which correspond to TSCCs. This is reflected in the transient specifications $(L_\textrm{tools}, [10, N_{\text{tr}}])$, $(L_\textrm{gas}, [12, N_{\text{tr}}])$, $(L_\textrm{supplies}, [15, N_{\text{tr}}]) \in \TransSpec{L}$. These specifications bound the expected total number of visitations to certain states in $\transientset{}$, where $N_{\text{tr}} = 200$. Figure~\ref{fig:y_Prinf} presents the values of the expected total number of times a state $s\in\transientset{}$ is visited under policy $\pi$, denoted by $\zeta_\pi(s)$, along with the optimal values $y^*_{s}$, obtained from Kallenberg's LP, $\mathrm{LP}_1$ \eqref{eqn:LP_EP}, $\mathrm{LP}_2$ \eqref{eqn:LP_Flow}, and  Algorithm \ref{alg:CPU}. As shown, the results match the expected number of visitations for the proposed methods for every state. 

For each transient specification $(L_i, [l_i, u_i]) \in \TransSpec{L}$, Table \ref{tbl_comparison_transient} shows $e^\top y^*_{L_i} := \sum_{f \in L_i} y^*_f$ and the expected total number of visitations achieved by the policy in corresponding states $\TransExpect_{\pi}(L_i) := \sum_{f\in L_i} \TransExpect_{\pi}(f)$. As shown, $\mathrm{LP}_1$, $\mathrm{LP}_2$ and Algorithm \ref{alg:CPU} yield policies that satisfy the given specifications
, while the policy derived from the Kallenberg LP does not. The last column shows the expected total number of visitations achieved by the policy on the larger (transient) island, where $\TransExpect_{\pi}(\transientset{}) = \sum_{f\in \transientset{}} \TransExpect_{\pi}(f)$. 
%


Figure~\ref{fig:HeatmapWithYConstraints} (\textit{right}) shows a heat map for the expected number of visits to the transient states, i.e. the large island.
The policy is calculated using $\mathrm{LP}_1$ \eqref{eqn:LP_EP}.
In addition to the constraints $(L_\textrm{tools}, [10, N_{\text{tr}}])$, $(L_\textrm{gas}, [12, N_{\text{tr}}])$, and $(L_\textrm{supplies}, [15, N_{\text{tr}}])$, we also add the constraint $(\transientset{} \setminus \left(L_\textrm{tools} \cup L_\textrm{gas} \cup L_\textrm{supplies}\right), [0, 10])$ to reduce the amount of time spent in transient states with no resources.
As shown, the agent meets the specifications largely by visiting states $s_{13}$, $s_{16}$, and $s_{29}$ to collect tools, gas, and supplies, respectively. 

\begin{table*}[b!]
	\centering
	\resizebox{14cm}{!}{%
		\begin{tabular}{c rr rr rr rr}
			\toprule
			\multirow{4}{*}{\textbf{Method}} & \multicolumn{6}{c}{\textbf{Transient Specifications ($N_{\textrm{tr}} = 200$)}} & \multicolumn{2}{c}{\textbf{Results}}  \\ 
			\cmidrule(lr){2-7} \cmidrule(lr){8-9}
			& \multicolumn{2}{c}{\textbf{Tools ($\mathbf{\ge 10}$)}} & \multicolumn{2}{c}{\textbf{Gas ($\mathbf{\ge 12}$)}} & \multicolumn{2}{c}{\textbf{Supplies ($\mathbf{\ge 15}$)}} & \multirow{4}{*}{$\mathbf{R^*}$} & \multirow{4}{*}{$\TransExpect_{\pi}(\transientset{})$} \\ 
			\cmidrule(lr){2-3} \cmidrule(lr){4-5} \cmidrule(lr){6-7}
			
			& $\boldsymbol{y^*}$ & $\TransExpect_{\pi}$
			& $\boldsymbol{y^*}$ & $\TransExpect_{\pi}$
			& $\boldsymbol{y^*}$ & $\TransExpect_{\pi}$
			
			& &
			\\
			\midrule
			\multicolumn{1}{c}{CPU}
			& 19.22 & 19.22
			& 15.51 & 15.51
			& 21.15 & 21.15
			& 0.3621 & 200 \\
			\multicolumn{1}{c}{CP}
			& 18.97 & 18.97
			& 15.26 & 15.26
			& 20.67 & 20.67
			& 0.3607 & 200 \\
			\multicolumn{1}{c}{EP}
			& 19.32 & 19.32
			& 15.51 & 15.51
			& 21.15 & 21.15
			& 0.3547 & 200 \\
			\multicolumn{1}{c}{Kallenberg}
			& 19.34 & \textbf{\tred{5.55}}
			& 15.53 & \textbf{\tred{3.95}}
			& 21.17 & \textbf{\tred{5.82}}
			& 0.3621 & \textbf{56.5}  \\
			\bottomrule
		\end{tabular}%
	}
	\caption{Bold red text indicates violated transient specifications. Constraints are specified in the header for each label type.}
	\label{tbl_comparison_transient}
\end{table*}


\subsection{Empirical Study}
\label{sec:exp_empirical}
In this section, we simulate the policies derived from our LPs to show the validity of our formulations and to further demonstrate the failure of the Kallenberg formulation to yield optimal rewards and meet specifications.

Let $S_t$ and $A_t$ denote the state and action, respectively, of the Frozen Island example at time $t$ assuming policy $\pi$ and initial distribution $\beta$. 
The average number of visits $\AvgVisitsNoarg{\pi}{n}$ and average reward $\AvgReward{\pi}{n}$ up to time $n$ are defined as

\begin{gather}
\AvgVisits{\pi}{n}{L} = \frac{1}{n} \sum_{t=1}^n \IndicatorFunction_{L} (S_t), \hspace*{12pt} \IndicatorFunction_{L}(s)=\left\{
\begin{array}{ll}
1& s \in L \\
0& s \notin L
\end{array}\right. \\ 
\AvgReward{\pi}{n} = \frac{1}{n} \sum_{t=1}^n R(S_t, A_t, S_{t+1}).
\end{gather}
We take an ensemble average over 5000 paths.

\begin{figure}
    \centering
    \includegraphics[scale=0.7]{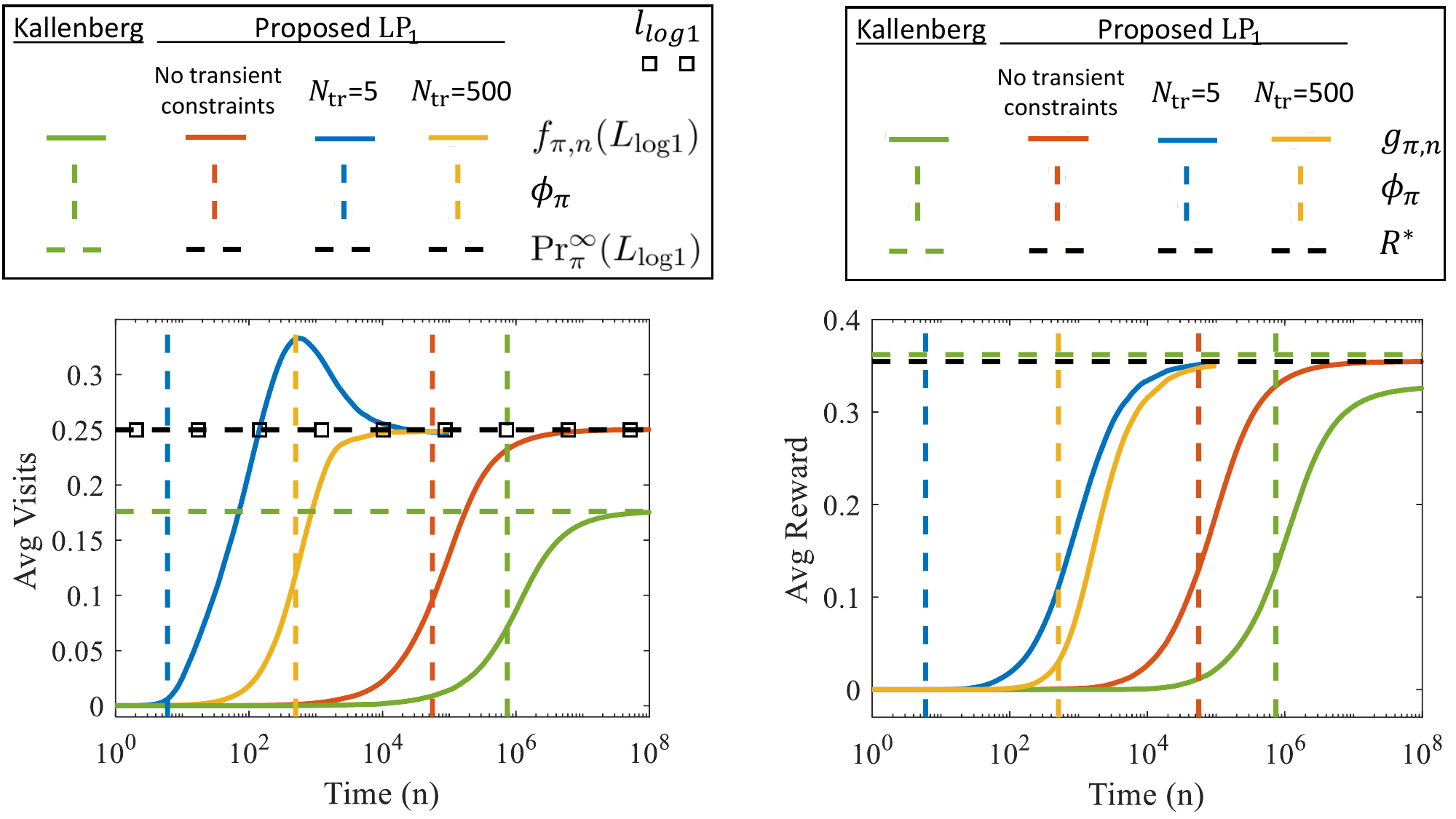}
    \caption{Execution of policy, showing (\textit{left}) average visits and (\textit{right}) average reward up to time $n$.}
    \label{fig:SimConstraints}
\end{figure}

First, we solve LP (4.7.6) of \citeA{kallenberg1983linear}.
In Figure~\ref{fig:SimConstraints} (\textit{left}), the solid green line shows the average number of visits to the states in $L_\textrm{log1} = \{s_{34}, s_{36}, s_{38}, s_{43}\}$, and the horizontal dashed green line indicates the steady-state distribution.
The square markers show the lower bound of the specification on the logs.
While the value of $\AvgVisits{\pi}{n}{L_\textrm{log1}}$ converges to the steady-state distribution, the policy fails to meet the steady-state specification. This follows from the fact that $\prbetapi \ne x^*$.

Next, we produce an EP policy by executing $\mathrm{LP}_1$ \eqref{eqn:LP_EP}, using no transient specifications.
The average number of visits to the states in $L_\textrm{log1}$ is shown as a solid red line in Figure~\ref{fig:SimConstraints} (\textit{left}).
Not only does the average number of visits converge to the corresponding steady-state distribution (dashed black line), but the specification is met.
We observe similar results for CP and CPU policies as well.

In Figure~\ref{fig:SimConstraints} (\textit{right}), the solid green and red lines show the average reward for the Kallenberg LP and our proposed $\mathrm{LP}_1$, respectively.
The dashed green line indicates $R^*$ for Kallenberg's formulation. As can be seen, for similar reasons as before, the average reward converges to a reward other than that output by the LP.
On the other hand, $\mathrm{LP}_1$ converges to the corresponding LP reward $R^*$ (black dashed line).



The vertical green and red dashed lines in Figure~\ref{fig:SimConstraints} indicate the average time of entry into $\recurrentset{}$ for $\mathrm{LP}_1$ and Kallenberg's LP, respectively, where the time of entry $\FirstRecurrent{\pi}$ is given by

\begin{gather}
\FirstRecurrent{\pi} = \min \{n \mid S_n \in \recurrentset{}\}.
\end{gather}

In both cases, the agent spends an unduly amount of time in the transient states before transitioning to a recurrent set, which may be undesirable.
To reduce the amount of time spent in the transient states, we next introduce a transient specification $(\transientset{}, [0,N_\textrm{tr}])$ and rerun $\mathrm{LP}_1$.
The constant $N_{\textrm{tr}}$ is used to control the time of entry into the recurrent sets.
The results are shown for $N_\textrm{tr} = 5$ (blue lines) and $N_\textrm{tr} = 500$ (yellow lines).
In both cases, convergence of the average visits to $\prbetapi(L_\textrm{log1})$ occurs at a much faster rate, leading to a much faster accumulation of reward.

We now comment further on the simulation for $N_\textrm{tr} = 5$.
The policy produced by $\mathrm{LP}_1$ separates the first small island into two main subsets.
The agent tends to visit state $s_{33}$ repeatedly after entering the first small island, leading to an above average number of visits to $\textrm{log1}$ states. This results in an initial ``overshoot'' of $\prbetapi(L_\textrm{log1})$.
This effect is not seen for $N_\textrm{tr} = 500$ due to the averaging effect of $\AvgVisits{\pi}{n}{L}$.
Likewise, the policy tends to delay the entry of the agent into state $s_{48}$ where rewards are accumulated.
This delay is especially noticeable in $\AvgReward{\pi}{n}$ for $N_\textrm{tr} = 5$ due to the logarithmic time scale.

\begin{figure}
    \centering
    \includegraphics[scale=0.75]{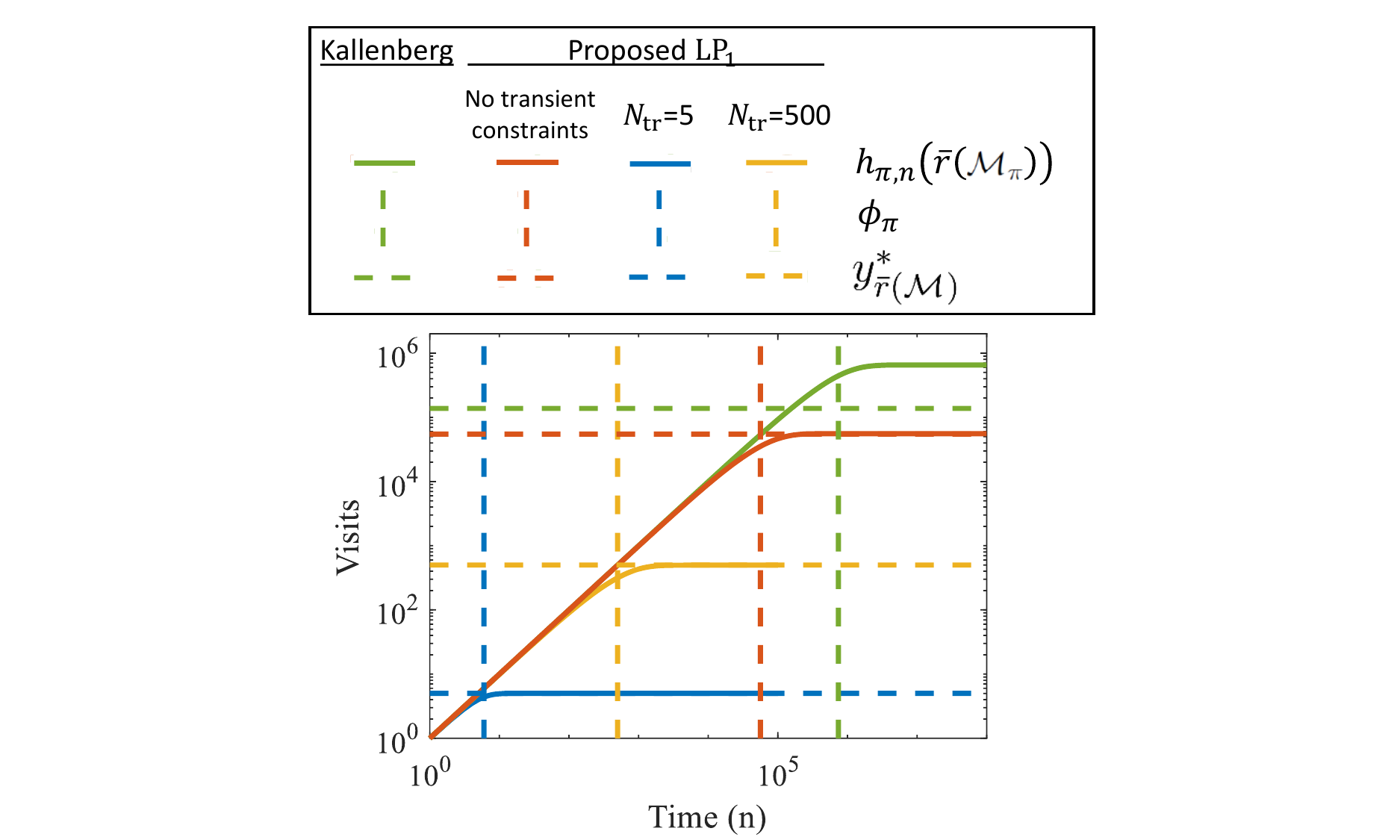}
    \caption{Execution of policy with transient specifications showing the number of visits to transient states up to time $n$.}
    \label{fig:ymin}
\end{figure}

In the same vein, we explore the simulated behavior of our policies in terms of the number of visits to transient states, where the number of visits $\VisitsNoarg{\pi}{n}(L)$ to states in $L\subseteq\transientset{}$ up to time $n$ is defined as
\begin{gather}
\Visits{\pi}{n}{L} = \sum_{t=1}^n \IndicatorFunction_{L} (S_t).
\end{gather}
In Figure~\ref{fig:ymin}, we show the number of visits to the transient states for the same policies as shown in Figure~\ref{fig:SimConstraints}.
For the policies produced by $\mathrm{LP}_1$, $\Visits{\pi}{n}{\transientset{}}$ converges to the optimal $y^*_{\transientset{}}$.
On the other hand, as described in Section~\ref{sec:exp_TRSPEC}, for the Kallenberg formulation we have $\TransExpect_{\pi} \ne y^*$ and so the derived policy fails to converge to $y^*_{\transientset{}}$.

\subsection{Comparison of Policies}
\label{sec:exp_compare_policies}
Recall that policies in $\edgepreservingset$ exercise all transitions in the TSCCs of an MDP. By contrast, policies in $\classpreservingset$ and $\uptounichainset$ are less restrictive in that they only preserve the state classification and the unichain property of these components, respectively. In turn, they often yield larger expected rewards while simultaneously satisfying desired specifications. In this section, we verify the correctness of such policies and compare their optimal rewards.

\begin{figure}[t!]
    \centering
    \includegraphics[trim = 0cm 1.25cm 0cm 0cm, clip, width = 15cm]{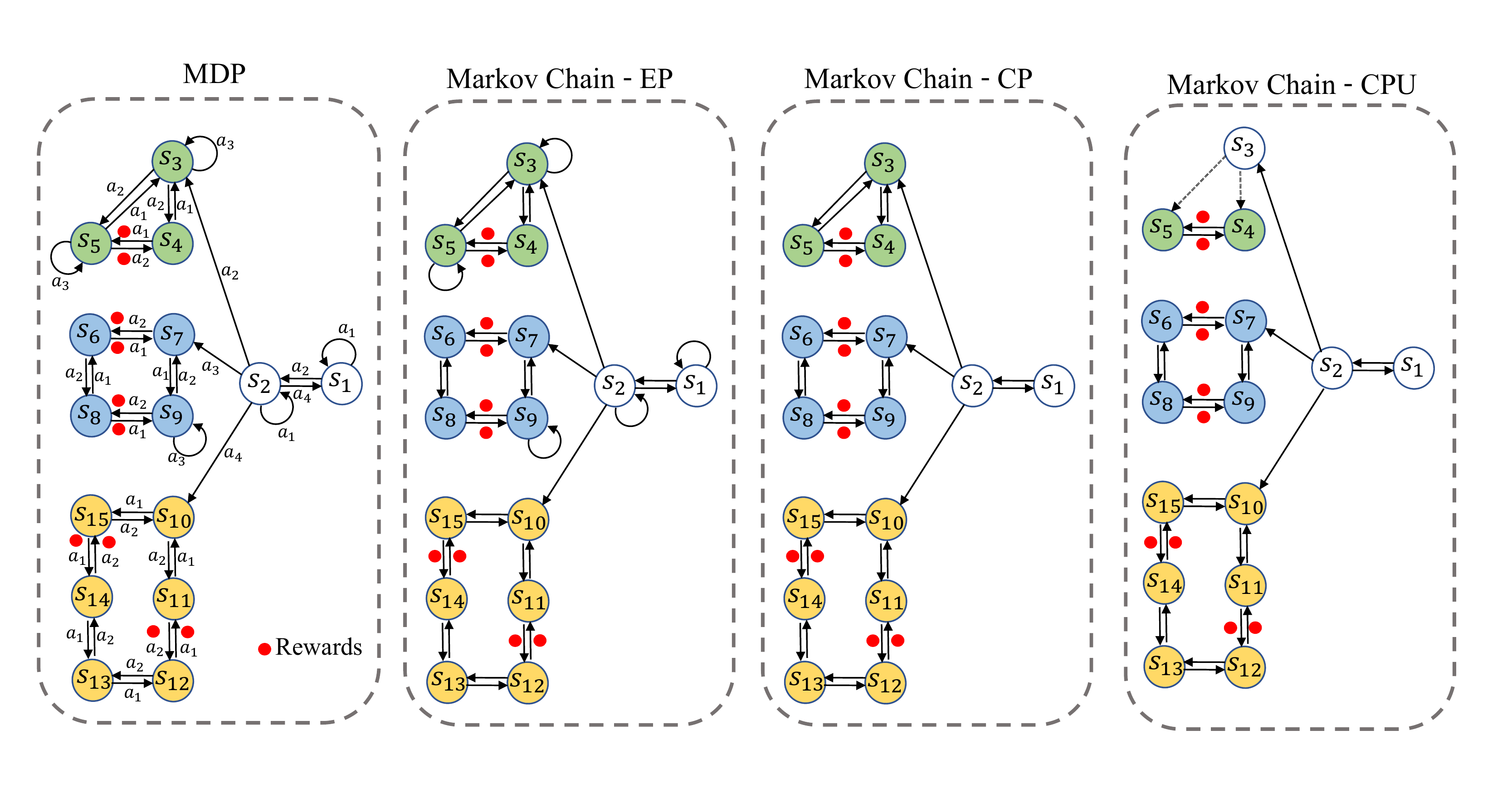}
    \caption{Markov chains induced by the EP, CP, and CPU policies. The first column shows the underlying MDP $\calM = (S,A,T,R,\beta)$. Transitions designated with circles have unit reward, otherwise the reward is $0$. The initial distribution $\beta$ is uniform over $S$.}
    \label{fig:MC_Ep_CP_CPU}
    \includegraphics[trim= 0cm 0.55cm 0cm 0cm, clip, width =14cm]{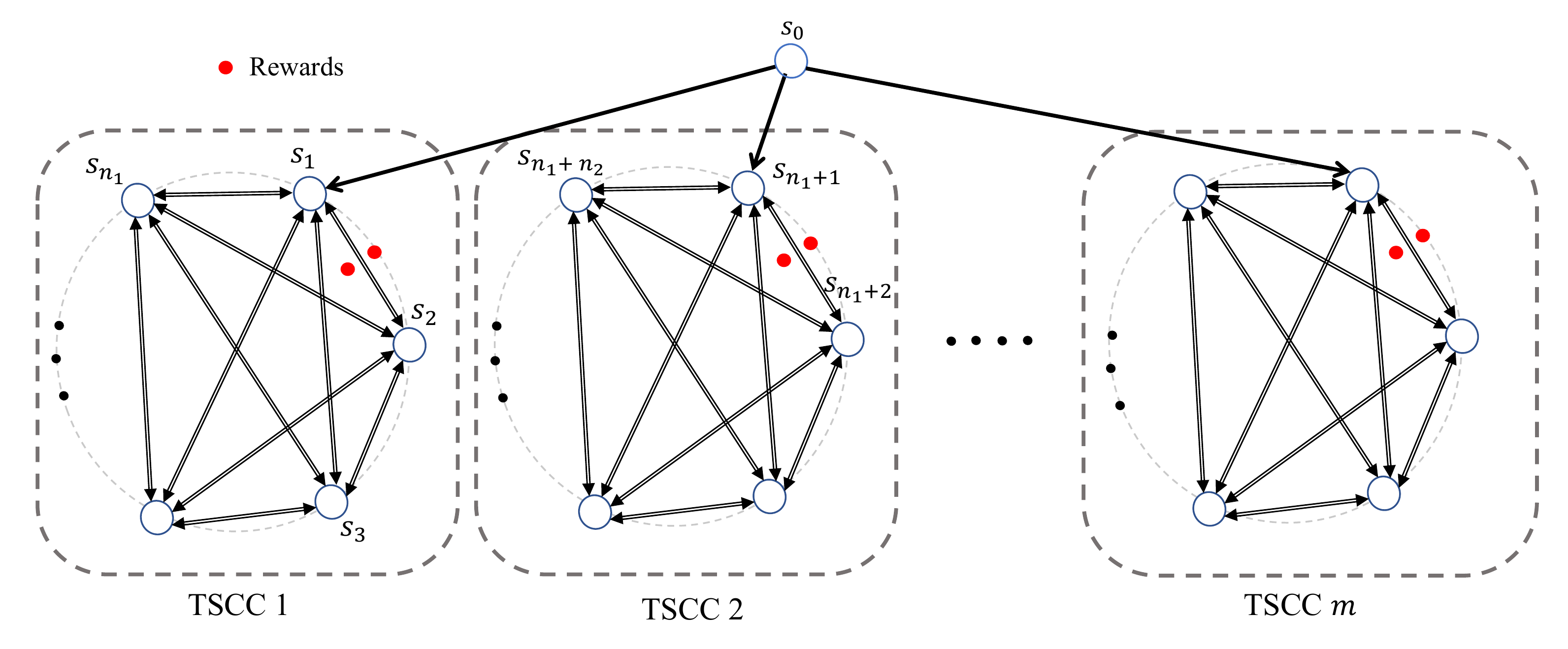}
    \caption{Toll Collector problem given by LMDP $\calM = (S,A,T,R,\beta,L,\SSSpec)$ consisting of $m$ fully-connected TSCCs $\tscc{k}{}, k\in[m]$ and $\transientset{} = \{s_0\}$. The $k$-th TSCC consists of $n_k$ states. State $s_0$ has $m$ actions, each of which leads to one of the $m$ TSCCs with probability $1$. For each state $s_i$ in $\tscc{k}{}$, there are $n_k - 1$ actions, each of which causes a transition to another state in $\tscc{k}{}$ with probability $1$. The reward function is defined such that, in each TSCC, there is a positive reward by taking the action that leads from some state $s_i$ to its neighbor $s_{i + 1}$ and vice-versa. That is, $R(s_i, \cdot, s_{i+1}) = R(s_{i+1}, \cdot, s_i) = 1$ for some $i$. These rewards are designated with red solid circles in each TSCC. All other rewards are $0$. The initial distribution $\beta$ is uniform over $S$. The labels and steady-state specifications are given by $L_k=\{s\in \tscc{k}{} : R(s,a,s')=0, \forall a\in A(s), s'\in\tscc{k}{} \}$ and $\SSSpec = (L_k, [l, 1])$, respectively, for all $k\in[\NumErgSets]$.}
    \label{fig:tax collector}
\end{figure}

Figure~\ref{fig:MC_Ep_CP_CPU} illustrates the Markov chains induced by policies in $\edgepreservingset, \classpreservingset$, and $\uptounichainset$, respectively, for the MDP shown in its first column. The policies are obtained from the optimal solutions of the corresponding LPs. 
As observed, the Markov chain induced by the EP policy (Column 2) contains all transitions in the TSCCs of the underlying MDP. The self loops of states $s_3$,$s_5$, and $s_9$ are missing in the Markov chain induced by the CP policy without affecting the recurrence of each of the TSCCs. In the case of the Markov chain induced by the CPU policy, the state $s_3$ is transient in the TSCC  $\{s_3,s_4,s_5\}$, but all TSCCs of $\calM$ remain unichain in the induced Markov chain. That is, each TSCC contains exactly one recurrent component.

In order to compare the performance of our EP, CP, and CPU policies, we define the Toll Collector example given by the LMDP $\calM$ of Figure~\ref{fig:tax collector}. 
In this problem, an agent must choose one of $m$ cities to visit, each of which corresponds to a TSCC of $\calM$. The $k$-th city consists of $n_k, k\in [m]$ counties represented as vertices and roads connecting these counties represented by edges. The roads with toll booths yield a positive reward for collecting a toll. However, the agent needs to spend some time on roads without toll booths in order to build them. 
We consider an instance of the Toll Collector problem for which $m=3$ and the number of states per TSCC $n_k = n, \forall k$. 
To highlight the gap between the optimal rewards of the different policies, we define the labels $L_k=\{s\in \tscc{k}{} : R(s,a,s')=0, \forall a\in A(s), s'\in\tscc{k}{} \}$ and $\SSSpec = (L_k, [l, 1])$, respectively, $\forall k\in[\NumErgSets]$. As such, per $\SSSpec$, the steady-state probability of states with no rewardful transitions is forced to be bounded below by $l$. We will use this steady-state specification with various values of $l$ to show that, for lower values of $l$, there is a significant gap in expected rewards observed by the various policies. As this $l$ value is increased, the gap can be shown to diminish.

\begin{figure}
     \centering
     \includegraphics[width=\textwidth]{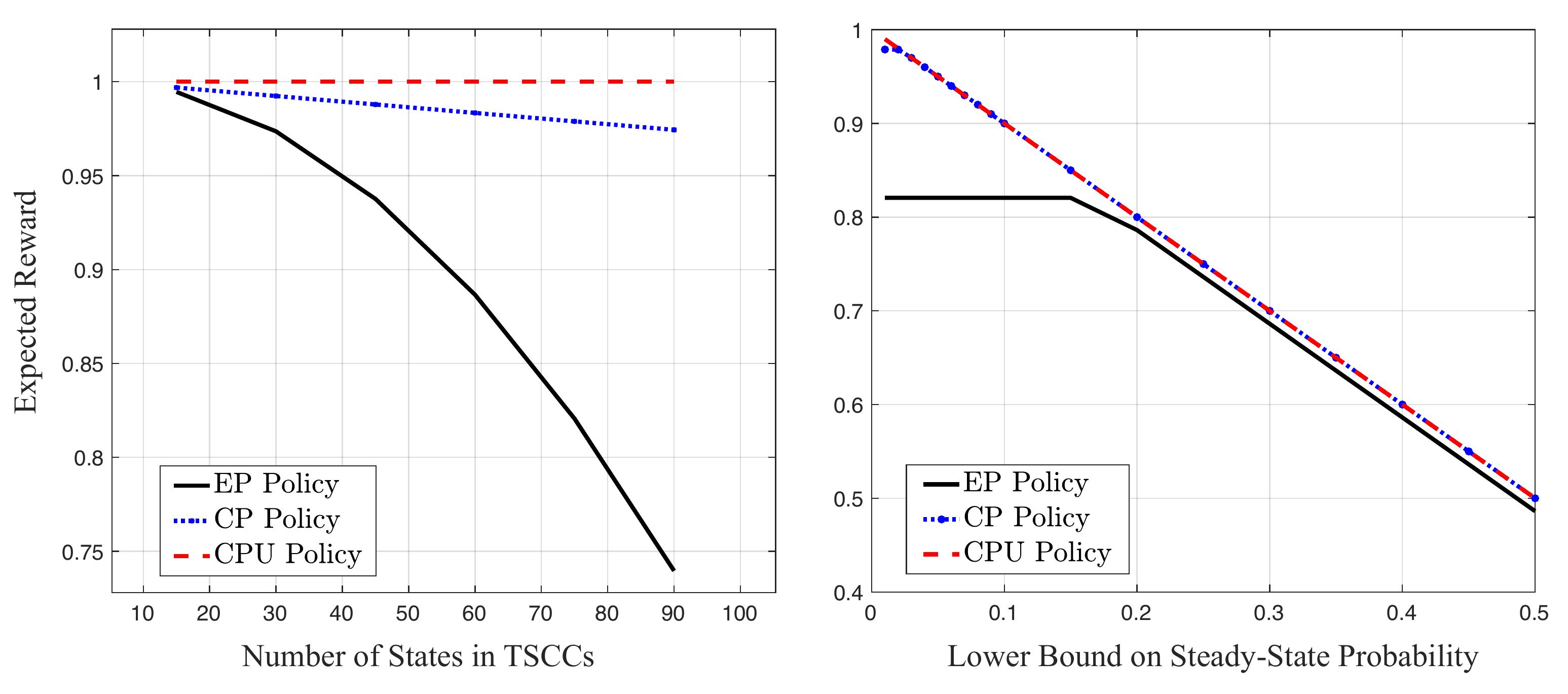}
 \caption{Comparison of EP, CP, and CPU policies for the Toll Collector example. (Left) Expected reward as function of the total number of states in each TSCC when $l = 0$. (Right) Expected reward as function of the lower bound $l$ on the steady-state probability when $n = 25$.}
     \label{fig:toll}
 \end{figure}

Figure~\ref{fig:toll} (left) compares the optimal rewards achieved by the different policies as a function of the total number of states in the TSCCs (i.e., $3 n$) when $l = 0$. As the number of states increases, the gap between the average reward of the EP policies and their CP and CPU counterparts increases. In this scenario, the EP policies incur a quadratic loss relative to CPU policies since they are forced to exercise all existing transitions equally and there are $O(n^2_k)$ such transitions in the $k$-th TSCC. On the other hand, an optimal CPU policy preserves the unichain property while exercising exclusively the two transitions with positive reward in each TSCC. A smaller loss is incurred by CP policies since they are only required to preserve the recurrence of the TSCCs and can thus restrict themselves to visiting the outer perimeter of each TSCC. In doing so, the CP policies incur a linear loss when compared to the CPU policies because they must visit every state in a TSCC infinitely often in order to preserve the recurrent classification of these states. 

Figure~\ref{fig:toll} (right) illustrates the average rewards as a function of the lower bound $l$ for the three types of policies when the number of states in each TSCC is $n=25$. 
When $l$ increases, the average reward gap between the different policies diminishes since the agent has to spend more time in states with no rewards to meet the desired specifications.  

\subsubsection{Operation of Algorithm \ref{alg:CPU}}
\begin{figure}[h]
    \centering
    \includegraphics[width = 15cm]{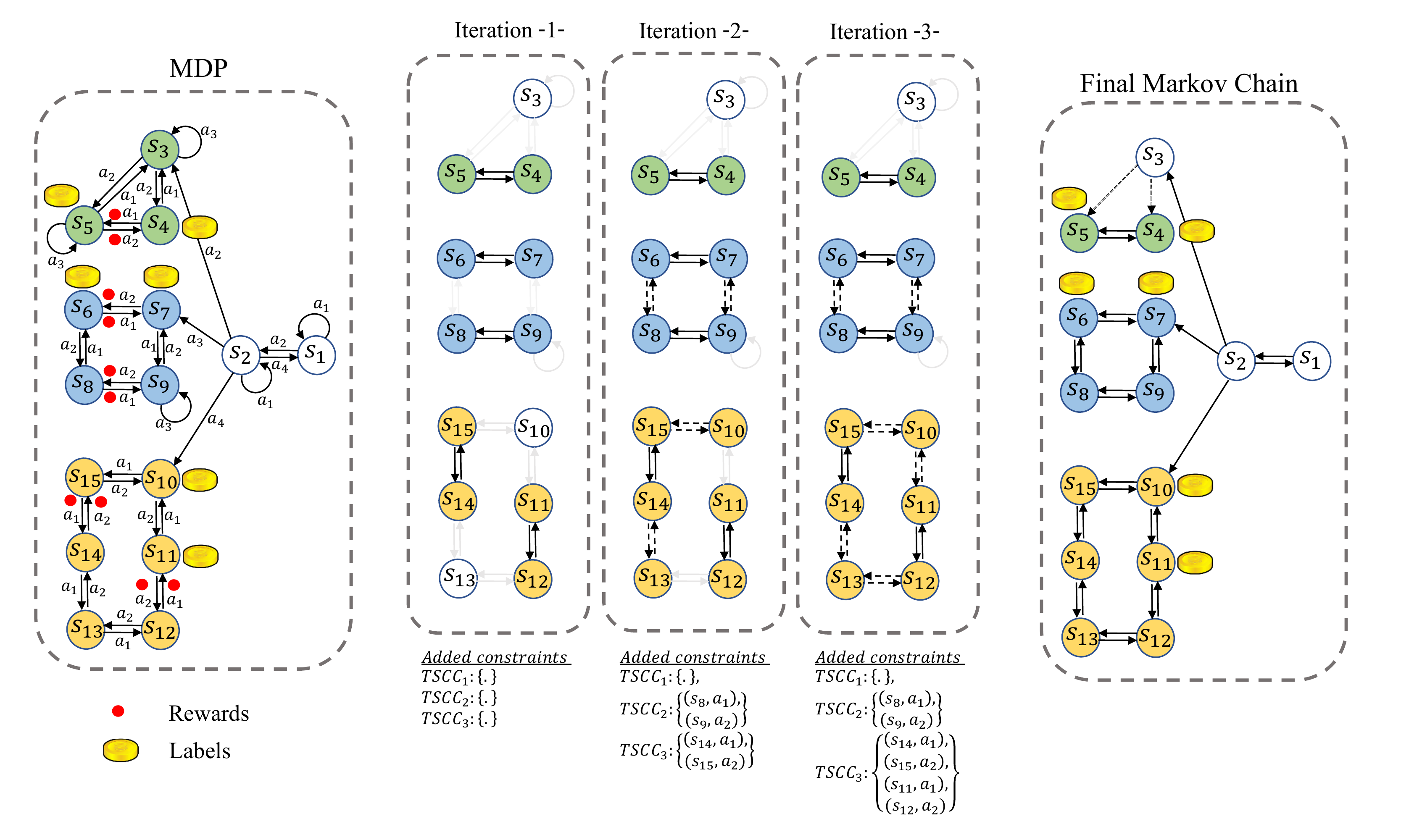}
    \caption{Illustration of the progress of Algorithm \ref{alg:CPU} for generating a policy in $\uptounichainset$. The LMDP $\calM = (S,A,T,R,\beta,L,\SSSpec)$, where $S$, $A$, and $R$ are given in the MDP (first column) and $\beta$ is uniform. labels are $L_\textrm{gold1}=\{s_4,s_5\}$, $L_\textrm{gold2}=\{s_6,s_7\}$, $L_\textrm{gold3}=\{s_{10},s_{11}\}$, and the steady-state specifications are $(L_\textrm{gold1}, [0.20, 1])$,$(L_\textrm{gold2}, [0.10, 1])$, $(L_\textrm{gold3}, [0.15, 1])$ (first column). In each iteration, we illustrate the support of the optimal solution for the TSCCs of $\calM$ (middle columns) along with the edges (state-action pairs) for a given cut. After three iterations, the support of the optimal solution corresponds to SCCs in each of the TSCCs. Every TSCC of $\calM$ is a unichain component in the Markov chain $\calM_\pi$ induced by the resulting policy (last column).}
    \label{fig:CPU Alg example}
\end{figure}

In this subsection, we present in detail the operation of the proposed Algorithm \ref{alg:CPU} to generate a CPU policy. Consider the LMDP given in Figure~\ref{fig:CPU Alg example}. For each iteration, LP~\eqref{eq:LP0_and_specs} is solved and the digraph of the support of the solution in each TSCC is shown (colored nodes). In the first iteration, for the first TSCC, states $s_4$ and $s_5$ form a SCC, while $s_3$ does not belong to the support of the solution. For both the second and third TSCCs, all respective states belong to the support but they do not form a SCC, thus we can find cut(s) (as can be seen at the bottom of the figure of the second iteration). 
In the second iteration, the dotted edges are added which results in one SCC for the second TSCC (no additional constraints are needed) but not for the third TSCC. Thus, we consider additional cuts as shown at the bottom of the figure of the third iteration. In the last iteration, the stopping criteria is met (the digraph in each of the three TSCCs is strongly connected). The final Markov chain induced by the CPU policy derived from the solution to the \ac{LP} of the third iteration is shown on the right side of Figure~\ref{fig:CPU Alg example}. States $\{s_3, s_4, s_5\}$ form a unichain component, states $\{s_6, s_7, s_8, s_9\}$ form a recurrent component, and states $\{s_{10}, s_{11}, s_{12}, s_{13}, s_{14}, s_{15}\}$ belong to a TSCC where all edges are preserved.

\subsection{Specifications on State-Action Pairs}
\label{sec:exp_product_space}
Up to this point, we have defined steady-state and transient specifications over states. However, the framework proposed can be used to synthesize policies with provably correct behavior on the level of state-action pairs as well. As an example, consider the LMDP $\calM = (S,A,T,R,\beta,L,\LMDPSpec)$ defined in Figure~\ref{fig:CPU Alg example} (\textit{left}). We define labels $L = (L^\infty,\translabelset)$ over state-action pairs, i.e., $\translabelset_{\textrm{tool}} = \{(s_{2},a_{1})\}$, and $L^\infty_{\textrm{gold}1} = \{(s_{4},a_{1})\}$, $L^\infty_{\textrm{gold}2} = \{(s_{6},a_{2})\}$, and $L^\infty_{\textrm{gold}3} = \{(s_{10},a_{1})\}$. The specifications are given as $\LMDPSpec = (\Phi^{\infty}_{L^\infty}, \Phi^{tr}_{L^{tr}})$, where the steady-state specifications are $(L^\infty_{\textrm{gold}1}, [0.10, 1])$, $(L^\infty_{\textrm{gold}2}, [0.12, 1])$, and $(L^\infty_{\textrm{gold}3}, [0.20, 1])$, and the transient specifications are given as $(\translabelset_{\textrm{tool}}, [20, 50])$. We also set the average total number of visitations $N_\textrm{tr} = 50$. Table \ref{state-action table} shows the steady-state distributions $\prbetapi(s,a)$ and the expected number of visitations $\TransExpect_{\pi}(s,a)$ for the labeled sets, as well as the total number of visitations $\TransExpect_{\pi}(\transientset{})$ to the set $\transientset{}$ for EP, CP and CPU policies. As shown, the policies meet both steady-state and transient specifications defined over the product space $S\times A$.

\begin{table*}[!t]
	\centering
	\begin{tabular}{c rrr rr}
			\toprule
			\multirow{3}{*}{\textbf{Policy}} & \multicolumn{5}{c}{\textbf{Specifications}} \\ 
			\cmidrule(lr){2-6} 
			& \multicolumn{3}{c}{\textbf{Steady-State}} & \multicolumn{2}{c}{\textbf{Transient}} \\ 
			\cmidrule(lr){2-4} \cmidrule(lr){5-6}
			
			& $\prbetapi(s_4,a_1)$ & $\prbetapi(s_6,a_2)$ & $\prbetapi(s_{10},a_1)$ & $\TransExpect_{\pi}(s_2,a_1)$ & $\TransExpect_{\pi}(\transientset{})$
			\\
			\midrule
			\multicolumn{1}{c}{EP}
			& 0.11 & 0.12 & 0.20 & 26.2 & 50   \\
			\multicolumn{1}{c}{CP}
			& 0.10 & 0.12 & 0.20 & 31.03 & 50   \\
			\multicolumn{1}{c}{CPU}
			& 0.11 & 0.12 & 0.20 & 28.82 & 50  \\
			\bottomrule
		\end{tabular}%
	\caption{The policies meet transient and steady-state specifications on state-action pairs for the MDP defined in Figure~\ref{fig:CPU Alg example}.}
	\label{state-action table}
\end{table*}

\begin{figure}[h]
    \centering
    \includegraphics[width = 15cm]{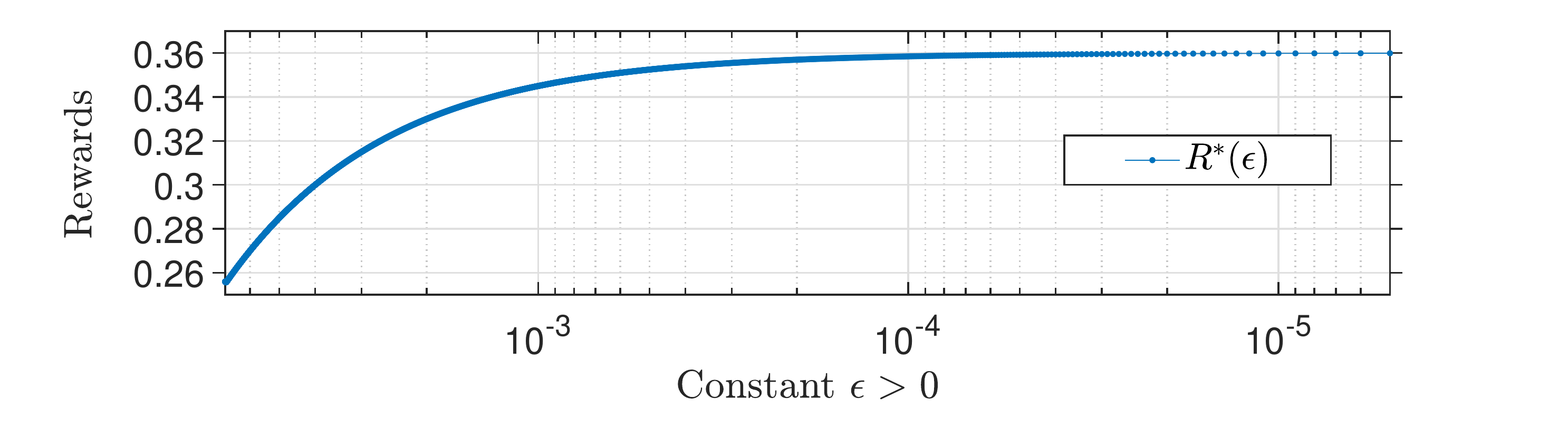}
    \caption{\textcolor{black}{Convergence of the average expected reward as we vary the parameter $\epsilon$ in \eqref{eqn:LP_EP_modified} for the MDP example of Figure~\ref{fig:CPU Alg example}.}}
    \label{fig:impact of eps example}
\end{figure}

{\color{black} \subsection{Modified LP and Policy Set}
\label{subsec:exist_issues}
\textbf{Impact of $\epsilon$ in $\mathrm{LP}_1(\epsilon)$.}
\label{sec: eps imapct}
In this section, we use the MDP example of Figure~\ref{fig:CPU Alg example} to investigate the impact of the parameter $\epsilon > 0$ on the total reward induced by an optimal EP policy. In particular, we solve $\mathrm{LP}_1(\epsilon)$ with descending values of $\epsilon$ and compute the optimal reward $R^*(\epsilon)$. As shown in Figure~\ref{fig:impact of eps example}, $R^*(\epsilon)$ increases monotonically as we decrease $\epsilon$ with diminishing return, and converges to nearly $0.36$ as $\epsilon\rightarrow 0$. For 
values of $\epsilon$ below $10^{-4}$, the change in average reward if we further decrease $\epsilon$ is insignificant. 
Therefore, in our experiments we have set $\epsilon = 10^{-4}$. }\par
{\color{black}
\noindent\textbf{Policies with bounded support.} Here, we verify the result of Theorem \ref{thm:modifiedEP}. Consider the example of Fig. \ref{fig:uni_multi_chain} where $R(s_2,a_1) = R(s_3,a_1) = R(s_3,a_2) = 0.1$ and $R(s_2,a_2) = 0.5$. Recalling that $\delta$ is a lower bound on the support of the policies in $\edgepreservingset(\delta)$ in \eqref{eq:modifiedEP}, we can show that the optimal average reward over $\edgepreservingset(\delta)$ is \sloppy $\max_{\pi\in\edgepreservingset(\delta)} R_\pi^\infty = 0.5 (1-\delta)^2 + 0.2\delta(1-\delta) + 0.1\delta^2$, achieved by the policy $\pi^*$ which has $\pi^*(s_2|a_2) = \pi^*(s_3|a_1) = 1-\delta$, and $\pi^*(a_1|s_2) = \pi^*(a_2|s_3) = \delta$. The reward $R^*(\delta)$ of the policy $\pi$ in \eqref{eq:policy_LP1} obtained from the optimal solution to $\mathrm{LP}_1(\delta)$ in \eqref{eqn:LP_EP_modified} is $R^*(\delta) = 0.5 - 1.2\delta$, 
where $\pi(a_1|s_2) = \delta/(1-2\delta)$, $\pi(a_2|s_2) = (1-\delta)/(1-2\delta)$ and $\pi(a_1|s_3) = \pi(a_2|s_3) = 1/2$. Fig. \ref{fig:thm_modifiedEP} shows that the difference $R_{\pi^*}^\infty - R^*(\delta)\rightarrow 0$ as $\delta\rightarrow 0$ as per Theorem \ref{thm:modifiedEP}.
\begin{figure}[h]
    \centering
    \includegraphics[width = 15cm]{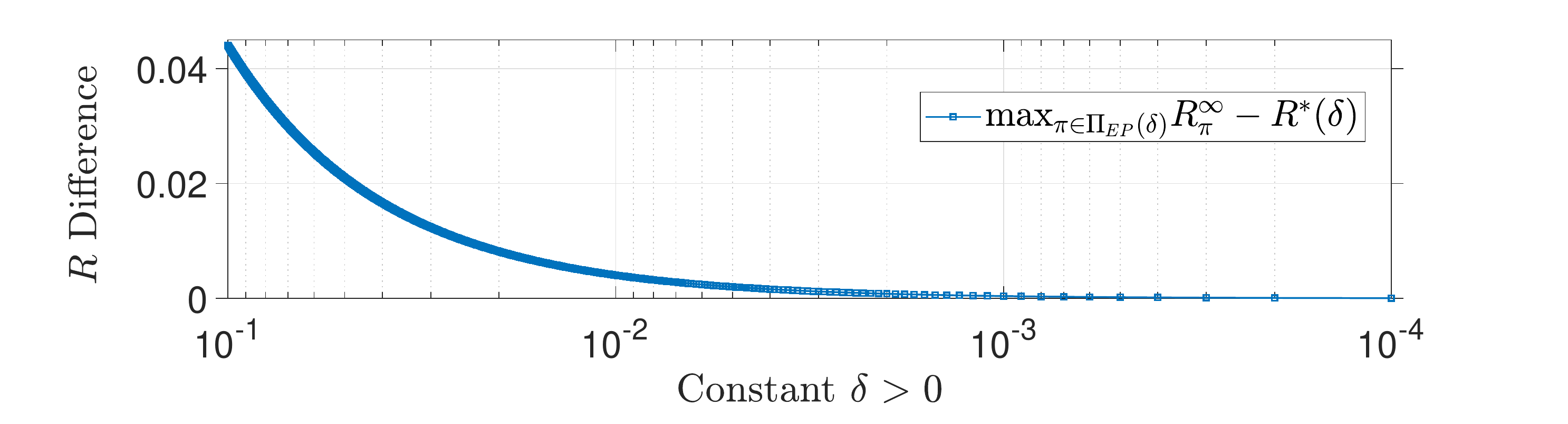}
    \caption{\textcolor{black}{The difference $R_{\pi^*}^\infty - R^*(\delta)\rightarrow 0$ as $\delta\rightarrow 0$, where $\pi^* = \argmax_{\pi\in\edgepreservingset(\delta)} R_\pi^\infty$ and $R^*(\delta)$ is the reward of the policy obtained from an optimal solution to $\mathrm{LP}_1(\delta).$}}
    \label{fig:thm_modifiedEP}
\end{figure}
}


\subsection{Scalability}
\label{sec:exp_runtime}
We demonstrate the scalability of the proposed formulations using two sets of experiments. The first set of experiments are performed on a standard desktop with 16GB of RAM using the Matlab CVX package for convex optimization \cite{cvx,gb08}. We also perform a second set of experiments on a standard desktop of 128GB of RAM using the commercial CPLEX Optimizer, which provides a higher-precision mathematical solver for large-scale linear 
programming. 

\begin{figure}[h]
    \centering
    \includegraphics[width = 11cm]{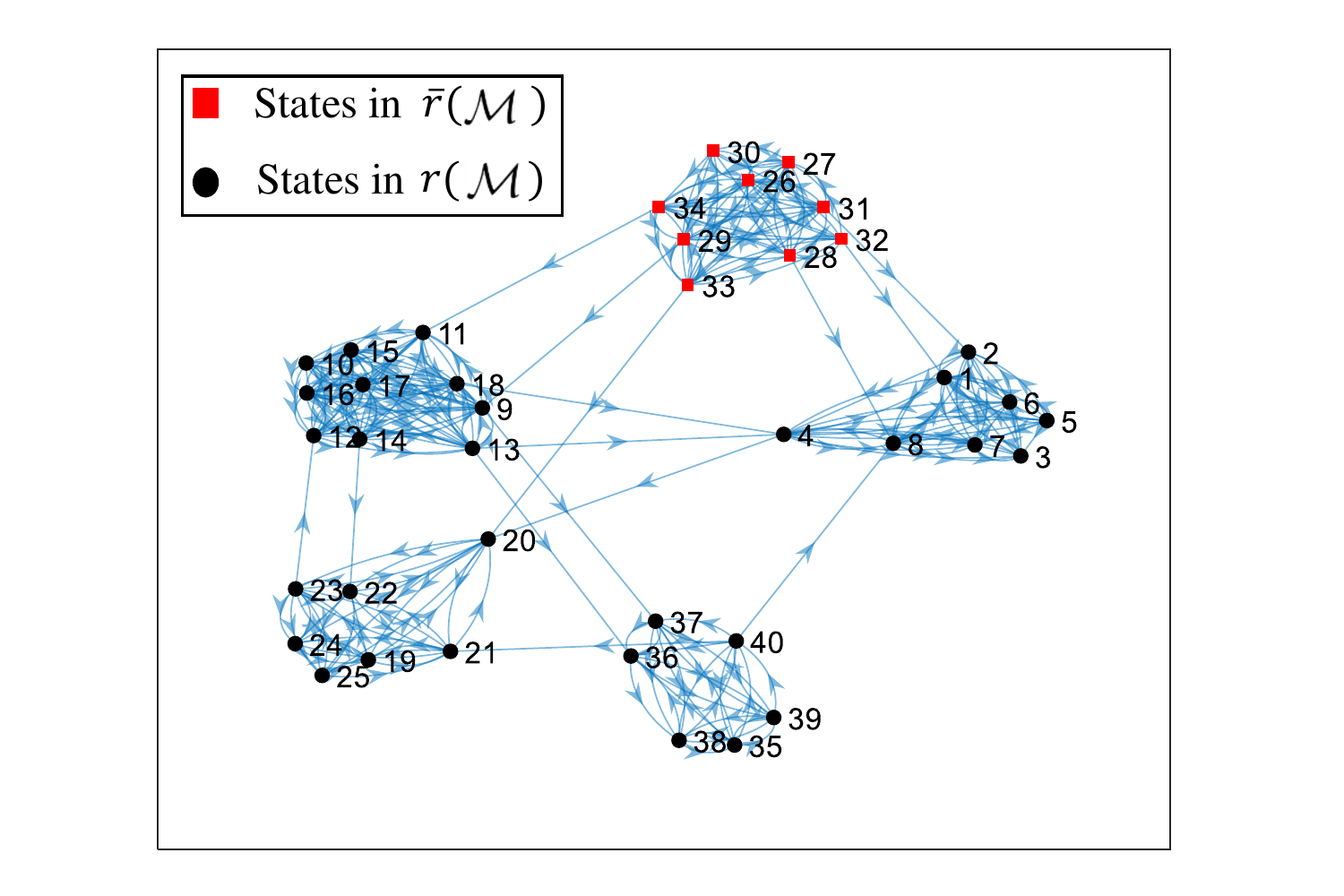}
    \caption{A NetworkX random 40-node digraph used to generate a LMDP for the third example of Table~\ref{run_time_all}.}
    \label{fig:netX}
\end{figure}

For the first set of experiments, we experiment with instances of increasing size of the Toll Collector problem (Figure~\ref{fig:tax collector}), the Frozen Islands environment (Figure \ref{fig:example}) and random partition graphs from the NetworkX library \cite{hagberg2008exploring}. The Toll Collector problem uses an MDP with three TSCCs, each of size $n$, while
the Frozen Islands environment consists of an $n \times n$ grid. 
We also experiment with random MDPs constructed from $n$-node directed Gaussian partition graphs generated using the NetworkX toolbox \cite{hagberg2008exploring}. For such graphs, the cluster sizes are drawn from a normal distribution with mean and variance $n/5$, and two nodes within the same cluster are connected with probability $p_{in}$, while two nodes in different clusters are connected with probability $p_{out}$ \cite{brandes2003experiments}. 
For these partition graphs, the state space corresponds to the vertex set, the number of actions is equal to the maximum node outdegree and the transitions are deterministic. 
The initial distribution is uniform over the set $\transientset{}$ and the rewards are selected such that only the first action from every state yields a positive reward, i.e., $R(s \in \recurrentset{}, a_1, \cdot)=1$ and 0 otherwise. 
%
An instance of an MDP constructed from a 40-node Gaussian partition graph is illustrated in Figure~\ref{fig:netX}. The specifications for the three environments are given in the caption of Table~\ref{run_time_all}. 

For each example, we generate EP, CP and CPU policies using $\mathrm{LP}_1, \mathrm{LP}_2$ and Algorithm \ref{alg:CPU}, respectively, and report on the runtime as we increase $n$. All instances were verified to meet the given specifications. The results are summarized in Table~\ref{run_time_all} demonstrating the scalability of the proposed formulations. As shown, $\mathrm{LP}_2$ incurs the largest runtime as it incorporates additional variables in the flow constraints to enforce the recurrence of the TSCCs. 

\begin{table*}[!t]
	\resizebox{\textwidth}{!}{%
		\begin{tabular}{c rrrrr rrrr rrrr}
			\toprule
			\multirow{4}{*}{\textbf{Policy}} & \multicolumn{13}{c}{\textbf{Example}} \\ 
			\cmidrule(lr){2-14} 
			& \multicolumn{5}{c}{\textbf{Toll Collector, $3n$}} & \multicolumn{4}{c}{\textbf{Frozen Lake, $n\times n$}} & \multicolumn{4}{c}{\textbf{Random Gaussian, $n$}}\\ 
			\cmidrule(lr){2-6} \cmidrule(lr){7-10} \cmidrule(lr){11-14}
			
			& 15 & 30 & 45 & 60 & 75 & $8\times 8$ & $12\times 12$ & $16\times 16$ & $20\times 20$ & 20 & 40 & 60 & 80
			\\
			\midrule
			\multicolumn{1}{c}{EP}
			& 0.5 & 1.18 & 2.4 & 4.7 & 6.99 & 1.96 & 2.87 & 5.17 & 7.97 & 0.61 & 1.58 & 2.34 & 4.16 \\
			\multicolumn{1}{c}{CP}
			& 1.16 & 3.85 & 8.37 & 15.1 & 24.48 & 5.08 & 25.81 & 108.49 & 431.38 & 4.56 & 5.84 & 19.06 & 35.74 \\
			\multicolumn{1}{c}{CPU}
			& 0.45 & 0.87 & 1.54 & 2.48 & 3.72 & 1.97 & 4.77 & 7.6 & 12.62 & 4.8 & 6.74 & 12.04 & 17.35\\
			\bottomrule
		\end{tabular}%
	}
	\caption{Average runtime results (in seconds) for 20 instances of the Toll Collector, Frozen Islands, and Gaussian partition graphs of increasing problem size $n$. The Toll Collector MDP consists of three TSCCs, each of size $n$. The detailed LMDP parameters are given in the caption of Figure~\ref{fig:tax collector} with a steady-state specification lower bound $l=0.05$. 
	The three-island problem described in 
	Figure \ref{fig:example} forms an $n \times n$ grid. In each of the smaller islands, logs are randomly distributed over $1/4$ of the states and a canoe (fishing rod) is placed in the top-left (bottom-right) tile. For these experiments, we have the constraints $(L_{\textrm{log1}} \cup L_{\textrm{log2}}, [0.3, 1]), (L_{\textrm{canoe1}} \cup L_{\textrm{canoe2}}, [0.05, 1])$ and reward function $R(\cdot, \cdot, L_{\textrm{fish1}} \cup L_{\textrm{fish2}}) = 1, R(\cdot, \cdot, S \setminus L_{\textrm{fish1}} \cup L_{\textrm{fish2}}) = 0$. For the Gaussian partition graphs, we define a steady-state specification $(L,[0.05,1])$, where $L=\{s_i\}$, for some $s_i \in \recurrentset{}$. The probability of intra-cluster connection $p_{in}=0.9$ and the probability of inter-cluster connection $p_{out}$ is $0.05, 0.01, 0.01, 0.005$ for the $20, 40, 60, 80$ nodes, respectively.   
	}
	\label{run_time_all}
\end{table*}

To further examine the scalability of the LPs underlying the different policies to much larger problem sizes, additional experiments are conducted using the CPLEX 12.8 solver. We run simulations of the LP in (\ref{eq:LP0_and_specs}), $\mathrm{LP}_1$ (with the positivity constraints) and $\mathrm{LP}_2$ (with the flow constraints) 
for random instances of the Frozen Islands problem. The runtime results are reported in Table \ref{CPLEXResults}. For a $64\times 64$ and a $128\times 128$ grid, $\mathrm{LP}_2$ (the most complex) is solved in about $20$ seconds and $15$ minutes, respectively, demonstrating the effectiveness of the developed formulations even for MDPs with over ten thousand states.

\begin{table*}[!t]
	\centering
		\begin{tabular}{c rrrrr}
			\toprule
			\multirow{3}{*}{\textbf{LP}} & \multicolumn{5}{c}{\textbf{Frozen Islands Example}} \\ 
			\cmidrule(lr){2-6} 
			& \multicolumn{5}{c}{\textbf{Size}, $n \times n$} \\ 
			\cmidrule(lr){2-6} 
			
			& $8\times 8$ & $16\times 16$ & $32\times 32$ & $64\times 64$ & $128\times 128$
			\\
			\midrule
			\multicolumn{1}{c}{$\mathrm{LP}_1$ \eqref{eqn:LP_EP}}
			& 0.0001 & 0.0017 & 0.0170 & 0.1187 & 20.306 \\ 
			\multicolumn{1}{c}{$\mathrm{LP}_2$ \eqref{eqn:LP_Flow}}
			& 0.003 & 0.038 & 0.595 & 20.251 & 933.821  \\
			\multicolumn{1}{c}{$\mathrm{LP}_3$ \eqref{eq:LP0_and_specs}}
			& 0.0001 & 0.0018 & 0.0168 & 0.1553 & 5.425 \\ 
			\bottomrule
		\end{tabular}%
	\caption{Average runtime (in seconds) of 20 instances per LP for the three-island problem described in 
	Figure \ref{fig:example}. These islands combined form an $n \times n$ grid. In each of the smaller islands, logs are randomly distributed over $1/4$ of the states and a canoe (fishing rod) is placed in the top-left (bottom-right) tile. For these experiments, we have the constraints $(L_{\textrm{log1}} \cup L_{\textrm{log2}}, [0.3, 1]), (L_{\textrm{canoe1}} \cup L_{\textrm{canoe2}}, [0.05, 1])$ and reward function $R(\cdot, \cdot, L_{\textrm{fish1}} \cup L_{\textrm{fish2}}) = 1, R(\cdot, \cdot, S \setminus L_{\textrm{fish1}} \cup L_{\textrm{fish2}}) = 0$.}
	\label{CPLEXResults}
\end{table*}

%
	


\section{Conclusion}
\label{sec:conclusions}
A framework for steady-state policy synthesis in general MDPs was developed to derive policies that satisfy constraints on the steady-state behavior of an agent. Linear programming solutions were proposed and their correctness proved for classes of edge-preserving and class-preserving policies. The framework also
enables policies that meet specified constraints on the expected number of times the agent visits transient states. Numerical simulations of the resulting policies demonstrate that our approach overcomes limitations in the literature. 

The article provides the first solution to the highly understudied problem of steady-state planning over stationary policies in constrained expected average reward multichain MDPs. The policies derived come with rigorous guarantees on the asymptotic long-term behavior of agents. The research findings have bearing on the fields of explainable, safe and trustworthy AI, where there is increased concern about explaining AI decisions, ensuring safety constraints are met, and building trust in the behavior of autonomous agents.
\acks{
This research was supported in part by the Air Force Research Laboratory through the Information Directorate’s Information Institute$^\text{®}$ contract number FA8750-20-3-1003 and FA8750-20-3-1004, the Air Force Office of Scientific Research through Award 20RICOR012, and the National Science Foundation through CAREER Award CCF-1552497 and Award CCF-2106339.} 

\begin{appendices}
{\color{black}
\setcounter{section}{0}
\renewcommand{\thesection}{Appendix \Alph{section}}
\section{Technical Lemmas}
\label{sec:appendx_proofaux}
%

\subsection*{Proof of Lemma \ref{lem:relation_bet_policies}}

Let $\pi\in\edgepreservingset$. Hence, $\recurrentset{\pi}=\recurrentset{}$ according to \eqref{eq:edge_preserving_set}. 
\textcolor{black}{Consider the set of states $\tscc{k}{}$ in a TSCC of $\cal M$, for some $k\in [m]$. We will show that this set is also a TSCC of ${\cal M}_\pi$. To this end, we first show that they form a SCC in the transition graph of ${\cal M}_\pi$. 
Since $\pi\in\edgepreservingset$, then $\pi(a|s) > 0, \forall s\in \recurrentset{}, a\in A(s)$. Hence, every action $a\in A(s)$ available in state $s\in\tscc{k}{}$ is played with non-zero probability. From \eqref{eq:M_pi}, for a pair of states $s,s'\in\tscc{k}{}$, $T_\pi(s'|s) > 0$ if $\exists a\in A(s)$ such that $T(s'|s,a) > 0$. 
Thus, for every directed path  between a pair of nodes in $\tscc{k}{}$ in the transition graph of $\cal M$, there is a similar path between the same nodes in the transition graph of ${\cal M}_\pi$. 
Therefore, the states $\tscc{k}{}$ also form a SCC in the transition graph of ${\cal M}_\pi$. 
Also, the set $\tscc{k}{}$ is reachable in ${\cal M}_\pi$ since $\tscc{k}{}\subseteq\recurrentset{\pi}$.  
Finally, there are no outgoing edges to states in $S\setminus\tscc{k}{\pi}$ since the edge set of the transition graph of ${\cal M}_\pi$ is a subset of the edge set of the transition graph of ${\cal M}$.
We conclude that $\tscc{k}{\pi} = \tscc{k}{}, \forall k\in[m]$. From the definition of the set of CP policies in \eqref{eq:class_preserving_set}, it follows that $\pi\in\classpreservingset$, proving that $\edgepreservingset\subseteq\classpreservingset$}. 
The two conditions in \eqref{eq:class_preserving_set} are special cases of the more general requirements in \eqref{eq:uptounichainset}, hence $\classpreservingset\subseteq\uptounichainset$. 
\qed

\smallbreak

The following lemma gives a characterization of the Markov chain state classification induced by a policy \eqref{eq:policy_LP1} derived from a feasible point of the constrained set $Q_0$ in \eqref{eq:Q0}.}
\begin{lemma} 
Given an MDP $\calM$, let $(x,y)\in Q_0$ defined in \eqref{eq:Q0} 
and $\pi := \pi(x,y)$ as in \eqref{eq:policy_LP1}. The following holds for the Markov chain $\calM_\pi$.
\begin{enumerate}[(a)]
\item If $s\in\transientset{}$, then $s\in\transientset{\pi}$, i.e., $\transientset{}\subseteq\transientset{\pi}$. \label{item:1}
\item If $s\in\tscc{k}{}\cap E_x$ for some $k\in [m]$, then $s\in\recurrentset{\pi}$. 
As a consequence, if $s\in\tscc{k}{}\cap\transientset{\pi}$, then $s\in \overline{E}_x$, i.e., $x_s = 0$. \label{item:2}
\end{enumerate}
\label{thm:class_under_pi}
\end{lemma}

\subsection*{Proof of Lemma \ref{thm:class_under_pi}}
\textcolor{black}{First, we show part \ref{item:1}, according to which every state in $\transientset{}$ is either transient or isolated in the Markov chain $\calM_\pi$ induced by a policy of the form \eqref{eq:policy_LP1} derived from a point in $Q_0$}. Consider $f\in\transientset{}$. From constraint \ref{LPmain_transx}, we have $x_f = 0$. Thus, from constraint \ref{LPmain_y}, \eqref{eq:policy_LP1} \textcolor{black}{and the fact that $f$ is only reachable from states in $\transientset{}$,}  
\begin{align}
y_f &= \beta_f + \sum_{f' \in \transientset{}} \sum_{a \in A(f')} y_{f'a} T(f | f', a)   \nn\\
& = \beta_f \! + \! \sum_{f'\in \transientset{}} y_{f'} \!\! \sum_{a\in A(f')} T(f | f', a) \pi(a | f') \nn\\ 
& = \beta_f \! + \! \sum_{f'\in\transientset{}} y_{f'} T_\pi (f | f')
\label{eq:y_of_node_in_F}
\end{align}
\textcolor{black}{Note that the second equality above follows from the definition of $\pi$ in  \eqref{eq:policy_LP1} for a state $f\notin E_x$}. Two cases arise. If $f\notin E_y$, then $y_f = 0$. Hence, $\beta_f = 0$ and $T_{\pi}(f|f') = 0, \forall f'\in E_y$. 
{\color{black} Thus, we have shown that every state $f$ in $\overline{E}_y\cap\transientset{}$ can only be reached from states in $\overline{E}_y\cap\transientset{}$, and that all such states have zero initial probability. Thus, every such state is either isolated or resides in an isolated component.} 
Therefore,  $f\in \transientset{\pi}$, {\color{black} where $\transientset{\pi}$ consists of transient or isolated states}. Now consider the other case where $f\in E_y$, i.e., $y_f > 0$. Assume, for the sake of contradiction, that $f\in\recurrentset{\pi}$. Hence, $f\in F$, for some TSCC $F$ (this subsumes the case where $f$ is absorbing with $|F| = 1$). Then, it must be that $F\subseteq \transientset{}$ since $f$ is not reachable from states $\Allstates\setminus \transientset{}$ even under an EP policy. Summing \eqref{eq:y_of_node_in_F} over the set $F$, we have 
\begin{align}
\sum_{f'\in F} y_{f'} &= \sum_{f'\in F} \beta_{f'} + \sum_{j \in F}\sum_{f'\in \transientset{}} y_{f'}T_\pi(j | f')\nonumber\\
& = \sum_{f'\in F} \beta_{f'}  + \sum_{f'\in \transientset{}\setminus F} y_{f'} \sum_{j \in F} T_\pi(j | f') + \sum_{f' \in F} y_{f'}\:,
\end{align}
\textcolor{black}{where the second equality is due to the closure of the set $F$, implying that $\sum_{j\in F} T_\pi(j|f') = 1$ for $f'\in F$}. It follows that $\beta_{f'} = 0, \forall f'\in F$, and $T_\pi(f|f') = 0, \forall f'\in (\transientset{}\setminus F) \cap E_y$. Therefore, $F\subseteq\transientset{\pi}$, yielding a contradiction. Hence, $f\in \transientset{\pi}$. We conclude that $\transientset{}\subseteq\transientset{\pi}$.

Next, we prove part \ref{item:2} \textcolor{black}{which states that every state $s$ in a TSCC of $\calM$ for which $x_s > 0$ is both recurrent and non-isolated in $\calM_\pi$}. Consider a state $s\in \tscc{k}{}\cap E_x$ for some $k\in [m]$, so $x_s > 0$. Assume, for the sake of contradiction, that $s\in\transientset{\pi}$, i.e., the state $s$ is either transient or isolated. If $s$ is transient, then the column of the matrix $T_\pi^\infty$ corresponding to state $s$ is zero. Therefore, from constraint \ref{LPmain_x} in \eqref{eq:Q0}, we have $x_s = 0$, i.e., $s\notin E_x$, yielding a contradiction. If $s\in F$ for some isolated component $F$, then 
\[
\sum_{s'\in F} (x_{s'} + y_{s'}) = \sum_{s'\in F} \sum_{f\in F}\sum_{a\in A(f)} y_{fa} T(s'|f,a)
\]
by summing constraint \ref{LPmain_y} over states $s'\in F$, and using the fact that $\beta_f = 0, \forall f\in F$ and that $s'\in F$ is only reachable from states in the isolated set $F$. Since $\sum_{s'\in F} T(s'|f,a) = 1, \forall f\in F, a\in A(f)$, by the closure of $F$, we get that $\sum_{s'\in F} x_{s'} = 0$ by interchanging the order of the sums, i.e., $s\in\overline{E}_x$, also yielding a contradiction.   
Hence, $s\in\recurrentset{\pi}$. 

\textcolor{black}{The second clause of Lemma \ref{thm:class_under_pi} \ref{item:2} remains to be proved, i.e., $s\in\tscc{k}{}\cap\transientset{\pi}\implies x_s = 0$.} Consider $s\in\tscc{k}{}\cap\transientset{\pi}$. Thus, $s\notin \recurrentset{\pi}$, so it follows from the result we have just shown that $s\in\transientset{}\cup\overline{E}_x$. However, since $s\in\tscc{k}{}$ for some $k$, then $s\notin\transientset{}$. Hence, $s\in\overline{E}_x$.  
\qed  

\textcolor{black}{Next, we state and prove two lemmas that will be useful in the proof of Lemma \ref{thm:mainCPU}, which establishes a sufficient condition for the existence of a one-to-one correspondence between a feasible point in $Q_0$ and the steady-state distribution of the Markov chain induced by the policy in \eqref{eq:policy_LP1} derived from this solution. }

\begin{lemma}
\label{lem:x_lp_is_statdist}
Given MDP $\calM$, if $(x,y)\in Q_0$, where $Q_0$ is the set of points in \eqref{eq:Q0}, then $x$ is a stationary distribution of the Markov chain $\calM_\pi$ induced by the policy $\pi$ in \eqref{eq:policy_LP1}. 
\end{lemma}

\subsection*{Proof of Lemma \ref{lem:x_lp_is_statdist}}
First, consider $s'\in \transientset{}$, and define $X_0:=\{x: (x,y)\in Q_0 \text{ for some } y\}$. Since $x\in X_0$, we have $x_{s'} = 0$ by constraint \ref{LPmain_transx}. Also, 
\begin{align}
    \sum_{s\in S} x_s T_\pi(s'|s) = \sum_{s\in\transientset{}} x_s T_\pi(s'|s) = 0 \:,
\end{align}
where the first equality holds since $s'\in\transientset{}$ is only reachable from states $s\in\transientset{}$ even when all edges defining possible transitions in the MDP are preserved.
Next, consider $s'\in \Allstates\setminus\transientset{}$. We have
\begin{align}
    x_{s'} :=\sum_{a\in A(s')}x_{s'a} &= \sum_{s\in\Allstates}\sum_{a\in A(s)} x_{sa} T(s'|s,a)
    = 
    \sum_{s\in E_x} \sum_{a\in A(s)} x_{s} \pi(a|s)T(s'|s,a) 
    = \sum_{s\in \Allstates} x_s T_\pi(s'|s)
\end{align}
The first equality follows from the fact that $x\in X_0$, the second from the definition of $\pi$ in \eqref{eq:policy_LP1}, and the last from the definition of $T_\pi$ in \eqref{eq:M_pi} and that $x_s = 0, \forall s\in S\setminus E_x$. Finally, $x^\top e = 1$, by summing constraints \ref{LPmain_y} over all $s'\in\Allstates$.   
\qed

\begin{lemma}
Given an MDP $\calM$, let $(x,y)\in Q_0$ and $\pi := \pi(x,y)$ as in \eqref{eq:policy_LP1}. If $\pi\in\uptounichainset$, then the subvector $x_{\tscc{k}{\pi}}$ of $x$ must satisfy the following identity for all $k\in[\NumErgSets]$
\begin{align}
x_{\tscc{k}{\pi}}^\top e = \beta_{\tscc{k}{\pi}}^\top e + \beta_{\transientset{\pi}}^\top P_{\pi,k}  \:,
    \label{eq:sum_within_Ek}
\end{align}
where, $P_{\pi,k} = [\absorbprob{f}{k}], f\in \transientset{\pi}$, is the vector of absorption probabilities from $\transientset{\pi}$ into $\tscc{k}{\pi}$ under policy $\pi$. 
\label{lem:sum_within_Ek}
\end{lemma}
%
\subsection*{Proof of Lemma \ref{lem:sum_within_Ek}}
To show \eqref{eq:sum_within_Ek}, note that, since $\pi\in\uptounichainset$, we have that $\tscc{k}{\pi}\subseteq\tscc{k}{}, k\in [m]$, where $\tscc{k}{\pi}\subset\recurrentset{\pi}$ denotes the $k$-th TSCC of $\calM_\pi$. Since $(x,y)\in Q_0$, by summing constraints \ref{LPmain_y} in \eqref{eq:Q0} over the set $\tscc{k}{\pi}$, we get
\begin{align}
\sum_{s\in\tscc{k}{\pi}} \beta_s = \sum_{s\in\tscc{k}{\pi}} x_s + \sum_{s\in \tscc{k}{\pi}} y_s - \sum_{s\in\tscc{k}{\pi}}\sum_{s'\in \tscc{k}{\pi}\cup \transientset{\pi}}\sum_{a\in A(s')} T(s|s',a) y_{s'a}
\label{eq:sum_betas_0}
\end{align}
where we used the fact that $\tscc{k}{\pi}$ is only reachable from states in $\tscc{k}{\pi}\cup \transientset{\pi}$. \textcolor{black}{By breaking the summation in the last term on the RHS of \eqref{eq:sum_betas_0} over states $s'$ in the union of the disjoint sets $\tscc{k}{\pi}$ and $\transientset{\pi}$ and interchanging the order of the summations over $s$ and $s'$}, the last term in \eqref{eq:sum_betas_0} simplifies to
\begin{align}
& \sum_{s'\in \tscc{k}{\pi}}\sum_{a\in A(s')} y_{s'a} \sum_{s\in\tscc{k}{\pi}}T(s|s',a)  + \sum_{s'\in\transientset{\pi}}\sum_{s\in\tscc{k}{\pi}}\sum_{a\in A(s')} T(s|s',a)y_{s'a}\nonumber\\
& = \sum_{s'\in\tscc{k}{\pi}} y_{s'} + \sum_{s'\in\transientset{\pi}} y_{s'}\sum_{s\in\tscc{k}{\pi}}T_{\pi}(s | s')\:,
\label{eq:simplify_ys}
\end{align}
where the first term on the RHS of the equality \eqref{eq:simplify_ys} follows from the closure of $\tscc{k}{\pi}$ (which implies that $\sum_{s\in\tscc{k}{\pi}}T(s|s',a) = 1$ for $s'\in\tscc{k}{\pi}$), and the second term from the definition of the policy in \eqref{eq:policy_LP1} for states in $\overline{E}_x$ (noting that $s'\in \transientset{\pi}$ implies $x_{s'} = 0$). 
\textcolor{black}{Replacing \eqref{eq:simplify_ys} in  \eqref{eq:sum_betas_0}, we get that}
\begin{align}
\sum_{s\in\tscc{k}{\pi}} \beta_s = \sum_{s\in \tscc{k}{\pi}} x_s - \sum_{s'\in\transientset{\pi}} y_{s'}\sum_{s\in\tscc{k}{\pi}}T_{\pi}(s | s')
\label{eq:sum_betas}
\end{align}

%
We proceed to further simplify the \textcolor{black}{second term on the RHS of \eqref{eq:sum_betas}}. Since $x_s = 0, \forall s\in\transientset{\pi}$, it follows from constraint \ref{LPmain_y} of \eqref{eq:Q0} and \eqref{eq:policy_LP1} that 
\begin{align}
   y_s = \beta_s \!+\! \sum_{s'\in\transientset{\pi}} y_{s'} \sum_{a\in A(s')}\pi(a|s') T(s|s',a)
   = \beta_s \!+\! \sum_{s'\in \transientset{\pi}} y_{s'} T_\pi(s|s'), ~\forall s\in\transientset{\pi}\:.
   \label{eq:y_f}
\end{align}
In matrix form, this can be rewritten as
\begin{align}
y_{\transientset{\pi}} = (I-Z_\pi^\top)^{-1}\beta_{\transientset{\pi}}\:,
\label{y_f}
\end{align}
where $Z_\pi = [z_{s's}] \in [0,1]^{|\transientset{\pi}|\times |\transientset{\pi}|}$, with $z_{s's} := T_\pi(s|s')$.
Hence, the second summation in \eqref{eq:sum_betas} can be written as
\begin{align}
\sum_{s'\in\transientset{\pi}} y_{s'}\sum_{s\in \tscc{k}{\pi}}T_{\pi}(s | s') & = y^\top L_{\pi,k} e 
= \beta_{\transientset{\pi}}^\top(I-Z_\pi)^{-1} L_{\pi,k} e \:,
\label{eq:F_to_Ek}
\end{align}
where $L_{\pi,k}$ is the submatrix of $T_\pi$ of transitions from $\transientset{\pi}$ to $\tscc{k}{}$ under policy $\pi$ as in \eqref{eq:T_form}. From \eqref{eq:sum_betas} and \eqref{eq:F_to_Ek}, 
\begin{align}
\sum_{s\in \tscc{k}{\pi}} x_s = \beta_{\tscc{k}{\pi}}^\top e + \beta_{\transientset{\pi}}^\top(I-Z_{\pi})^{-1} L_{\pi,k} e  \:.
\end{align}
Since the vector $P_{\pi,k}$ is the scaled (by the inverse of $\eta_s$) $s$-th column of the submatrix of the matrix $T^\infty_\pi$ defining transitions from $\transientset{\pi}$ to $\tscc{k}{\pi}$, we have \cite{feller:1968,puterman1994markov}, 
\begin{align}
 P_{\pi,k} = (I-Z_\pi)^{-1} L_{\pi,k} e \:, 
 \label{eq:abs_probs}
\end{align}
which proves the identity \eqref{eq:sum_within_Ek} of Lemma \ref{lem:sum_within_Ek}.
\qed

\textcolor{black}{We can readily state the next Lemma which establishes the aforementioned sufficiency condition.}  
\begin{lemma}
\label{thm:mainCPU}
Given an MDP $\calM$, let $(x,y)\in Q_0$ and $\pi := \pi(x,y)$ as in \eqref{eq:policy_LP1}. If $\pi\in\uptounichainset$, then $\prbetapi = x$. 
\end{lemma}
Before we prove Lemma \ref{thm:mainCPU}, we remark that this result also holds for policies in $\edgepreservingset$ and $\classpreservingset$ since these are subsets of $\uptounichainset$ per Lemma \ref{lem:relation_bet_policies}. 

\subsection*{Proof of Lemma \ref{thm:mainCPU}}
{\color{black} We seek to show that the steady-state distribution of the Markov chain $\calM_\pi$ induced by the policy $\pi$ \eqref{eq:policy_LP1} derived from a feasible point $(x,y)\in Q_0$ matches $x$, provided that $\pi$ is a CPU policy, where $Q_0$ is as defined in \eqref{eq:Q0}. 

First, we consider the states in $\transientset{\pi}$.} We have that $\text{Pr}^\infty_\pi(s) = 0, \forall s\in \transientset{\pi}$ since such states are either transient or isolated in the Markov chain ${\cal M}_\pi$ induced by policy $\pi$. Next, we argue that $x_s = 0$ for all such states. From Lemma \ref{thm:class_under_pi} \ref{item:1}, we have that $\transientset{}\subseteq \transientset{\pi}$. {\color{black} For states $s\in\transientset{}$, $x_s = 0$ by constraint \ref{LPmain_transx} in  $\eqref{eq:Q0}$. Thus, we have shown that $\text{Pr}^\infty_\pi(s) = x_s$ for every $s\in\transientset{}$}.  Now, consider a state $s\in\transientset{\pi}\setminus \transientset{}$. The state $s$ must belong to $r_k({\cal M})\cap\bar{r}({\cal M}_\pi)$ for some $k\in[\NumErgSets]$, where $m$ is the number of TSCCs in $\calM$. Hence, $x_s = 0$ by Lemma \ref{thm:class_under_pi} \ref{item:2}. Therefore, we have argued that $x_s = \prbetapi(s) = 0, \forall s\in\transientset{\pi}$.

Second, we consider states in $\recurrentset{\pi}$. 
According to Lemma \ref{lem:x_lp_is_statdist}, $x$ satisfies
\begin{align}
   x_{\tscc{k}{\pi}}^\top &= x_{\tscc{k}{\pi}}^\top T_{\pi,k}, \forall k\in[m] \:,
   \label{eq:stat_dist_Ek}
\end{align}
where $T_{\pi,k}$ is the submatrix of $T_\pi$ of transitions between states in $\tscc{k}{\pi}$. We have also shown that $x_{\tscc{k}{\pi}}$ satisfies the identity \eqref{eq:sum_within_Ek} stated in Lemma \ref{lem:sum_within_Ek}.

Given the definition of $\prbetapi(s)$ in Lemma \ref{lem:ssdist} and \eqref{eq:T_star_form}, $\prbetapi(s) = \eta_s \left( \beta_{\tscc{k}{}}^\top e + \beta_{\transientset{}}^\top P_{\pi,k} \right)$. 
Hence, 
\begin{align}
    \sum_{s\in\tscc{k}{\pi}} \prbetapi(s) = \beta_{\tscc{k}{\pi}}^\top e + \beta_{\transientset{\pi}}^\top P_{\pi,k} \:.
    \label{eq:sumXpi_within_Ek}
\end{align}
{\color{black} From \eqref{eq:sum_within_Ek} and \eqref{eq:sumXpi_within_Ek}, 
we conclude that 
\begin{align}
\sum_{s\in \tscc{k}{\pi}} \prbetapi(s) = \sum_{s\in \tscc{k}{\pi}} x_s.
\label{eq:sumx_equal_sump}
\end{align}
\textcolor{black}{The ergodic theorem of Markov chains asserts that the solution to $x^\top T = x^\top, \text {where } x^\top e = 1, x\geq 0$, is unique iff $T$ is the transition matrix of a unichain \cite{Gallager_book,altman1999constrained}. From \eqref{eq:stat_dist_Ek}, \eqref{eq:sumXpi_within_Ek} 
and \eqref{eq:sumx_equal_sump}, we have shown that \[
x_k^\top T_{\pi,k} = x_k^\top, \text{ where } x_k^\top e = c_k, ~x_k\geq 0 \:
\] 
for TSCCs $k\in[m]$, where $x_k:= x_{\tscc{k}{\pi}}$, $c_k$ is the RHS of \eqref{eq:sumXpi_within_Ek}, and $~\sum_{k=1}^m c_k = 1$. Further, since $\pi\in\uptounichainset$, every TSCC is a unichain. Hence,} by the ergodic theorem, 
the solution $x_{\tscc{k}{\pi}}$ to \eqref{eq:stat_dist_Ek} and \eqref{eq:sum_within_Ek} is unique for each component $\tscc{k}{\pi}, k\in [m]$, thus $x$ is equal to the unique steady-state distribution, i.e., $x = \prbetapi$.}
\qed

\smallbreak
We also make use of the following lemma in the proof of the converse part of 
Theorem \ref{thm:mainEP}. 
The lemma 
establishes that all occupation measures induced by the policies of interest are $Q_0$-feasible.
\begin{lemma}
\label{thm:XPi_in_LP}
Given MDP $\calM$, let $X_0:=\{x: (x,y)\in Q_0 \text{ for some } y\}$, where $Q_0$ is as defined in \eqref{eq:Q0}.  
Then, $\setssdistbeta(\uptounichainset)\subseteq X_0$.  
\end{lemma}

\subsection*{Proof of Lemma \ref{thm:XPi_in_LP}}
We show that the steady-state distribution induced by every CPU policy is in $X_0$. To this end, let $x\in \setssdistbeta(\uptounichainset)$, i.e., $\exists\pi\in\uptounichainset: \prbetapi = x$, where $\prbetapi$ is as defined in Lemma \ref{lem:ssdist}. Therefore, $x$ is a stationary distribution of the Markov chain $\calM_\pi$, in which $\tscc{k}{\pi},k\in[\NumErgSets]$ are TSCCs and states $\transientset{\pi}$
are either transient or isolated. 
Hence, $ x^\top = x^\top T_\pi$. Therefore,
\begin{align}
    x_{s'} &:= \sum_{a\in A(s')} x_{s'a}= \sum_{s\in \Allstates} x_s T_\pi(s'|s) = \sum_{s\in \Allstates} \sum_{a\in A(s)} x_s \pi(a|s) T(s'|s,a) \nonumber\\
    &=  \sum_{s\in \Allstates} \sum_{a\in A(s)} x_{sa} T(s'|s,a)\:,
\end{align}
{\color{black} where the last equality follows since $\prbetapi(s,a) = \prbetapi(s)\pi(a|s)$.} Thus, the steady-state distribution $x$ satisfies constraint \ref{LPmain_x} in \eqref{eq:Q0}. From the definition of $\uptounichainset$ in \eqref{eq:uptounichainset}, every $f\in\transientset{}$ is either transient or isolated under $\pi$. Thus, $x_{\transientset{}} := \{\prbetapi(f,a)\}_{f\in\transientset{}, a} = 0$, 
satisfying constraint \ref{LPmain_transx}. 

{\color{black} The variables $y_{fa}, f\in \transientset{\pi}, a\in A(f)$, can be set as in \eqref{y_f}, i.e., choose $y_{fa} = \beta_{\transientset{\pi}}^\top (I-Z_\pi)^{-1} e_f \pi(a|f), f\in \transientset{\pi}, a\in A(f)$, where $Z_\pi$ is the submatrix of $T_\pi$ defined in \eqref{eq:T_form}, which satisfies the constraints \ref{LPmain_y} as we have already shown in \eqref{eq:y_f}. The remaining variables $y_{sa}, s\in\tscc{k}{\pi}, a\in A(s)$, can now be chosen in terms of $x_{sa}, y_{fa}, T(s'|s,a)$ and $\beta$ such that the corresponding constraints \ref{LPmain_y} are satisfied. 
Thus, for the given $x$, we have shown the existence of a feasible $y$ such that $(x,y)\in Q_0$. Therefore, $x\in X_0$.}
\qed

\renewcommand{\thesection}{Appendix \Alph{section}}
\section{Proof of Main Theorems}
\label{sec:appendx_proofmain}

\subsection*{Proof of Theorem \ref{thm:pi_in_ext_edge_set}}
Since $(x,y)$ is a feasible point of $\mathrm{LP}_1$, we have that $(x,y)\in Q_0$ per \eqref{eqn:LP_EP}. From Lemma \ref{thm:class_under_pi} \ref{item:1}, $\transientset{}\subseteq\transientset{\pi}$, \textcolor{black}{thus $\recurrentset{}\supseteq\recurrentset{\pi}$.  
Consider a state $s\in\recurrentset{}$.} Then, $s\in\tscc{k}{}$ for some $k\in[m]$. From the positivity constraint \ref{LPmain_strictposx} of $\mathrm{LP}_1$, \textcolor{black}{we also have that $x_s > 0$, i.e., $s\in E_x$}. \textcolor{black}{Since $s\in\tscc{k}{}\cap E_x$, it follows that $s\in\recurrentset{\pi}$ by Lemma \ref{thm:class_under_pi} \ref{item:2}. Therefore, $\recurrentset{}\subseteq\recurrentset{\pi}$.} 
We conclude that $\recurrentset{\pi} = \recurrentset{}$. 
\textcolor{black}{From constraint \ref{LPmain_strictposx}, $x_{sa}>0, \forall s\in \recurrentset{}, a\in A(s)$. It follows from the definition of $\pi$ in \eqref{eq:policy_LP1} for states $s\in E_x$ that $\pi(a|s) > 0, \forall s\in \recurrentset{}, a\in A(s)$. We have shown that $\pi$ satisfies both requirements in \eqref{eq:edge_preserving_set}, hence $\pi\in\edgepreservingset$.}    
\qed

\subsection*{Proof of Theorem \ref{thm:mainEP}}
$(\implies)$ \textcolor{black}{First, we show that if \eqref{eqn:LP_EP} is feasible, then there exist an EP policy that meets the specifications $\SSSpec$}. Let $(x,y)\in Q_1$ denote a feasible solution to \eqref{eqn:LP_EP} and let $\pi$ be defined as in \eqref{eq:policy_LP1}. By Theorem \ref{thm:pi_in_ext_edge_set}, $\pi\in\edgepreservingset$. \textcolor{black}{By Lemma \ref{lem:relation_bet_policies}, we also have that $\pi\in\uptounichainset$}. Invoking Lemma \ref{thm:mainCPU}, we conclude that $\prbetapi(s,a) = x_{sa}, s\in S, a\in A(s)$, \textcolor{black}{i.e., $x$ is equal to the steady-state distribution of the Markov chain $\calM_\pi$ induced by policy $\pi$. Since $x$ satisfies constraint \ref{LPmain_SSspecs}, this implies that $\calM_\pi$ meets the specifications $\SSSpec$}.
\smallbreak 

\noindent$(\impliedby)$ \textcolor{black}{Now, we show the converse, that is, the existence of an EP policy that meets the specifications implies that $\mathrm{LP}_1$ in \eqref{eqn:LP_EP} is feasible}. Define $V:=\{x: \ref{LPmain_SSspecs} \text{ and } \ref{LPmain_strictposx} \text{ satisfied}\}$. Thus, we have that $X_{\textit{LP}_1} = X_0 \cap V$, where $X_{\textit{LP}_1} = \{x: (x,y)\in Q_1 \text{ for some } y\}$. Suppose $\exists \pi\in\edgepreservingset$ that satisfies the specifications $\SSSpec$ as in the statement of Theorem \ref{thm:mainEP}. Then $\prbetapi\in\setssdistbeta(\edgepreservingset)$ is well-defined as in Lemma \ref{lem:ssdist}. 
\textcolor{black}{We have $\prbetapi(s) := (\beta^\top \cesarolimit{T_\pi})_s > 0, \forall s\in\recurrentset{\pi}$, since all such states are recurrent in the Markov chain ${\cal M}_\pi$.  
Since $\pi\in\edgepreservingset$, $\pi(a|s) > 0, \forall s\in \recurrentset{}, a\in A(s)$, from \eqref{eq:edge_preserving_set}. Hence, by Lemma \ref{lem:ssdist}, $\prbetapi(s,a)>0, \forall s\in \recurrentset{}, a\in A(s)$. 
Therefore, $\prbetapi\in V$.} 
Hence, $\setssdistbeta(\edgepreservingset)\cap V$ is non-empty. Set $x_{sa} = \prbetapi(s,a), s\in S, a\in A(s)$. Recall that $\transientset{\pi} = \transientset{}$ since $\pi\in\edgepreservingset$, so we have $x_{sa}=\prbetapi(s,a) = 0, \forall s \in \transientset{}$. From Lemma \ref{thm:XPi_in_LP}, $\setssdistbeta(\edgepreservingset)\subseteq X_0$, \textcolor{black}{where we also use the fact that $\setssdistbeta(\edgepreservingset)\subseteq\setssdistbeta(\uptounichainset)$ as a consequence of Lemma \ref{lem:relation_bet_policies}}. The variables $y_{sa}$ can be defined in terms of $x_{sa}, T(s'|s,a)$ and $\beta$ such that the constraints \ref{LPmain_y} are satisfied. Hence, $X_{\textit{LP}_1}$, and in turn $Q_1$, is non-empty. 
The optimality of $\pi^*$ follows from the optimality of $(x^*, y^*)$, Theorem \ref{thm:pi_in_ext_edge_set} and the established equality $\mathrm{Pr}^\infty_{\pi^*} = x^*$. 
\qed

\subsection*{Proof of Theorem \ref{thm:pol_in_CP}}
Let $f\in\transientset{}$. From Lemma \ref{thm:class_under_pi} \ref{item:1}, $f\in\transientset{\pi}$. Now consider $s\in\tscc{k}{}$ for some $k\in[m]$. As argued earlier, every state in $\tscc{k}{}$ is reachable from $s$ given constraints \ref{LP_flowCap}, \ref{LP_flowTransfer}, \ref{LP_flowIn} of \eqref{eqn:LP_Flow}. In addition, $s$ is reachable from all states in $\tscc{k}{}$, which follows from constraints \ref{LP_flowRevInit}, \ref{LP_flowRevCap}, \ref{LP_flowRevTransfer}, \ref{LP_flowRevIn}.  
Hence, $s\in\recurrentset{\pi}$. Therefore, $\recurrentset{}\subseteq\recurrentset{\pi}$. Since we have already shown that $\transientset{}\subseteq\transientset{\pi}$, we conclude that $\recurrentset{\pi} = \recurrentset{}$. Therefore, $\pi\in\classpreservingset$ defined in \eqref{eq:class_preserving_set}.  
\qed

\subsection*{Proof of Theorem \ref{thm:mainCP}}
The proof follows the same reasoning as that of Theorem \ref{thm:mainEP}.

\noindent $(\implies)$ Let $(x,y,f,f^\text{rev})\in Q_2$ denote a feasible solution to \eqref{eqn:LP_Flow} and let $\pi$ be defined as in \eqref{eq:policy_LP1}. By Theorem \ref{thm:pol_in_CP}, $\pi\in\classpreservingset$. Invoking Lemma \ref{thm:mainCPU} and Lemma \ref{lem:relation_bet_policies}, we have that $\prbetapi(s,a) = x_{sa}, s\in S, a\in A(s)$, which implies that $\calM_\pi$ meets the specifications $\SSSpec$ per constraint \ref{LPmain_SSspecs}. 

\noindent$(\impliedby)$ Define $V:=\{x: \ref{LPmain_SSspecs} \text{ and } \ref{LP_flowInit}-\ref{LP_flowVars} \text{ satisfied}\}$. Thus, we have that $X_{\textit{LP}_2} = X_0 \cap V$, where $X_{\textit{LP}_2} = \{x: (x,y,f,f^\text{rev})\in Q_2\}$. Suppose $\exists \pi\in\classpreservingset$ that satisfies the specifications $\SSSpec$ as in the statement of Theorem \ref{thm:mainCP}. Hence, $\tscc{k}{\pi}, k\in[m]$ are the recurrent components of $\calM_\pi$. Then, $\prbetapi\in\setssdistbeta(\classpreservingset)$ is well-defined as in Lemma \ref{lem:ssdist}. 
We can set $x_{sa} = \prbetapi(s,a) = \pi(a|s)\prbetapi(s)$ for every $s\in S, a\in A(s)$. The flow variables in \ref{LP_flowInit} - \ref{LP_flowRevInit} can be defined in terms of $x_{sa}$ and $T(s'|s,a)$ such that the constraints \ref{LP_flowTransfer} - \ref{LP_flowRevIn} are satisfied. Hence, $x\in V$, i.e., $\setssdistbeta(\classpreservingset)\cap V$ is non-empty. 
By Lemma \ref{thm:XPi_in_LP}, $X_{\textit{LP}_2}$ and $Q_2$ are non-empty. 
The optimality of $\pi^*$ follows from the optimality of $(x^*, y^*)$, Theorem \ref{thm:pol_in_CP} and the established equality $\mathrm{Pr}^\infty_{\pi^*} = x^*$. 
\qed

\subsection*{Proof of Theorem \ref{thm:CPU}}
Assume $(x,y)\in Q^*$. We have that  $V_k^+(x)\subseteq\tscc{k}{\pi}$ for $\pi$ in \eqref{eq:policy_LP1}
by Lemma \ref{thm:class_under_pi} \ref{item:2}. Further, consider $s\in\tscc{k}{}\cap\overline{E}_x$. By constraint \ref{LPmain_x} and the definition of $\pi$ in \eqref{eq:policy_LP1}, $T_\pi(s|s') = 0, \forall s'\in V_k^+(x)$. For the sake of contradiction, assume $s\in\recurrentset{\pi}$. Hence, $s\in F\subseteq\tscc{k}{}$ for some TSCC $F$ of $\calM_\pi$. Summing constraints \ref{LPmain_y} over the set $F$, we get that $\beta_s = 0, \forall s\in F$ and $T_\pi(s|s') = 0, s'\in\transientset{}, s\in F$. Hence, $s\in\transientset{\pi}$, yielding a contradiction. We conclude that $V_k^+(x) = \recurrentset{\pi}\cap\tscc{k}{}$. 
Therefore, if for every $k\in[m]$ we have that the subgraph $(V_k^+(x),E_k^+(x))$ is strongly connected, then $V_k^+(x)$ is a SCC in $\calM_\pi$, for $\pi$ in \eqref{eq:policy_LP1}. Hence, $V_k^+(x)$ 
is the unique TSCC  $\tscc{k}{\pi}$ in the set $\tscc{k}{}$,  
i.e., $\pi\in\uptounichainset$. The result now follows from Lemma \ref{thm:mainCPU}.  
\qed

{\color{black}
\subsection*{Proof of Theorem \ref{thm:EP_with_eps}}
(1) Every feasible solution of $\mathrm{LP}_1(\epsilon)$ is also $\mathrm{LP}_1$-feasible. Hence, the result follows as an immediate consequence of Theorem \ref{thm:pi_in_ext_edge_set}. \\
(2) The proof of part (2) follows the same reasoning as in the proof of the converse of Theorem \ref{thm:mainEP}. Specifically, we have shown that, if $\pi\in\edgepreservingset$, then $\prbetapi(s,a)>0, \forall s\in\recurrentset{\pi},a\in A(s)$. Hence, $\exists\epsilon > 0$ such that $\prbetapi\in V'$, where $V':=\{x: \ref{LPmain_SSspecs}  \text{ and } (v)' \text{ satisfied)}\}$. Therefore, $X_{\textit{LP}_1(\epsilon)} := X_0\cap V'$ is non-empty, and in turn $\mathrm{LP}_1(\epsilon)$ is feasible. 
\\
(3) 
Let $\epsilon_n\rightarrow 0, n\in\mathbb{N}$, be a monotonically decreasing sequence, $\pi_n^*$ the EP policy in \eqref{eq:policy_LP1} corresponding to an optimal solution to $\mathrm{LP}_1(\epsilon_n)$, and $R_n:= R^\infty_{\pi_n^*}(\beta)$. The sequence $(R_n)_{n\in\mathbb{N}}$ is monotonically non-decreasing since $R_n\geq R_m$ whenever $\epsilon_n < \epsilon_m$. Further, from \eqref{eq:avgreward_equiv}, we have that the sequence is bounded above since $\sup_{\pi\in\edgepreservingset} R^\infty_\pi(\beta) \leq r_\textrm{max}$, where $r_\textrm{max}:=\max_{s\in S, a\in A(s)} R(s,a)$. 
Since the sequence $(R_n)_{n\in\mathbb{N}}$ is both increasing and bounded, it converges to the limit $\sup_n R_n$ by the monotone convergence theorem \cite{royden}. We are only left to show that $\sup_n R_{n} = \sup_{\pi\in\edgepreservingset} R_\pi^\infty$. To this end, assume for the sake of contradiction that $\sup_n R_n < \sup_{\pi\in\edgepreservingset} R_\pi^\infty$. 
Since the RHS of the inequality is the least upper bound on the average reward of EP policies, then for any $\delta > 0, \exists \pi'\in\edgepreservingset: R_{\pi'}^\infty > \sup_{\pi\in\edgepreservingset} R_\pi^\infty - \delta$. We can choose $\delta$ small enough such that $R_{\pi'}^\infty > \sup_n R_n$. From part (2) above,  $\exists\epsilon > 0$, such that $\mathrm{Pr}^\infty_{\pi'}$ is $\mathrm{LP}_1(\epsilon')$-feasible for all $\epsilon'\leq\epsilon$. Hence, from the definition of $\pi_n^*$, we get that $\sup_n R_n \geq R_{\pi'}^\infty$, yielding a contradiction. 
\qed
}

{\color{black}
\subsection*{Proof of Theorem \ref{thm:modifiedEP}}
\noindent (1) Let $x$ be $\mathrm{LP}_1(\delta)$-feasible. Since the feasible set for $\mathrm{LP}_1(\delta)$ is a subset of the feasible set of $\mathrm{LP}_1$, then $\pi\in\Pi_{EP}$ by Theorem \ref{thm:pi_in_ext_edge_set}. Therefore, we only need to verify the bounded support requirement in \eqref{eq:modifiedEP}. For $s\in\recurrentset{}$, we have that $x_{sa}\geq\delta > 0, a\in A(s)$, from constraint $(v)'$ in $\mathrm{LP}_1(\delta)$. Hence, $\pi(a|s) = x_{sa}/x_s \geq\delta$. Therefore, $\pi\in\edgepreservingset(\delta)$.\\

\noindent (2) Assume $\pi\in\edgepreservingset(\delta)$ and meets the specifications $\SSSpec$. Noting that $\edgepreservingset(\delta)\subset\edgepreservingset$, then there exists an $0< \epsilon\leq \delta$ such that $\prbetapi$ is a feasible solution of $\mathrm{LP}_1(\epsilon)$, which follows from part (2) of Theorem \ref{thm:EP_with_eps}. Hence, $\max_{\pi\in\edgepreservingset(\delta)}R_\pi^\infty(\beta)\leq R^*(\epsilon)$ since $R^*(\epsilon)$ is the optimal value of $\mathrm{LP}_1(\epsilon)$, where $\epsilon\leq\delta$ is a function of $\delta$. As $\delta\rightarrow 0$, the sequence of rewards $R^*(\delta)$ is monotonically non-decreasing and bounded above. Hence, as $\delta\rightarrow 0$, the sequence $R^*(\delta)$ converges to a limit. Every convergent sequence is a Cauchy sequence \cite{royden}, i.e., the elements of the sequence become arbitrarily close to each other as $\delta\rightarrow 0$. Hence, $R^*(\epsilon) - R^*(\delta) \rightarrow 0$, as $\delta\rightarrow 0$. 
\qed
}

\subsection*{Proof of Theorem \ref{thm:dual_cone}}
The cone $V(x,y)$ in \eqref{eq:cone} is the cone of feasible directions from a feasible point $(x,y)$, i.e., directions $v = (h,z)$ along which $\exists\lambda > 0$ such that $(x,y) + \lambda(h,z)$ is feasible. The sets $u(x)$, $l(x)$, $n(x)$ and $m(y)$ denote the sets of active (upper and lower) specification and non-negativity (of state-action variables $x$ and $y$) constraints, respectively. Since the rewards vector $R$ is an interior point of the dual cone $V^*(x,y)$ designated in the statement of Theorem \ref{thm:dual_cone}, moving away from $(x,y)$ along any feasible direction can only reduce the value of the objective, i.e., $\sum_{s\in S}\sum_{a\in A(s)} R(s,a) h_{sa} < 0$. Hence, $(x,y)$ is the unique optimal solution to \eqref{eq:LP0_and_specs}. We have already shown that the set of occupation measures induced by policies for which   $\transientset{}\subseteq\transientset{\pi}$ is contained in the feasible set of $\eqref{eq:LP0_and_specs}$. Since $\pi$ in \eqref{eq:policy_LP1} is one such policy by Lemma \ref{thm:class_under_pi} \ref{item:1}, we have $\prbetapi = x$ and $\pi$ meets the specifications $\SSSpec$. The uniqueness of $\pi$ in this class of policies follows from the established uniqueness of the optimal solution $x$.         
\qed


\subsection*{Proof of Proposition \ref{thm:unichainpol}}
By Lemma \ref{thm:class_under_pi}, we have that $f\in\transientset{\pi}$. If $\pi\in\uptounichainset$, then the condition of Lemma~\ref{thm:mainCPU} is met, and it follows from \eqref{y_f} that $y_f = \beta_{\transientset{\pi}}^\top (I-Z_\pi)^{-1} e_f = \TransExpect_{\pi}(f)$.
\qed

\end{appendices}

%% file: main.bbl
\begin{thebibliography}{}

\bibitem[\protect\BCAY{Akshay, Bertrand, Haddad,\ \BBA\ H{\'e}lou{\"e}t}{Akshay
  et~al.}{2013}]{SSC}
Akshay, S., Bertrand, N., Haddad, S., \BBA\ H{\'e}lou{\"e}t, L. \BBOP2013\BBCP.
\newblock \BBOQ The steady-state control problem for {Markov} decision
  processes\BBCQ\
\newblock In {\Bem International Conference on Quantitative Evaluation of
  Systems}, \BPGS\ 290--304, Berlin Heidelberg. Springer.

\bibitem[\protect\BCAY{Altman}{Altman}{1998}]{altman_total_cost_98}
Altman, E. \BBOP1998\BBCP.
\newblock \BBOQ Constrained {Markov} decision processes with total cost
  criteria: {Lagrangian} approach and dual linear program\BBCQ\
\newblock {\Bem Mathematical Methods of Operations Research}, {\Bem 48\/}(3),
  387--417.

\bibitem[\protect\BCAY{Altman}{Altman}{1999}]{altman1999constrained}
Altman, E. \BBOP1999\BBCP.
\newblock {\Bem Constrained Markov decision processes}.
\newblock CRC Press, Boca Raton.

\bibitem[\protect\BCAY{Altman, Boularouk,\ \BBA\ Josselin}{Altman
  et~al.}{2019}]{Altman2019_11}
Altman, E., Boularouk, S., \BBA\ Josselin, D. \BBOP2019\BBCP.
\newblock \BBOQ Constrained {Markov} decision processes with total expected
  cost criteria\BBCQ\
\newblock In {\Bem Proceedings of the 12th EAI International Conference on
  Performance Evaluation Methodologies and Tools}, \BPGS\ 191--192. ACM.

\bibitem[\protect\BCAY{Atia, Beckus, Alkhouri,\ \BBA\ Velasquez}{Atia
  et~al.}{2020}]{ijcai2020}
Atia, G., Beckus, A., Alkhouri, I., \BBA\ Velasquez, A. \BBOP2020\BBCP.
\newblock \BBOQ Steady-state policy synthesis in multichain {Markov} decision
  processes\BBCQ\
\newblock In {\Bem Proceedings of the 29th International Joint Conference on
  Artificial Intelligence (IJCAI)}, \BPGS\ 4069--4075. International Joint
  Conferences on Artificial Intelligence Organization.

\bibitem[\protect\BCAY{Ayala, Andersson,\ \BBA\ Belta}{Ayala
  et~al.}{2014}]{CSLControllerSynthesis}
Ayala, A.~M., Andersson, S.~B., \BBA\ Belta, C. \BBOP2014\BBCP.
\newblock \BBOQ Formal synthesis of control policies for continuous time markov
  processes from time-bounded temporal logic specifications\BBCQ\
\newblock {\Bem IEEE Transactions on Automatic Control}, {\Bem 59\/}(9),
  2568--2573.

\bibitem[\protect\BCAY{Baiocchi}{Baiocchi}{2010}]{remoteControl}
Baiocchi, D. \BBOP2010\BBCP.
\newblock {\Bem Confronting Space Debris: Strategies and Warnings from
  Comparable Examples Including Deepwater Horizon}.
\newblock Rand Corporation.

\bibitem[\protect\BCAY{Baumgartner, Thi{\'e}baux,\ \BBA\ Trevizan}{Baumgartner
  et~al.}{2018}]{occupationPLTL}
Baumgartner, P., Thi{\'e}baux, S., \BBA\ Trevizan, F. \BBOP2018\BBCP.
\newblock \BBOQ Heuristic search planning with multi-objective probabilistic
  {LTL} constraints\BBCQ\
\newblock In {\Bem Sixteenth International Conference on Principles of
  Knowledge Representation and Reasoning}, \BPGS\ 415--424.

\bibitem[\protect\BCAY{Bertsekas}{Bertsekas}{2005}]{bertsekas2005dynamic}
Bertsekas, D. \BBOP2005\BBCP.
\newblock {\Bem Dynamic programming and optimal control}, \lowercase{\BVOL}~2.
\newblock Athena Scientific, Belmont, Mass.

\bibitem[\protect\BCAY{Bertsimas\ \BBA\ Tsitsiklis}{Bertsimas\ \BBA\
  Tsitsiklis}{1997}]{linear_opt_book}
Bertsimas, D.\BBACOMMA\  \BBA\ Tsitsiklis, J. \BBOP1997\BBCP.
\newblock {\Bem Introduction to Linear Optimization\/} (1st \BEd).
\newblock Athena Scientific.

\bibitem[\protect\BCAY{Bhatnagar\ \BBA\ Lakshmanan}{Bhatnagar\ \BBA\
  Lakshmanan}{2012}]{ConstrainedActorCritic_12}
Bhatnagar, S.\BBACOMMA\  \BBA\ Lakshmanan, K. \BBOP2012\BBCP.
\newblock \BBOQ An online actor--critic algorithm with function approximation
  for constrained {Markov} decision processes\BBCQ\
\newblock {\Bem Journal of Optimization Theory and Applications}, {\Bem
  153\/}(3), 688--708.

\bibitem[\protect\BCAY{Boussemart\ \BBA\ Limnios}{Boussemart\ \BBA\
  Limnios}{2004}]{minFailureRate_7}
Boussemart, M.\BBACOMMA\  \BBA\ Limnios, N. \BBOP2004\BBCP.
\newblock \BBOQ Markov decision processes with asymptotic average failure rate
  constraint\BBCQ\
\newblock {\Bem Communications in Statistics-Theory and Methods}, {\Bem
  33\/}(7), 1689--1714.

\bibitem[\protect\BCAY{Boussemart, Limnios,\ \BBA\ Fillion}{Boussemart
  et~al.}{2002}]{minFailureRate_8}
Boussemart, M., Limnios, N., \BBA\ Fillion, J. \BBOP2002\BBCP.
\newblock \BBOQ Non-ergodic {Markov} decision processes with a constraint on
  the asymptotic failure rate: general class of policies\BBCQ\
\newblock {\Bem Stochastic models}, {\Bem 18\/}(1), 173--191.

\bibitem[\protect\BCAY{Brafman\ \BBA\ De~Giacomo}{Brafman\ \BBA\
  De~Giacomo}{2019}]{LDLfPlanning}
Brafman, R.~I.\BBACOMMA\  \BBA\ De~Giacomo, G. \BBOP2019\BBCP.
\newblock \BBOQ Planning for {LTLf /LDLf} goals in non-{Markovian} fully
  observable nondeterministic domains\BBCQ\
\newblock In {\Bem Proceedings of the Twenty-Eighth International Joint
  Conference on Artificial Intelligence (IJCAI)}, \BPGS\ 1602--1608.
  International Joint Conferences on Artificial Intelligence Organization.

\bibitem[\protect\BCAY{Brandes, Gaertler,\ \BBA\ Wagner}{Brandes
  et~al.}{2003}]{brandes2003experiments}
Brandes, U., Gaertler, M., \BBA\ Wagner, D. \BBOP2003\BBCP.
\newblock \BBOQ Experiments on graph clustering algorithms\BBCQ\
\newblock In {\Bem European Symposium on Algorithms}, \BPGS\ 568--579.
  Springer.

\bibitem[\protect\BCAY{Brockman, Cheung, Pettersson, Schneider, Schulman,
  Tang,\ \BBA\ Zaremba}{Brockman et~al.}{2016}]{OpenAIGym}
Brockman, G., Cheung, V., Pettersson, L., Schneider, J., Schulman, J., Tang,
  J., \BBA\ Zaremba, W. \BBOP2016\BBCP.
\newblock \BBOQ {OpenAI Gym}\BBCQ\
\newblock {\Bem arXiv preprint arXiv:1606.01540}.

\bibitem[\protect\BCAY{Camacho\ \BBA\ McIlraith}{Camacho\ \BBA\
  McIlraith}{2019}]{LTLfPlanning}
Camacho, A.\BBACOMMA\  \BBA\ McIlraith, S.~A. \BBOP2019\BBCP.
\newblock \BBOQ Strong fully observable non-deterministic planning with {LTL}
  and {LTLf} goals\BBCQ\
\newblock In {\Bem Proceedings of the Twenty-Eighth International Joint
  Conference on Artificial Intelligence (IJCAI)}, \BPGS\ 5523--5531.
  International Joint Conferences on Artificial Intelligence Organization.

\bibitem[\protect\BCAY{Courcoubetis\ \BBA\ Yannakakis}{Courcoubetis\ \BBA\
  Yannakakis}{1995}]{cour_yann_jacm_1995}
Courcoubetis, C.\BBACOMMA\  \BBA\ Yannakakis, M. \BBOP1995\BBCP.
\newblock \BBOQ The complexity of probabilistic verification\BBCQ\
\newblock {\Bem Journal of the ACM}, {\Bem 42\/}(4), 857–907.

\bibitem[\protect\BCAY{De~Ghellinck}{De~Ghellinck}{1960}]{DeGhellinck1960}
De~Ghellinck, G. \BBOP1960\BBCP.
\newblock \BBOQ Les probl\`{e}mes de d\'{e}cisions s\'{e}quentielles\BBCQ\
\newblock {\Bem Cahiers du Centre d’Etudes de Recherche Op{\'e}rationnelle},
  {\Bem 2\/}(2), 161--179.

\bibitem[\protect\BCAY{De~Giacomo, Felli, Patrizi,\ \BBA\ Sardina}{De~Giacomo
  et~al.}{2010}]{MuCalculusPlanning}
De~Giacomo, G., Felli, P., Patrizi, F., \BBA\ Sardina, S. \BBOP2010\BBCP.
\newblock \BBOQ Two-player game structures for generalized planning and agent
  composition\BBCQ\
\newblock In {\Bem Twenty-Fourth AAAI Conference on Artificial Intelligence}.

\bibitem[\protect\BCAY{Denardo\ \BBA\ Fox}{Denardo\ \BBA\
  Fox}{1968}]{10.2307/2099444}
Denardo, E.~V.\BBACOMMA\  \BBA\ Fox, B.~L. \BBOP1968\BBCP.
\newblock \BBOQ Multichain {Markov} renewal programs\BBCQ\
\newblock {\Bem SIAM Journal on Applied Mathematics}, {\Bem 16\/}(3), 468--487.

\bibitem[\protect\BCAY{Derman}{Derman}{1970}]{Derman:1970}
Derman, C. \BBOP1970\BBCP.
\newblock {\Bem Finite State Markovian Decision Processes}.
\newblock Academic Press, Inc., Orlando, FL, USA.

\bibitem[\protect\BCAY{Engesser, Bolander,\ \BBA\ Nebel}{Engesser
  et~al.}{2017}]{DELPlanning}
Engesser, T., Bolander, T., \BBA\ Nebel, B. \BBOP2017\BBCP.
\newblock \BBOQ Cooperative epistemic multi-agent planning with implicit
  coordination\BBCQ\
\newblock In {\Bem Proceedings of the 3rd {Workshop} on {Distributed} and
  {Multi}-{Agent} {Planning} ({DMAP})}, \BPG~68.

\bibitem[\protect\BCAY{{Feinberg}}{{Feinberg}}{2009}]{4927531}
{Feinberg}, E.~A. \BBOP2009\BBCP.
\newblock \BBOQ Adaptive computation of optimal nonrandomized policies in
  constrained average-reward {MDP}s\BBCQ\
\newblock In {\Bem IEEE Symposium on Adaptive Dynamic Programming and
  Reinforcement Learning}, \BPGS\ 96--100.

\bibitem[\protect\BCAY{Feinberg}{Feinberg}{2000}]{fein_or_2000}
Feinberg, E.~A. \BBOP2000\BBCP.
\newblock \BBOQ Constrained discounted {Markov} decision processes and
  {Hamiltonian} cycles\BBCQ\
\newblock {\Bem Mathematics of Operations Research}, {\Bem 25\/}(1), 130--140.

\bibitem[\protect\BCAY{Feller}{Feller}{1968}]{feller:1968}
Feller, W. \BBOP1968\BBCP.
\newblock {\Bem An Introduction to Probability Theory and its Applications\/}
  (3rd \BEd)., \lowercase{\BVOL}~1.
\newblock Wiley.

\bibitem[\protect\BCAY{Gallager}{Gallager}{2013}]{Gallager_book}
Gallager, R.~G. \BBOP2013\BBCP.
\newblock {\Bem Stochastic Processes: Theory for Applications}.
\newblock Cambridge University Press, New York.

\bibitem[\protect\BCAY{Grant\ \BBA\ Boyd}{Grant\ \BBA\ Boyd}{2008}]{gb08}
Grant, M.\BBACOMMA\  \BBA\ Boyd, S. \BBOP2008\BBCP.
\newblock \BBOQ Graph implementations for nonsmooth convex programs\BBCQ\
\newblock In Blondel, V., Boyd, S., \BBA\ Kimura, H.\BEDS, {\Bem Recent
  Advances in Learning and Control}, Lecture Notes in Control and Information
  Sciences, \BPGS\ 95--110. Springer-Verlag Limited.

\bibitem[\protect\BCAY{Grant\ \BBA\ Boyd}{Grant\ \BBA\ Boyd}{2014}]{cvx}
Grant, M.\BBACOMMA\  \BBA\ Boyd, S. \BBOP2014\BBCP.
\newblock \BBOQ {CVX}: Matlab software for disciplined convex programming,
  version 2.1\BBCQ.

\bibitem[\protect\BCAY{Guo\ \BBA\ Zavlanos}{Guo\ \BBA\
  Zavlanos}{2018}]{LTLPlanning}
Guo, M.\BBACOMMA\  \BBA\ Zavlanos, M.~M. \BBOP2018\BBCP.
\newblock \BBOQ Probabilistic motion planning under temporal tasks and soft
  constraints\BBCQ\
\newblock {\Bem IEEE Transactions on Automatic Control}, {\Bem 63\/}(12),
  4051--4066.

\bibitem[\protect\BCAY{Hagberg, Swart,\ \BBA\ S~Chult}{Hagberg
  et~al.}{2008}]{hagberg2008exploring}
Hagberg, A., Swart, P., \BBA\ S~Chult, D. \BBOP2008\BBCP.
\newblock \BBOQ Exploring network structure, dynamics, and function using
  networkx\BBCQ\
\newblock \BTR, Los Alamos National Lab.(LANL), Los Alamos, NM (United States).

\bibitem[\protect\BCAY{Jamroga}{Jamroga}{2004}]{ATLPlanning}
Jamroga, W. \BBOP2004\BBCP.
\newblock \BBOQ Strategic planning through model checking of atl formulae\BBCQ\
\newblock In {\Bem International Conference on Artificial Intelligence and Soft
  Computing}, \BPGS\ 879--884. Springer.

\bibitem[\protect\BCAY{Jha, Raman, Sadigh,\ \BBA\ Seshia}{Jha
  et~al.}{2018}]{C2TLPlanning}
Jha, S., Raman, V., Sadigh, D., \BBA\ Seshia, S.~A. \BBOP2018\BBCP.
\newblock \BBOQ Safe autonomy under perception uncertainty using
  chance-constrained temporal logic\BBCQ\
\newblock {\Bem Journal of Automated Reasoning}, {\Bem 60\/}(1), 43--62.

\bibitem[\protect\BCAY{Kallenberg}{Kallenberg}{1983}]{kallenberg1983linear}
Kallenberg, L. C.~M. \BBOP1983\BBCP.
\newblock {\Bem Linear programming and finite Markovian control problems}.
\newblock Mathematisch Centrum, Amsterdam.

\bibitem[\protect\BCAY{Kemeny\ \BBA\ Snell}{Kemeny\ \BBA\
  Snell}{1963}]{kemeny1963markov}
Kemeny, J.\BBACOMMA\  \BBA\ Snell, J.~L. \BBOP1963\BBCP.
\newblock {\Bem Finite Markov chains}.
\newblock Springer-Verlag, New York.

\bibitem[\protect\BCAY{Krass\ \BBA\ Vrieze}{Krass\ \BBA\
  Vrieze}{2002}]{10.2307/3690451}
Krass, D.\BBACOMMA\  \BBA\ Vrieze, O.~J. \BBOP2002\BBCP.
\newblock \BBOQ Achieving target state-action frequencies in multichain
  average-reward {Markov} decision processes\BBCQ\
\newblock {\Bem Mathematics of Operations Research}, {\Bem 27\/}(3), 545--566.

\bibitem[\protect\BCAY{Kwiatkowska\ \BBA\ Parker}{Kwiatkowska\ \BBA\
  Parker}{2013}]{PLTLPlanning}
Kwiatkowska, M.\BBACOMMA\  \BBA\ Parker, D. \BBOP2013\BBCP.
\newblock \BBOQ Automated verification and strategy synthesis for probabilistic
  systems\BBCQ\
\newblock In {\Bem Automated Technology for Verification and Analysis}, \BPGS\
  5--22. Springer.

\bibitem[\protect\BCAY{Lakshmanan\ \BBA\ Bhatnagar}{Lakshmanan\ \BBA\
  Bhatnagar}{2012}]{QLearningConstrained_13}
Lakshmanan, K.\BBACOMMA\  \BBA\ Bhatnagar, S. \BBOP2012\BBCP.
\newblock \BBOQ A novel {Q-learning} algorithm with function approximation for
  constrained {Markov} decision processes\BBCQ\
\newblock In {\Bem 50th Annual Allerton Conference on Communication, Control,
  and Computing (Allerton)}, \BPGS\ 400--405. IEEE.

\bibitem[\protect\BCAY{Lazar}{Lazar}{1983}]{constrainedRoutingMotivation}
Lazar, A. \BBOP1983\BBCP.
\newblock \BBOQ Optimal flow control of a class of queueing networks in
  equilibrium\BBCQ\
\newblock {\Bem IEEE Transactions on Automatic Control}, {\Bem 28\/}(11),
  1001--1007.

\bibitem[\protect\BCAY{Lindemann\ \BBA\ Dimarogonas}{Lindemann\ \BBA\
  Dimarogonas}{2017}]{STLPlanning}
Lindemann, L.\BBACOMMA\  \BBA\ Dimarogonas, D.~V. \BBOP2017\BBCP.
\newblock \BBOQ Robust motion planning employing signal temporal logic\BBCQ\
\newblock In {\Bem 2017 American Control Conference (ACC)}, \BPGS\ 2950--2955.
  IEEE.

\bibitem[\protect\BCAY{Manne}{Manne}{1960}]{10.2307/2627340}
Manne, A.~S. \BBOP1960\BBCP.
\newblock \BBOQ Linear programming and sequential decisions\BBCQ\
\newblock {\Bem Management Science}, {\Bem 6\/}(3), 259--267.

\bibitem[\protect\BCAY{Nilsson, Hussien, Balkan, Chen, Ames, Grizzle, Ozay,
  Peng,\ \BBA\ Tabuada}{Nilsson et~al.}{2015}]{correctByConstruction}
Nilsson, P., Hussien, O., Balkan, A., Chen, Y., Ames, A.~D., Grizzle, J.~W.,
  Ozay, N., Peng, H., \BBA\ Tabuada, P. \BBOP2015\BBCP.
\newblock \BBOQ Correct-by-construction adaptive cruise control: Two
  approaches\BBCQ\
\newblock {\Bem IEEE Transactions on Control Systems Technology}, {\Bem
  24\/}(4), 1294--1307.

\bibitem[\protect\BCAY{Norris}{Norris}{1997}]{norris_1997}
Norris, J.~R. \BBOP1997\BBCP.
\newblock {\Bem Markov Chains}.
\newblock Cambridge Series in Statistical and Probabilistic Mathematics.
  Cambridge University Press.

\bibitem[\protect\BCAY{Petrik\ \BBA\ Zilberstein}{Petrik\ \BBA\
  Zilberstein}{2009}]{petrik2009bilinear}
Petrik, M.\BBACOMMA\  \BBA\ Zilberstein, S. \BBOP2009\BBCP.
\newblock \BBOQ A bilinear programming approach for multiagent planning\BBCQ\
\newblock {\Bem Journal of Artificial Intelligence Research}, {\Bem 35},
  235--274.

\bibitem[\protect\BCAY{Pistore, Bettin,\ \BBA\ Traverso}{Pistore
  et~al.}{2014}]{CTLPlanning}
Pistore, M., Bettin, R., \BBA\ Traverso, P. \BBOP2014\BBCP.
\newblock \BBOQ Symbolic techniques for planning with extended goals in
  non-deterministic domains\BBCQ\
\newblock In {\Bem Sixth European Conference on Planning}.

\bibitem[\protect\BCAY{Privault}{Privault}{2018}]{Privault2018}
Privault, N. \BBOP2018\BBCP.
\newblock {\Bem Understanding Markov Chains: Examples and Applications}.
\newblock Springer Singapore, Singapore.

\bibitem[\protect\BCAY{Puterman}{Puterman}{1994}]{puterman1994markov}
Puterman, M. \BBOP1994\BBCP.
\newblock {\Bem Markov decision processes : discrete stochastic dynamic
  programming}.
\newblock Wiley, New York.

\bibitem[\protect\BCAY{Ross}{Ross}{1989}]{10.2307/171066}
Ross, K.~W. \BBOP1989\BBCP.
\newblock \BBOQ Randomized and past-dependent policies for {Markov} decision
  processes with multiple constraints\BBCQ\
\newblock {\Bem Operations Research}, {\Bem 37\/}(3), 474--477.

\bibitem[\protect\BCAY{Royden\ \BBA\ Fitzpatrick}{Royden\ \BBA\
  Fitzpatrick}{2010}]{royden}
Royden, H.\BBACOMMA\  \BBA\ Fitzpatrick \BBOP2010\BBCP.
\newblock {\Bem Real Analysis\/} (4th \BEd).
\newblock Pearson.

\bibitem[\protect\BCAY{Schwarting, Alonso-Mora,\ \BBA\ Rus}{Schwarting
  et~al.}{2018}]{motivation}
Schwarting, W., Alonso-Mora, J., \BBA\ Rus, D. \BBOP2018\BBCP.
\newblock \BBOQ Planning and decision-making for autonomous vehicles\BBCQ\
\newblock {\Bem Annual Review of Control, Robotics, and Autonomous Systems},
  {\Bem 1\/}(1), 187--210.

\bibitem[\protect\BCAY{Skwirzynski}{Skwirzynski}{1981}]{steadyStateNetworkBehavior}
Skwirzynski, J.~K. \BBOP1981\BBCP.
\newblock {\Bem New concepts in multi-user communication},
  \lowercase{\BVOL}~43.
\newblock Springer Science \& Business Media.

\bibitem[\protect\BCAY{Song, Feng,\ \BBA\ Zhang}{Song
  et~al.}{2015}]{PCTLPlanning}
Song, L., Feng, Y., \BBA\ Zhang, L. \BBOP2015\BBCP.
\newblock \BBOQ Planning for stochastic games with co-safe objectives\BBCQ\
\newblock In {\Bem Twenty-Fourth International Joint Conference on Artificial
  Intelligence}.

\bibitem[\protect\BCAY{Tarjan}{Tarjan}{1972}]{Tarjan1972DepthFirstSA}
Tarjan, R.~E. \BBOP1972\BBCP.
\newblock \BBOQ Depth-first search and linear graph algorithms\BBCQ\
\newblock {\Bem SIAM Journal on Computing}, {\Bem 1}, 146--160.

\bibitem[\protect\BCAY{Tian}{Tian}{2019}]{debrisMitigation}
Tian, Z. \BBOP2019\BBCP.
\newblock \BBOQ United states law and policy on space debris\BBCQ\
\newblock In {\Bem Space Security and Legal Aspects of Active Debris Removal},
  \BPGS\ 155--167. Springer.

\bibitem[\protect\BCAY{Trevizan, Thi{\'e}baux,\ \BBA\ Haslum}{Trevizan
  et~al.}{2017}]{occupationMeasurePlanning}
Trevizan, F., Thi{\'e}baux, S., \BBA\ Haslum, P. \BBOP2017\BBCP.
\newblock \BBOQ Occupation measure heuristics for probabilistic planning\BBCQ\
\newblock In {\Bem Twenty-Seventh International Conference on Automated
  Planning and Scheduling}.

\bibitem[\protect\BCAY{Trevizan, Thi{\'e}baux, Santana,\ \BBA\
  Williams}{Trevizan et~al.}{2016}]{iDual}
Trevizan, F., Thi{\'e}baux, S., Santana, P., \BBA\ Williams, B. \BBOP2016\BBCP.
\newblock \BBOQ Heuristic search in dual space for constrained stochastic
  shortest path problems\BBCQ\
\newblock In {\Bem Twenty-Sixth International Conference on Automated Planning
  and Scheduling}.

\bibitem[\protect\BCAY{Trevizan, Thi{\'e}baux, Santana,\ \BBA\
  Williams}{Trevizan et~al.}{2017}]{iDual2}
Trevizan, F., Thi{\'e}baux, S., Santana, P., \BBA\ Williams, B. \BBOP2017\BBCP.
\newblock \BBOQ I-dual: solving constrained ssps via heuristic search in the
  dual space\BBCQ\
\newblock In {\Bem Proceedings of the 26th International Joint Conference on
  Artificial Intelligence}, \BPGS\ 4954--4958.

\bibitem[\protect\BCAY{Velasquez}{Velasquez}{2019}]{IJCAI2019}
Velasquez, A. \BBOP2019\BBCP.
\newblock \BBOQ Steady-state policy synthesis for verifiable control\BBCQ\
\newblock In {\Bem Proceedings of the 28th International Joint Conference on
  Artificial Intelligence}, \BPGS\ 5653--5661. AAAI Press.

\bibitem[\protect\BCAY{Wongpiromsarn, Topcu, Ozay, Xu,\ \BBA\
  Murray}{Wongpiromsarn et~al.}{2011}]{TuLiP}
Wongpiromsarn, T., Topcu, U., Ozay, N., Xu, H., \BBA\ Murray, R.~M.
  \BBOP2011\BBCP.
\newblock \BBOQ {TuLiP}: a software toolbox for receding horizon temporal logic
  planning\BBCQ\
\newblock In {\Bem Proceedings of the 14th International Conference on Hybrid
  Systems: Computation and Control}, \BPGS\ 313--314. ACM.

\bibitem[\protect\BCAY{Wu\ \BBA\ Durfee}{Wu\ \BBA\
  Durfee}{2010}]{wu_durfee_jair_2010}
Wu, J.\BBACOMMA\  \BBA\ Durfee, E.~H. \BBOP2010\BBCP.
\newblock \BBOQ Resource-driven mission-phasing techniques for constrained
  agents in stochastic environments\BBCQ\
\newblock {\Bem Journal of Artificial Intelligence Research (JAIR)}, {\Bem 38},
  415--473.

\bibitem[\protect\BCAY{Zhou, Maity,\ \BBA\ Baras}{Zhou
  et~al.}{2016}]{MTLPlanning}
Zhou, Y., Maity, D., \BBA\ Baras, J.~S. \BBOP2016\BBCP.
\newblock \BBOQ Timed automata approach for motion planning using metric
  interval temporal logic\BBCQ\
\newblock In {\Bem 2016 European Control Conference (ECC)}, \BPGS\ 690--695.
  IEEE.

\end{thebibliography}
